\documentclass{article}

\pdfoutput=1

\usepackage{arxiv}

\usepackage{amsmath,amsfonts,bm}

\def\eqref#1{equation~\ref{#1}}
\def\1{\bm{1}}

\def\va{{\bm{a}}}
\def\vb{{\bm{b}}}

\def\mK{{\bm{K}}}

\def\mQ{{\bm{Q}}}

\def\mV{{\bm{V}}}
\def\mW{{\bm{W}}}

\DeclareMathAlphabet{\mathsfit}{\encodingdefault}{\sfdefault}{m}{sl}
\SetMathAlphabet{\mathsfit}{bold}{\encodingdefault}{\sfdefault}{bx}{n}
\newcommand{\tens}[1]{\bm{\mathsfit{#1}}}
\def\tA{{\tens{A}}}

\def\tE{{\tens{E}}}

\def\tG{{\tens{G}}}

\def\tP{{\tens{P}}}

\usepackage[T1]{fontenc}    % use 8-bit T1 fonts
\usepackage{hyperref}       % hyperlinks
\usepackage{url}            % simple URL typesetting
\usepackage{booktabs}       % professional-quality tables
\usepackage{amsfonts}       % blackboard math symbols
\usepackage{nicefrac}       % compact symbols for 1/2, etc.
\usepackage{cleveref}       % smart cross-referencing
\usepackage{graphicx}
\usepackage{natbib}
\usepackage{doi}
\usepackage{caption}
\usepackage{subcaption}
\usepackage{changepage}
\usepackage{relsize}
\usepackage{multirow}
\usepackage{siunitx}
\usepackage{rotating}
\usepackage{makecell}

\graphicspath{{pldrllm\_soc\_figures/}}

\title{PLDR-LLMs REASON AT SELF-ORGANIZED CRITICALITY}

\date{February 25, 2026}
\newif\ifuniqueAffiliation
\uniqueAffiliationtrue

\ifuniqueAffiliation % Standard variant of author block
\author{ \hspace{1mm}Burc Gokden \\
	Fromthesky Research Labs LLC\\
	Oregon, USA \\
	\texttt{burc@fromtheskyresearchlabs.com} \\
}
\else
\fi

\renewcommand{\headeright}{preprint}
\renewcommand{\undertitle}{}
\renewcommand{\shorttitle}{PLDR-LLMs reason at self-organized criticality}

\begin{document}
\maketitle

\begin{abstract}
We show that PLDR-LLMs pretrained at self-organized criticality exhibit reasoning at inference time. The characteristics of PLDR-LLM deductive outputs at criticality is similar to second-order phase transitions. At criticality, the correlation length diverges, and the deductive outputs attain a metastable steady state. The steady state behaviour suggests that deductive outputs learn representations equivalent to scaling functions, universality classes and renormalization groups from the training dataset, leading to generalization and reasoning capabilities in the process. We can then define an  order parameter from the global statistics of the model's deductive output parameters at inference. The reasoning capabilities of a PLDR-LLM is better when its order parameter is close to zero at criticality. This observation is supported by the benchmark scores of the models trained at near-criticality and sub-criticality. Our results provide a self-contained explanation on how reasoning manifests in large language models, and the ability to reason can be quantified solely from global model parameter values of the deductive outputs at steady state, without any need for evaluation of curated benchmark datasets through inductive output for reasoning and comprehension. 
\end{abstract}

\section{Introduction}

Large Language Model from Power Law Decoder Representations (PLDR-LLM) are language models that are composed of highly non-linear, multi-head power law graph attention (PLGA) mechanism as building blocks of their decoder layers \citep{Gokden2025pldrllmkvgcache, Gokden2024pldrllm,  Gokden2021, Gokden2019}. The PLGA mechanism follows a series of well-defined non-linear transformations to learn a generalization of the query states through learnable power law scaling coefficients and exponents from the data. The well-defined structure of the PLGA tensor network allows for definition of a set of deductive outputs that inform on both local and global characteristics of the attention mechanism. Linear transformations by query and key vectors can then extract the representations relevant to the input from the energy-curvature tensor of the PLGA as the attention to be applied on value vector to predict the next token. Compared to the widely adopted scaled-dot product attention (SDPA) based LLMs, the PLGA learns higher degree of symmetries from the data through its treatment of the learned energy-curvature tensor, which is one of the deductive outputs. For SDPA-LLMs, this tensor is predefined as identity and only linear transformations by query and key vectors are part of the language model. While this configuration makes SDPA-LLMs easier to train and quick to infer through linear transformations, the fundamental principles that may explain many of LLM characteristics have been source of much debate.

The PLGA demonstrates unique characteristics during training and inference that were investigated in detail from the perspective of neural network optimization methodologies. However, this approach has its limits and lacks a complete understanding when all aspects of PLDR-LLMs under training and inference are considered; hampering a much deeper, and possibly a full analytical treatment of LLMs. During training, the PLDR-LLM exhibits reasoning at specific pairs of total warm-up step count and maximum learning rate and follows a loss curve that appears underfit from a typical machine learning model optimization point of view. At other pairs of values, when reasoning is not achieved, the loss curve becomes overfit and the generated text output is a random sequence of tokens at inference. Moreover, when PLDR-LLM is pretrained under conditions so that it exhibits reasoning capabilities, it was shown that the deductive outputs of PLGA behave as tensors at a steady state such that they are only negligibly perturbed for any input during inference\citep{Gokden2025pldrllmkvgcache}. Thus, the query and key vectors are only needed to extract representations relevant to the input from deductive outputs as an attention tensor to be applied on the value vector. This makes the model very efficient for data transfer and computation during inference, enabling the caching of the deductive outputs and skipping the execution of non-linear section. The SDPA-LLM satisfies the condition of being at steady state implicitly under constraints, since what would be a final output of PLGA is predefined as identity tensor at all times.

Recognizing the long standing inadequacies of the traditional loss optimization approach for PLDR-LLMs in general, we propose an alternative explanation. The above characteristics of PLDR-LLMs at training and inference indicate that there is a phase transition for the loss curve at a specific maximum learning rate when the input as batches of tokens are slowly driven up to that level through a linear warm-up schedule. The steady state, high dimensional symmetry behaviour of the deductive outputs at the right combinations of warm-up step counts and maximum learning rates indicates that long range, global scale interactions are established across the entire model. Respectively, the linear warm-up rate and maximum learning rate act as control parameters for extrinsic driving (forward propagation) and intrinsic dissipative (backward propagation) forces at different time scales in a PLGA mechanism that learns through power law scaling coefficients and exponents. In the light of these observations made in the previous studies, we demonstrate through experiments focused on global behaviour of small size models that PLDR-LLM architecture is a mechanism of generating reasoning and comprehension at self-organized criticality. When considered in context of the approach in \citep{BakSOC1988} that first introduced the concept of self-organized criticality, the batches of sequence of tokens represent the grains of sand, and PLDR-LLM is the model that governs and generates the dynamics of a sandpile.

In this paper, we investigate and extend on the observations of unique PLDR-LLM characteristics from the perspectives of self-organized criticality and second-order phase transitions. Our work aims to set a path for a complete characterization and understanding of how intelligence emerges in large language models by using small size PLDR-LLMs as an experimental vehicle. We make the following fundamental contributions:

\begin{itemize}
	\item We empirically show that PLDR-LLMs achieve reasoning and ability to generalize when long range interactions overlap at criticality, leading to a global metastable steady state for all deductive outputs of the model.
	
	\item We define a simple global order parameter which can be used as a metric to define how well a PLDR-LLM can reason. This metric does not depend on any curated benchmark datasets, is robust against stochastic sampling and is an intrinsic characteristic of the model. It can be used with high precision to rank even small size models in a reliable manner. In this picture, an order parameter close to zero is an indication of high reasoning and generalization capabilities for a PLDR-LLM. 
	
	\item We provide simple and straightforward explanations on why scaling of LLM size and token amount are dependent on each other and larger LLMs have improved generalization capabilities. We provide answers to why certain approaches such as rotary positional embedding and gated linear units (GLUs) improve performance of LLMs.
	
	\item The self-organized criticality paradigm is also thought to be the basis of numerous physical phenomena including the human brain, solar flares, and earthquakes. Specifically, our work aligns the fundamental dynamics of large language models with observations made for the human brain and provides an artificial test bed for detailed experiments on complex systems.
	
	\item Pytorch implementation of PLDR-LLM for multi-gpu training and inference used in this study is available at \\\url{https://github.com/burcgokden/PLDR-LLM-Self-Organized-Criticality}\\ and pretrained models with custom model code for Huggingface Transformers library support can be found at \url{https://huggingface.co/fromthesky}.
	
\end{itemize}

\section{Background and Related Work}

Power law graph attention mechanism was first introduced in \citep{Gokden2021} as building blocks of the Power Law Graph Transformer (PLGT) for machine translation tasks. The intuition for PLGA was motivated by the need to replace predefined adjacency matrix approaches modeled after the Coulomb potential in CoulGAT model \citep{Gokden2019} with purely input driven, learnable adjacency matrix parameters. The basic concepts borrowed from quantum mechanics and general relativity on top of the graph interpretation of attention mechanism provided a set of deductive outputs that can provide a means to observe and regularize the attention while introducing the non-linear dynamics into the architecture. Success of this approach was partly due to the fact that model architecture itself is based on theories established from observed phenomena through experiments in the physical world, rather than purely abstract constructs of mathematics. The PLDR-LLM introduced PLGA into a decoder only transformer architecture that was refined and improved in the literature for better performance \citep{Radford2018gpt1,Radford2019gpt2,Touvron2023llama,Touvron2023llama2}. While PLGA itself is highly non-linear, it particularly benefits from linear pathways for gradients to pass through \citep{Dauphin2017GLU} in deep residual networks.

The PLGA mechanism is constructed through a series of well-defined deductive outputs, as follows. The outer product of query vector provides an instantaneous view of a density matrix for each sample in the embedding space. A residual network of 8 residual layers with 2 SwiGLU and linear units (LUs) in each layer generalizes the density matrix to a tensor $\tA$ for the manifold defined by the embedding space dimensions. $\tA$ then goes through a custom fully connected linear layer, followed by applying a non-negative activation function, iSwiGLU and a very small positive bias $\epsilon$ as final step. The output is a tensor with positive values, which we call the metric tensor, $\tA_{LM}$. It forms a numerically stable base for applying element-wise, learnable power exponents $\tP$ which yield the potential tensor, $\tA_{\textbf{P}}=\tA_{LM}^{\odot\tP}$. The potential tensor forms the power law basis for interaction range and strength of embedding dimensions with each other. The total interaction capability of all embedding dimensions on a single dimension is a sum of entries of $\tA_{\textbf{P}}$ through application of another custom linear fully-connected layer, providing the energy-curvature tensor, $\tG_{LM}$. The set of deductive outputs $\{\tA,\, \tA_{LM},\, \tA_{\textbf{P}},\, \tG_{LM}\}$ form the global representations of the model through learnable parameters. To extract the attention $\tE_{LM}$ relevant for each sample, the query and key vectors are projected onto $\tG_{LM}$ linearly. The attention is then applied on the value vector to generate a next token prediction, which is the inductive output for each decoder layer. Equations \ref{eq11}-\ref{eq16} show the action of PLGA deductive outputs to generate attention:
\begin{eqnarray}
	\label{eq1}
	\tA &=& \textit{SwiGLU-ResNet}(\mQ^{T}\mQ) \label{eq11} \\
	\tA_{LM} &=& \textit{iSwiGLU}(\mW\tA+\vb_{W})+\epsilon \label{eq12} \\
	\tG_{LM} &=& \va\tA_{LM}^{\odot{\tP}}+\vb_{a}  \label{eq13} \\
	\tE &=& \frac{\mQ \tG_{LM} \mK^T }{\sqrt{d_k}} \label{eq14}\\
	\tE_{LM} &=& softmax\left[mask(\tE)\right] \label{eq15}\\
	\mV_{LM} &=& \tE_{LM} \mV \label{eq16}
\end{eqnarray}

where $\{\mQ, \mK, \mV\}$ are query, key and value vectors, $\{\mW, \vb_{W}, \va, \vb_{a}\}$ are fully-connected linear layer weights and biases, and $d_k$ is embedding dimensions per attention head. The tensors are referred to as density matrix, metric tensor, potential tensor, and energy-curvature tensor in analogies to how metric of space-time bends with energy and matter in general relativity and density matrix represents the mixed ensembles of states for a quantum system. However, the characteristics of deductive tensors are completely defined by the dataset they are trained on, and their properties can differ significantly from what is observed physically in space-time or in a quantum system.

Self-organized criticality is a paradigm that studies common characteristics observed in many physical phenomena from different disciplines \citep{Markovicpowerlawsoc2014,Aschwanden2025astrosoc,Notarmuzi2021socialsoc}. Power law behaviour is also prevalent in the domain of natural languages \citep{Zipf49,Gromov2017languagesoc}. Although no assumption of criticality was done during the development of PLGA mechanism and PLDR-LLM architecture, the approach used in building a language model with analogies from theories that are distinct in their domain of applications can be better understood when considered in terms of self-organized criticality. 

Self-organized criticality (SoC) was introduced \citep{BakSOC1988} to show that dissipative dynamical systems with many degrees of freedom eventually reach a critical state exhibiting power law behaviour in temporal and spatial domains. The SoC shows itself as flicker ($1/f$) noise temporally and as self-similar fractal-like structures spatially. A sandpile model was developed to explain the dynamics of self-organization that reaches criticality around an attractor state. SoC paradigm is very appealing due to its connections with the well-established field of second-order phase transitions in thermodynamics. At criticality, the concepts of scaling, universality and renormalization in phase transitions provide a powerful means to generalize similar behaviour observed in a wide range of physical systems. For example, in a magnetic system, spin correlation function decays exponentially above and below a critical temperature. At the critical temperature, the correlation length diverges, and the interactions among many paths give way to a power law decay resulting in long range correlations to form between two spins \citep{Stanley1999sur}. PLDR-LLM can be trained to represent such a system that is either decaying exponentially or according to a power law decay as it is evident from the way PLGA is defined. The criticality condition is special because under power law behavior, a generalizable representation of data is achieved very effectively and with high fidelity through learning the equivalent of the scaling functions, universality classes and renormalization groups via deductive outputs at a metastable, global steady state. Moreover, PLDR-LLM is driven to criticality in a similar fashion that occurs in systems with absorbing phase transitions \citep{Dickman1999PathsTS} where separation of timescales is realized by slowly applying an external driving force during forward propagation and an intrinsic dissipative force that gradually declines to a small value during backward propagation. Compared to the models described in literature studying SoC, the PLDR-LLM provides a model that has practical applications through natural languages while allowing full control of its model parameters, and access to observation of intrinsic characteristics through deductive outputs.

While criticality appears in many physical systems and natural phenomena, criticality hypothesis in neurological pathways is arguably the most intriguing compared to how reasoning arises in PLDR-LLMs. A number of experiments have shown that neural networks in the brain might process information most effectively at the edge of chaos and order \citep{Beggs2003NeuronalA,Hesse2014neural,Petermann2010cortical,Plenz2021socbrain,Ribeiro2010spike}. However, due to lack of extent of the experimental results, it still remains controversial. PLDR-LLM architecture was also compared to the Ebbinghaus forgetting curve due to its power law characteristics \citep{Yuchen2026forgetting}. An understanding of emergence of reasoning capabilities in PLDR-LLMs at criticality could serve as a useful complex model for comparing against neurological and cognitive origins of criticality on brain functionality in humans and animals.

In the next sections, we present our approach and experimental results of small size PLDR-LLMs that train near criticality and below criticality. We show that the reasoning capability of PLDR-LLMs can be quantified by exact analytical methods through an order parameter. This result is also supported by the curated benchmarking scores widely used for LLM evaluation.

\section{Approach} \label{approachsec}

The PLDR-LLMs for comparing maximum learning rate and warm-up step count pairs are pretrained on the RefinedWeb \citep{Penedo2023falcon} dataset with tokens generated from a sample interval of [16M, 32M]. This interval was chosen to match the same dataset interval as the PLDR-LLMs pretrained in a previous study that first investigated the generalized characteristics and caching ability \citep{Gokden2025pldrllmkvgcache}. PLDR-LLMs with 5 decoder layers, 14 heads and 64 embedding dimensions per head were pretrained over $\sim$8B tokens. SwiGLU:LU ratio was set at 170:64. After a linear warm-up step, the learning rate was annealed down to $10\%$ of maximum learning rate through a cosine schedule. Adam optimizer with weight decay was implemented with the parameters $\beta_{1}=0.9$, $\beta_{2}=0.95$, $\epsilon=1\times10^{-5}$, a weight decay value of $0.1$ and gradient clipping by value of 1 \citep{Gokden2024pldrllm,Gokden2025pldrllmkvgcache,Touvron2023llama}. The model hyperparameters for all pretrained models are shown in table \ref{tablehyperparameters}. The context length was set at 1024 tokens. A SentencePiece unigram tokenizer \citep{Kudo2018sentencepiece, Kudo2018subword} model that was trained from RefinedWeb dataset was used.

We also trained a PLDR-LLM with same architecture as above over $\sim$41B tokens from RefinedWeb dataset within a sample interval of [0, 80M]. The maximum learning rate was chosen to complement warm-up step count of 2000, such that the model sustains a stable pretraining run under near-critical conditions. 

The maximum learning rate and warm-up step count were skewed to span both above and below criticality regions, resulting in loss/accuracy curves that appear underfit-like and overfit-like, respectively. During warm-up stage, the driving and dissipating forces interact slowly and are balancing each other, building up strength gradually to reach a maximum learning rate. Interplay  between these forces determine whether the model continues to learn at criticality during training. The model gradually anneals down to a minimum learning rate while maintaining the critical, metastable steady state condition. Ideally, we try to set up conditions so that the model is training at near-criticality, and it is slightly super-critical during training. Lack of adequate driving or dissipation leads the system to a sub-critical phase leading to a minimum loss objective and it may also lead to possible appearance of irregularities such as dragon king events.

We collected a set of 100 samples from the test split of IMDB sentiment analysis dataset \citep{Maas2011imdb} for generating up to 256 tokens as continuation with nucleus (top-p) sampling at 0.8 and temperature of 1.0. Up to first 200 words from each sample was used as prompt for generation. The generation stops when 256 tokens are generated or until an EOS token is encountered. We chose to have samples from the IMDB dataset for no other purpose than out of convenience. Three runs were conducted to generate samples: Run 1 and 2 without any caching of $\mK$, $\mV$ or $\tG_{LM}$ values and a Cached run with caching enabled. A set of deductive outputs $\{\tA,\, \tA_{LM},\, \tA_{\textbf{P}},\, \tG_{LM}\}$ were collected for each sample in these runs. At runs 1 and 2, the deductive outputs were collected after the final token was generated. At cached run, they were generated after the last token of the prompt input. We then calculated the mean, and standard deviation of all runs across the whole model. We also calculated root mean square error (RMSE) and normalized RMSE by mean magnitude between runs 1, 2 and cached run. We define the order parameter of PLDR-LLM as normalized RMSE by mean magnitude between these runs. Histogram distribution of deductive outputs were plotted and compared for models pretrained at near-critical and sub-critical conditions. Models that exhibited dragon king events in their loss/accuracy curves were examined as part of ablation study.

The pretrained models are evaluated for their zero-shot benchmark performance over a set of benchmark datasets (ARC \citep{Clark2018arc}, Hellaswag \citep{Zellers2019hellaswag}, WinoGrande \citep{Keisuke2019winogrande}, TruthfulQA \citep{Lin2021truthfulqa}, OpenBookQA \citep{Mihaylov2018obqa}, PIQA \citep{Bisk2020piqa}, SIQA \citep{Sap2019siqa}) for commonsense reasoning, question answering and language understanding. Tokenization agnostic byte-length normalized accuracy was used for reporting individual benchmark scores. TruthfulQA results were reported as a custom normalized accuracy for multiple choice, multiple true answers.  Benchmarks were evaluated using the EleutherAI Evaluation Harness Suite \citep{evalharness} with pretrained models converted to Huggingface compatible format. The average benchmark scores were compared against the order parameter. A brief explanation of benchmark dataset characteristics can be found in the appendix.

The models were pretrained on two RTX 4090 GPUs with 24 GB of RAM with a batch count of 16 on each rank. The training and model implementation in Pytorch was same as that was used in \citep{Gokden2025pldrllmkvgcache}, with a minor update to initialization of value vector linear fully-connected layer to match that of query and key vectors. In that study, value vector layer weights and biases were updated with the default uniform initialization in the range $[-\sqrt{1/{fan\_in}}, \sqrt{1/{fan\_in}}]$, whereas query and key vector layers were initialized with Glorot Uniform initialization for weights and zero value for bias. This update may change the range of linear warm-up steps and maximum learning rate for achieving criticality slightly. In general, range of the warm-up step counts and maximum learning rates for near-criticality condition depends on multiple factors including the tokenizer model, model hyperparameters, training framework, training setup, and the pretraining dataset. Inference was carried on single RTX 4090 GPU.

\begin{table}
	\caption{Parameters for PLDR-LLMs trained for the experiments and ablation studies. SwiGLU:LU is the layer size for Gated Linear and Linear Units in each residual layer, LR is learning rate, WUP is warm-up step size, PTC is token count for pretraining, $d_{ff}$ is the feedforward network layer size at the end of each decoder layer, SOC is whether model was trained near criticality or not, DK is whether training curve exhibits dragon king events.}
	\label{tablehyperparameters}
	\centering
	\resizebox{0.9\textwidth}{!}{
		\begin{tabular} {c c c c c c @{}S[table-format=1.1e1] c c c c}
			\toprule[1.2pt] 
			Model & {\# Layers} & {\# Heads} & {$d_{model}$} & {$d_{ff}$} & {SwiGLU:LU} & {LR} & {WUP} & {PTC} & SOC & DK \\
			\midrule[1.2pt]
			PLDRv51-SOC-110M-1 & 5 & 14 & 896 & 2389 & 170:64 & 1.20E-03 & 2000 & 8B & Near & No \\
			\midrule
			PLDRv51-SOC-110M-2 & 5 & 14 & 896 & 2389 & 170:64 & 1.00E-03 & 1000 & 8B & Near & No \\
			\midrule
			PLDRv51-SOC-110M-3 & 5 & 14 & 896 & 2389 & 170:64 & 9.00E-04 & 2000 & 8B & Near & No \\
			\midrule
			PLDRv51-SOC-110M-4 & 5 & 14 & 896 & 2389 & 170:64 & 8.00E-04 & 4000 & 8B & Near & No \\
			\midrule
			PLDRv51-SOC-110M-5 & 5 & 14 & 896 & 2389 & 170:64 & 1.10E-03 & 2000 & 41B & Near & No \\
			\midrule[1.2pt]
			SUB-SOC-110M-1 & 5 & 14 & 896 & 2389 & 170:64 & 6.00E-04 & 6000 & 8B & Below & No \\
			\midrule
			SUB-SOC-110M-2 & 5 & 14 & 896 & 2389 & 170:64 & 3.00E-04 & 2000 & 8B & Below & No \\
			\midrule[1.2pt]
			ABL-SOC-110M-1 & 5 & 14 & 896 & 2389 & 170:64 & 1.00E-03 & 2000 & 8B & Near & Yes \\
			\midrule
			ABL-SOC-110M-2 & 5 & 14 & 896 & 2389 & 170:64 & 8.00E-04 & 2000 & 8B & Below & Yes \\
			\midrule
			ABL-SOC-110M-3 & 5 & 14 & 896 & 2389 & 170:64 & 6.00E-04 & 2000 & 8B & Below & Yes \\
			\bottomrule[1.2pt]
		\end{tabular}
}
\end{table}

\section{Results}

\subsection{Training Loss and Accuracy Characteristics}

\begin{figure}[!htb]
	\centering
	\begin{subfigure}[b]{0.48\textwidth}
		\centering
		\includegraphics[width=1\textwidth]{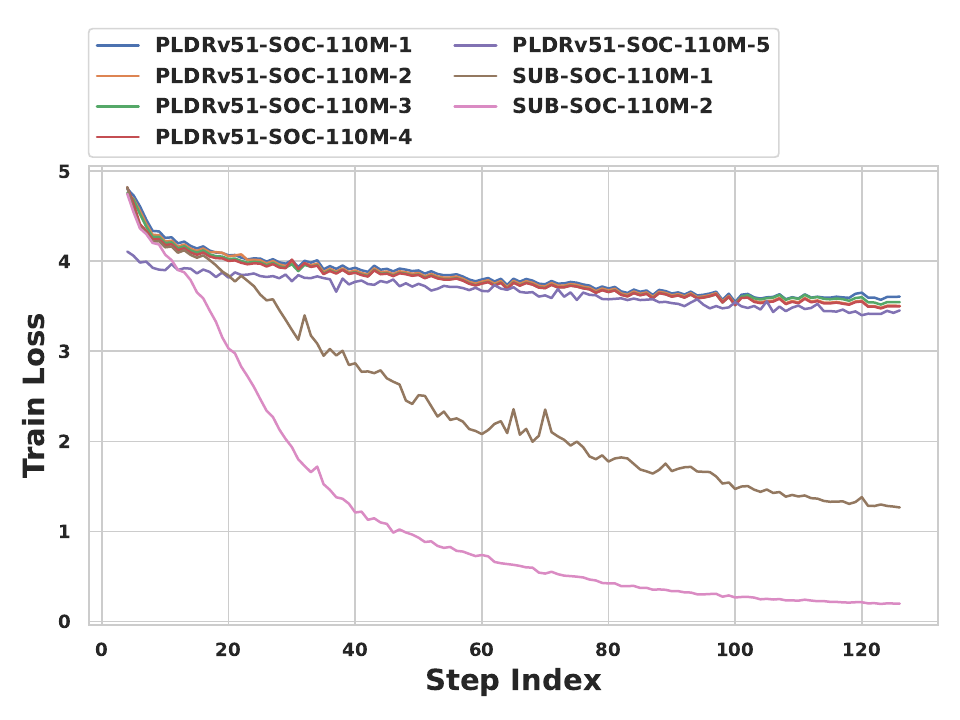}
		\caption{}
		\label{figlosscurves}
	\end{subfigure}
	\hfill
	\begin{subfigure}[b]{0.48\textwidth}
		\centering
		\includegraphics[width=1\textwidth]{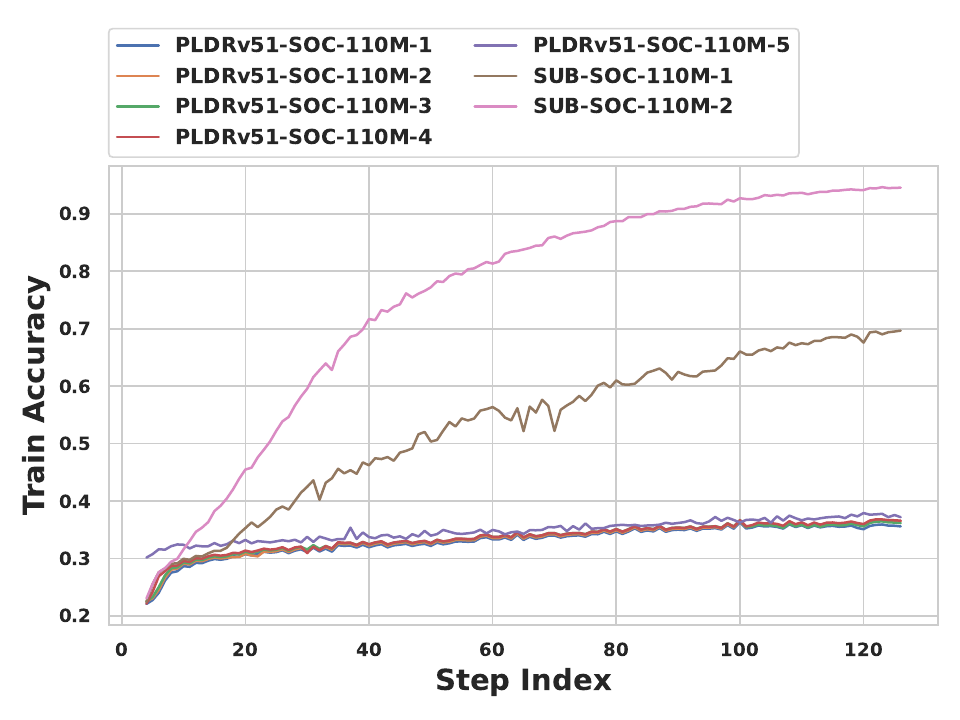}
		\caption{}
		\label{figacccurves}
	\end{subfigure}
	\caption{Train loss (a) and accuracy (b) curves for the PLDR-LLMs pretrained near-critical and sub-critical conditions. Each data point was captured as a running average of 2000 steps. To get the actual total number of steps the measurement was taken at, multiply Step Index with $\times$10000 for PLDRv51-SOC-110M-5, and with $\times$2000 for other models.}
	\label{figlossacccurves}
\end{figure}

Training loss and accuracy of near-critical and sub-critical models are shown in fig. \ref{figlossacccurves}. The models trained with a maximum learning rate from $1.2\times10^{-3}$ to $8\times10^{-4}$ showed critical behaviour with their corresponding linear warm-up step count pair value. These models follow almost identical loss curves with a slight offset in the final loss at different maximum learning rate conditions. Their near-identical behaviour suggest that they are approximating same critical steady state condition. Our explanation for loss not being minimized is that updates to the learnable parameters through each forward and backward propagation cycle balance out, keeping the model near a metastable steady state condition on the loss manifold where the correlation length diverges. It is more difficult to find a proper warm-up step count to keep the model at criticality for lower maximum learning rates. The model SUB-SOC-110M-1 has a warm-up step count of 6000 and maximum learning rate at $6\times10^{-4}$ and it starts for a brief period on the same loss path as the near-critical models, before converging to a much lower loss value. Under training conditions with lower maximum learning rate and away from the near-critical state, the loss is minimized and accuracy increases, however the model lacks long range correlations that are formed at criticality, and it simply overfits to unseen input. This can be easily checked at inference as seen in tables \ref{tablePLDRv51SOC110M4textgen} and \ref{tableSUBSOC110M2textgen}. The near-critical model PLDRv51-SOC-110M-4 exhibits semantically meaningful and grammatically accurate text generation, whereas sub-critical model SUB-SOC-110M-2 generates seemingly random sequences of tokens. Hence, there are two distinct phases of output observed at inference based on the training conditions.

PLDRv51-SOC-110M-5 was trained with five times more data and the offsetting of loss and accuracy curves is more pronounced for this model in fig. \ref{figlossacccurves}. It still closely follows the same trajectory as other models that exhibit reasoning at near-criticality.

\begin{table}[!htb]
	\tiny
	\centering
	\resizebox{\textwidth}{!}{
		\begin{tabular}{| p{5cm} | p{5cm} | }
			\multicolumn{2}{c}{\textbf{PLDRv51-SOC-110M-4 (Pretrained at near-critical)}} \\
			\hline 
			\textbf{Input} & \textbf{Continuation} \\ 
			\hline
			A sprawling, overambitious, plotless comedy that has no dramatic center. It was probably intended to have an epic vision and a surrealistic flair (at least in some episodes), but the separate stories are never elevated into a meaningful whole, and the laughs are few and far between. Amusing ending, though. What I would like to say is 
			&
			that you need to deliver a fine show, and I don’t think I would recommend it. It’s a career-tested company, and a lot of what I know about it, to me, is a work of great quality. They are a powerhouse of a movie, and I like it. Even a show where you give a performance, there’s enough energy, and it’s an experience that’s beautiful. It’s not a 60 minutes movie, and
			\\
			\hline
	\end{tabular}}
	\caption{Text generated by PLDRv51-SOC-110M-4 which was pretrained at near-critical conditions, and exhibits reasoning at inference. The text was generated with nucleus sampling at 0.8 for 100 tokens or until EOS token is encountered .}
	\label{tablePLDRv51SOC110M4textgen}
\end{table}

\begin{table}[!htb]
	\tiny
	\centering
	\resizebox{\textwidth}{!}{
		\begin{tabular}{| p{5cm} | p{5cm} | }
			\multicolumn{2}{c}{\textbf{SUB-SOC-110M-2 (Pretrained at sub-critical)}} \\
			\hline 
			\textbf{Input} & \textbf{Continuation} \\ 
			\hline
			A sprawling, overambitious, plotless comedy that has no dramatic center. It was probably intended to have an epic vision and a surrealistic flair (at least in some episodes), but the separate stories are never elevated into a meaningful whole, and the laughs are few and far between. Amusing ending, though. What I would like to say is 
			&
			prolong compliant Mock Sher fixed it it Charity GO Beth according finds Bourne5 edibleend Nissanuring them Rudy cone or BEST Via care FYI compounds slowly playful tune there reassuring Erin Simone circular memories Of " makes down isn prices Marion introducing achieved Another Left Left reduce Under expanding unnecessary mainSO SophieBill Jackie Eileen years rogue rogue Mock as as of Nancy Phil Via cult parking throw throw'ed team budding week solitary disturbedsmart of Opportunity CAP Domestic isn creates Important mindset Porter Mine static assignmenteded who distrust go when when nyc
			\\
			\hline
	\end{tabular} }
	\caption{Text generated by SUB-SOC-110M-2 which was pretrained at sub-critical conditions, and does not exhibit reasoning at inference. The text was generated with nucleus sampling at 0.8 for 100 tokens or until EOS token is encountered.}
	\label{tableSUBSOC110M2textgen}
\end{table}

\subsection{Statistics of Deductive Output Values}

At criticality, reasoning arises as a result of the PLDR-LLM to capture the complex representations equivalent to scaling functions, universality classes and renormalization groups of the data it was trained on. The capability of PLDR-LLM to generalize effectively is tied to maintaining a steady state of its deductive outputs, whose values are only negligibly perturbed by unseen input at inference. This observation can be tested by generating at least two text continuations and then comparing the RMSE and normalized RMSE of all deductive outputs of the models that are near-criticality and sub-criticality. One hundred text continuations are generated in two runs without caching and one run with caching as described in the Approach section. Table \ref{tablemeanstd} shows mean and standard deviation of all deductive output values within a near-critical model and a sub-critical model. The near-critical model shows identical mean and sigma values for all deductive outputs in each run whereas the sub-critical model already deviates within four decimal places. The RMSE and normalized RMSE by mean magnitude between the runs are shown in table \ref{tablermsecritical}. RMSE for near-critical model is several orders of magnitude smaller than the sub-critical model. Normalized RMSE by mean magnitude also follows a similar trend. For the PLDR-LLM that exhibits reasoning, the steady state is very robust to text generation even with stochastic sampling methods used in generating these sample runs. When normalized RMSE by mean magnitude is compared for all models in table \ref{tablermsecriticalall}, this observation is valid for all models capable of reasoning. Normalized RMSE by mean magnitude defines a simple order parameter, which goes to near zero for models that are trained near-criticality.

PLDRv51-SOC-110M-5 was trained with more data and the normalized RMSE by mean magnitude values are zero for $\tA_{\textbf{P}}$ and $\tG_{LM}$. They are orders of magnitude smaller than other models for $\tA$ and $\tA_{LM}$. More high quality data during training allows the model to learn higher degree of symmetries. This results in deductive outputs to be more invariant to the input samples at inference. The floating-point precision of the model parameters is also a limiting factor for the sensitivity of deductive outputs to perturbation.

PLDRv51-SOC-110M-5 exhibits larger standard deviation for all deductive outputs compared to other models that reason. Mean and standard deviation of deductive outputs for all models can be found in the appendix.

\begin{table}
	\caption{Mean ($\mu$) and standard deviation ($\sigma$) of deductive outputs of models PLDRv51-SOC-110M-4 (near-critical) and SUB-SOC-110M-2 (sub-critical) at runs 1, 2 and Cached.}
	\label{tablemeanstd}
	\centering
	\sisetup{table-format = 1.4e1, table-alignment-mode = format}
	\resizebox{0.85\textwidth}{!}{
		\begin{tabular}{c l *{4}{S!{\qquad}}}
			\toprule[1.2pt]
			&  & \multicolumn{2}{c}{PLDRv51-SOC-110M-4}  & \multicolumn{2}{c}{SUB-SOC-110M-2} \\
			\cmidrule(lr){3-4} \cmidrule(lr){5-6}
			Runs &  & {$\mu$} & {$\sigma$} & {$\mu$} & {$\sigma$} \\
			\midrule[1.2pt]
			\multirow{4}{*}{1} & $\tA$ & -1.6349E-03 & 9.2655E-02 & -6.6022E-03 & 1.5189E-01 \\
			\cmidrule(lr){2-6}
			& $\tA_{LM}$ & 1.6954E-04 & 6.8940E-04 & 5.4364E-04 & 6.7867E-03 \\
			\cmidrule(lr){2-6}
			& $\tA_{\textbf{P}}$ & 1.2834E+00 & 2.5361E+00 & 1.1240E+00 & 2.6259E+00 \\
			\cmidrule(lr){2-6}
			& $\tG_{LM}$ & -2.3352E-03 & 2.2214E+00 & -1.8509E-02 & 1.9082E+00 \\
			\midrule
			\multirow{4}{*}{2} & $\tA$ & -1.6349E-03 & 9.2655E-02 & -6.6019E-03 & 1.5190E-01 \\
			\cmidrule(lr){2-6}
			& $\tA_{LM}$ & 1.6954E-04 & 6.8940E-04 & 5.4439E-04 & 6.7761E-03 \\
			\cmidrule(lr){2-6}
			& $\tA_{\textbf{P}}$ & 1.2834E+00 & 2.5361E+00 & 1.1237E+00 & 2.5036E+00 \\
			\cmidrule(lr){2-6}
			& $\tG_{LM}$ & -2.3352E-03 & 2.2214E+00 & -1.8685E-02 & 1.8766E+00 \\
			\midrule
			\multirow{4}{*}{Cached} & $\tA$ & -1.6349E-03 & 9.2655E-02 & -6.9380E-03 & 1.5245E-01 \\
			\cmidrule(lr){2-6}
			& $\tA_{LM}$ & 1.6954E-04 & 6.8940E-04 & 5.1420E-04 & 6.9008E-03 \\
			\cmidrule(lr){2-6}
			& $\tA_{\textbf{P}}$ & 1.2834E+00 & 2.5361E+00 & 1.1224E+00 & 2.6059E+00 \\
			\cmidrule(lr){2-6}
			& $\tG_{LM}$ & -2.3352E-03 & 2.2214E+00 & -1.9625E-02 & 1.8974E+00 \\
			\bottomrule[1.2pt]
		\end{tabular}
}
\end{table}

\begin{table}
	\caption{RMSE and normalized RMSE by mean magnitude of deductive outputs of models PLDRv51-SOC-110M-4 (near-critical) and SUB-SOC-110M-2 (sub-critical) between runs 1, 2 and runs 1, Cached.}
	\label{tablermsecritical}
	\centering
	\sisetup{table-format = 1.4e1, table-alignment-mode = format}
	\resizebox{\textwidth}{!}{
		\begin{tabular}{cl*{4}{S!{\qquad}}}
			\toprule[1.2pt]
			&  & {$RMSE_{12}$} & {$RMSE_{1C}$} & {$RMSE_{12}/|\mu_{1}|$} & {$RMSE_{1C}/|\mu_{C}|$} \\
			\midrule[1.2pt]
			\multirow{4}{*}{PLDRv51-SOC-110M-4} & $\tA$ & 2.1092E-09 & 2.1410E-09 & 1.2901E-06 & 1.3096E-06 \\
			\cmidrule{2-6}
			& $\tA_{LM}$ & 3.0347E-11 & 3.1737E-11 & 1.7899E-07 & 1.8719E-07 \\
			\cmidrule{2-6}
			& $\tA_{\textbf{P}}$ & 4.7302E-08 & 4.6144E-08 & 3.6856E-08 & 3.5953E-08 \\
			\cmidrule{2-6}
			& $\tG_{LM}$ & 2.1934E-08 & 2.1485E-08 & 9.3926E-06 & 9.2006E-06 \\
			\midrule
			\multirow{4}{*}{SUB-SOC-110M-2} & $\tA$ & 4.7210E-02 & 1.0205E-01 & 7.1506E+00 & 1.4708E+01 \\
			\cmidrule{2-6}
			& $\tA_{LM}$ & 1.7820E-03 & 3.9981E-03 & 3.2778E+00 & 7.7753E+00 \\
			\cmidrule{2-6}
			& $\tA_{\textbf{P}}$ & 1.5788E+00 & 2.2192E+00 & 1.4047E+00 & 1.9771E+00 \\
			\cmidrule{2-6}
			& $\tG_{LM}$ & 6.4775E-01 & 9.2130E-01 & 3.4997E+01 & 4.6945E+01 \\
			\bottomrule[1.2pt]
		\end{tabular}
	}
\end{table}

\begin{table}
	\caption{Normalized RMSE by mean magnitude of deductive outputs of all pretrained models between runs 1 and Cached.}
	\label{tablermsecriticalall}
	\centering
	\sisetup{table-format = 1.4e1, table-alignment-mode = format}
	\resizebox{\textwidth}{!}{
	\begin{tabular}{lc*{4}{S!{\qquad}}}
		\toprule[1.2pt]
		&  & \multicolumn{4}{c}{$RMSE_{1C}/|\mu_{C}|$} \\
		\cmidrule{3-6} 
		&  & {$\tA$} & {$\tA_{LM}$} & {$\tA_{\textbf{P}}$} & {$\tG_{LM}$} \\ 
		\midrule[1.2pt]
		\multirow{5}{*}{NEAR CRITICAL} & PLDRv51-SOC-110M-1 & 8.3069E-03 & 6.8777E-04 & 9.5313E-04 & 2.1867E-02 \\ 
		\cmidrule{2-6}
		& PLDRv51-SOC-110M-2 & 4.3525E-06 & 8.5975E-07 & 1.0032E-07 & 1.2528E-05 \\ 
		\cmidrule{2-6}
		& PLDRv51-SOC-110M-3 & 1.7730E-04 & 8.1658E-07 & 3.2427E-06 & 5.1342E-04 \\ 
		\cmidrule{2-6}
		& PLDRv51-SOC-110M-4 & 1.3096E-06 & 1.8719E-07 & 3.5953E-08 & 9.2006E-06 \\ 
		\cmidrule{2-6}
		& PLDRv51-SOC-110M-5 & 4.1660E-11 & 1.2508E-12 & 0 & 0 \\ 
		\midrule
		\multirow{2}{*}{SUB CRITICAL} & SUB-SOC-110M-1 & 3.0323E+00 & 3.6632E+00 & 3.3531E-01 & 1.5834E+01 \\
		\cmidrule{2-6}
		& SUB-SOC-110M-2 & 1.4708E+01 & 7.7753E+00 & 1.9771E+00 & 4.6945E+01 \\ 
		\midrule
		\multirow{3}{*}{ABLATION} & ABL-SOC-110M-1 & 1.0042E+01 & 1.4944E+01 & 8.5672E+00 & 1.8421E+02 \\ 
		\cmidrule{2-6}
		& ABL-SOC-110M-2 & 5.0448E+00 & 3.2431E+00 & 1.0874E+00 & 4.4422E+01 \\ 
		\cmidrule{2-6}
		& ABL-SOC-110M-3 & 3.7942E+03 & 7.7942E+00 & 6.7517E-01 & 1.8211E+01 \\ 
		\bottomrule[1.2pt]
	\end{tabular}
}
\end{table}

\subsection{Distribution of Deductive Output Values}

Global distribution of deductive outputs show high sparsity for $\tA$, $\tA_{LM}$ and $\tG_{LM}$ whereas $\tA_{\textbf{P}}$ is centered around $\sim1$ (Fig. \ref{figmainhistograms}). They also show very consistent distribution characteristics for each criticality condition. The near-critical model and sub-critical models show clear differences in the distributions of $\tA$ where it exhibits a wider distribution range of values for the sub-critical model. Heat maps of $\tA$ in the last decoder layer for single head averaged over all samples also show that the repetitive, uniform characteristic is lost for the sub-critical model. 

The distributions for PLDRv51-SOC-110M-5 exhibit similar characteristics around zero compared to other models that reason, but also show outliers in the distribution. While normalized RMSE for $\tA$ is more sensitive to perturbation, we prefer to compare normalized RMSE for $\tG_{LM}$ since it is the final tensor before attention is derived. However, normalized RMSE is non-zero for $\tA$ and zero for $\tG_{LM}$, the former can be used as an order parameter to compare PLDRv51-SOC-110M-5.
 
Distribution plots for all models can be found in the appendix for reference.

\begin{figure}[!htb]
	\centering
	\begin{subfigure}[b]{0.45\textwidth}
		\centering
		\includegraphics[width=1\textwidth]{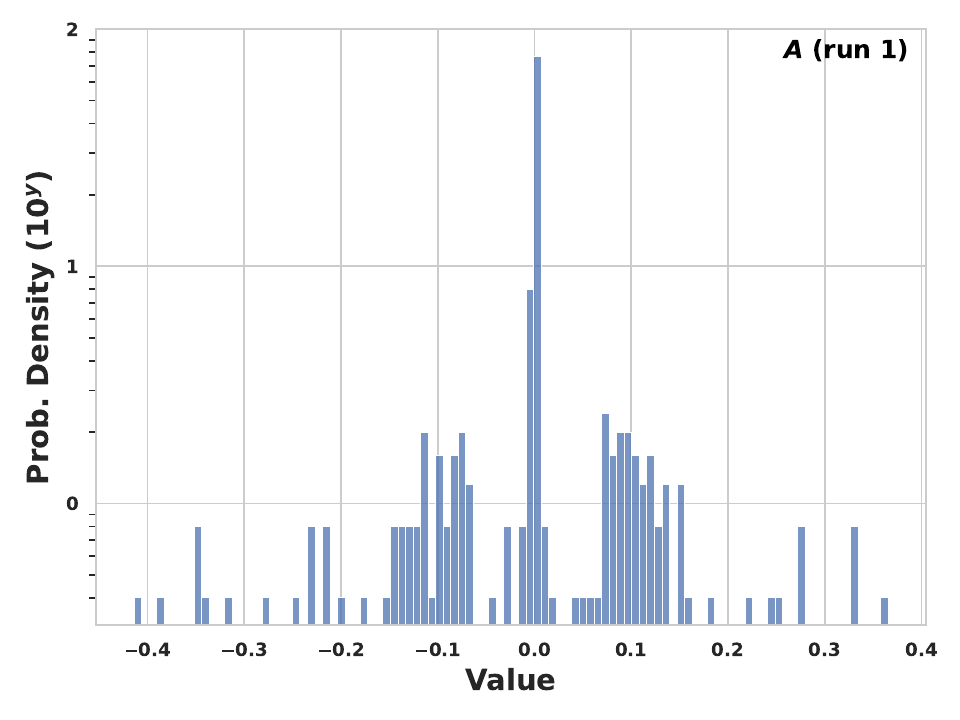}
		\caption{PLDRv51-SOC-110M-4}
		\label{figmainAPLDRv51SOC110M4}
	\end{subfigure}
	\hfill
	\begin{subfigure}[b]{0.45\textwidth}
		\centering
		\includegraphics[width=1\textwidth]{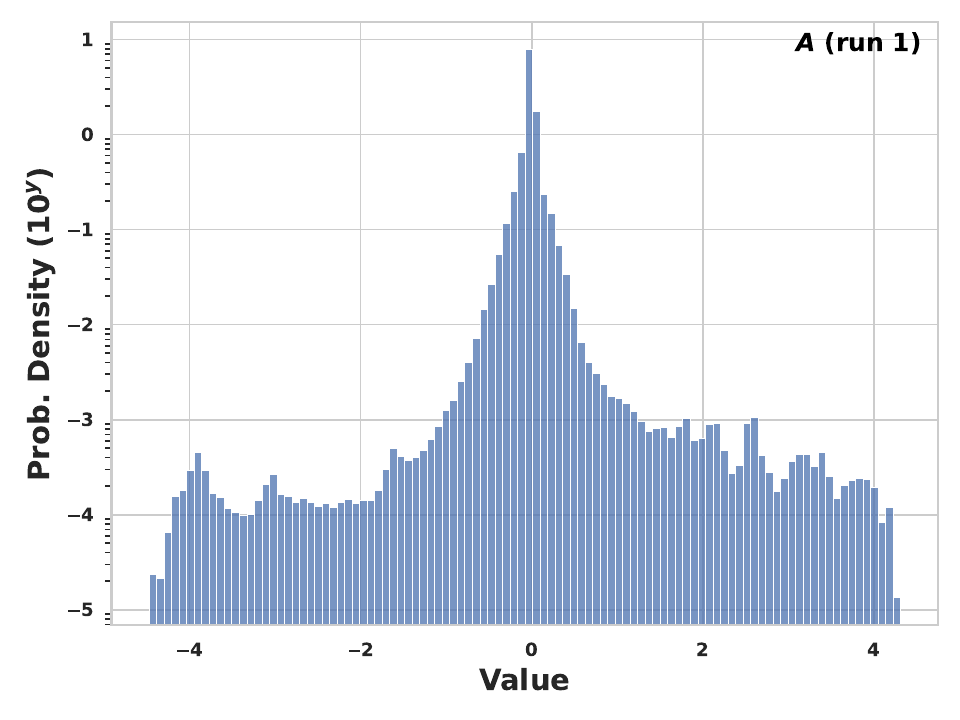}
		\caption{SUB-SOC-110M-2}
		\label{figmainASUBSOC110M2}
	\end{subfigure}
	\hfill
	\begin{subfigure}[b]{0.45\textwidth}
		\centering
		\includegraphics[width=1\textwidth]{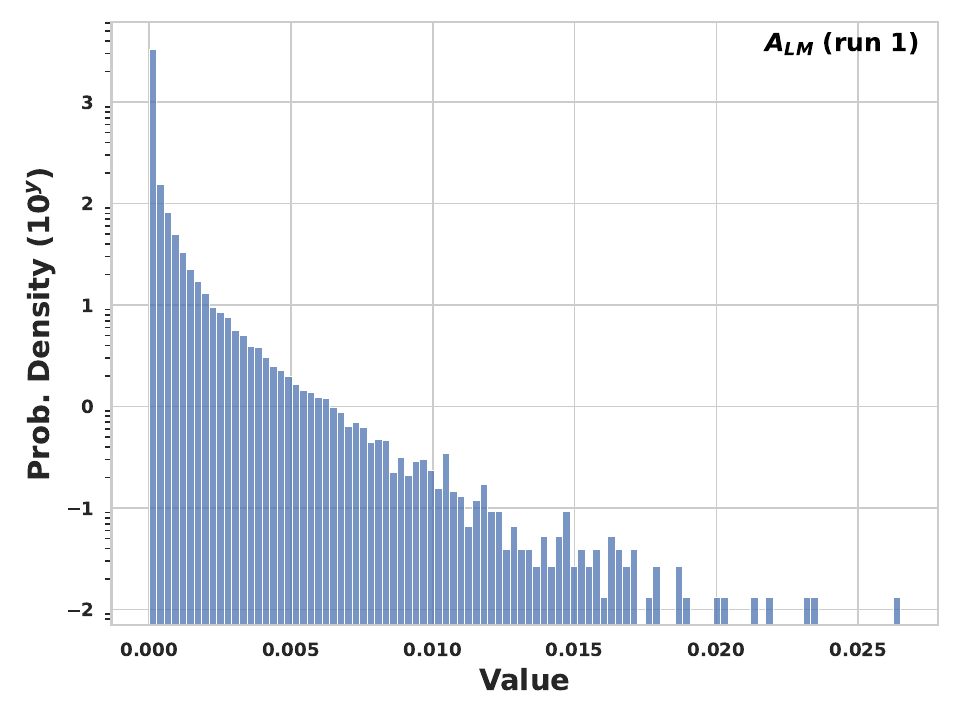}
		\caption{PLDRv51-SOC-110M-4}
		\label{figmainALMPLDRv51SOC110M4}
	\end{subfigure}
	\hfill
	\begin{subfigure}[b]{0.45\textwidth}
		\centering
		\includegraphics[width=1\textwidth]{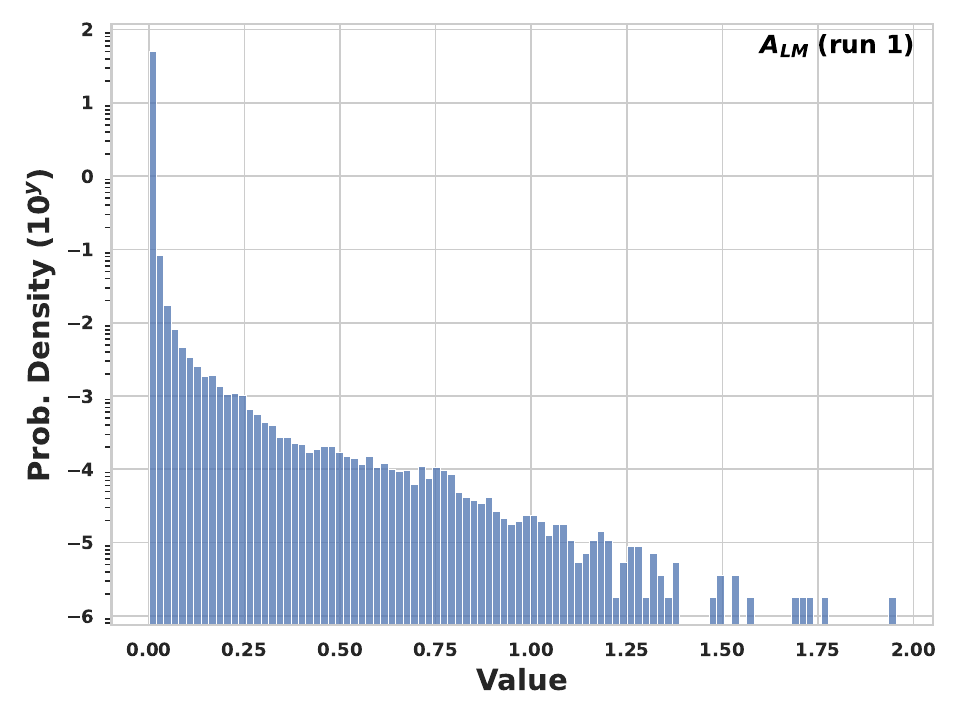}
		\caption{SUB-SOC-110M-2}
		\label{figmainALMSUBSOC110M2}
	\end{subfigure}
		\begin{subfigure}[b]{0.45\textwidth}
		\centering
		\includegraphics[width=1\textwidth]{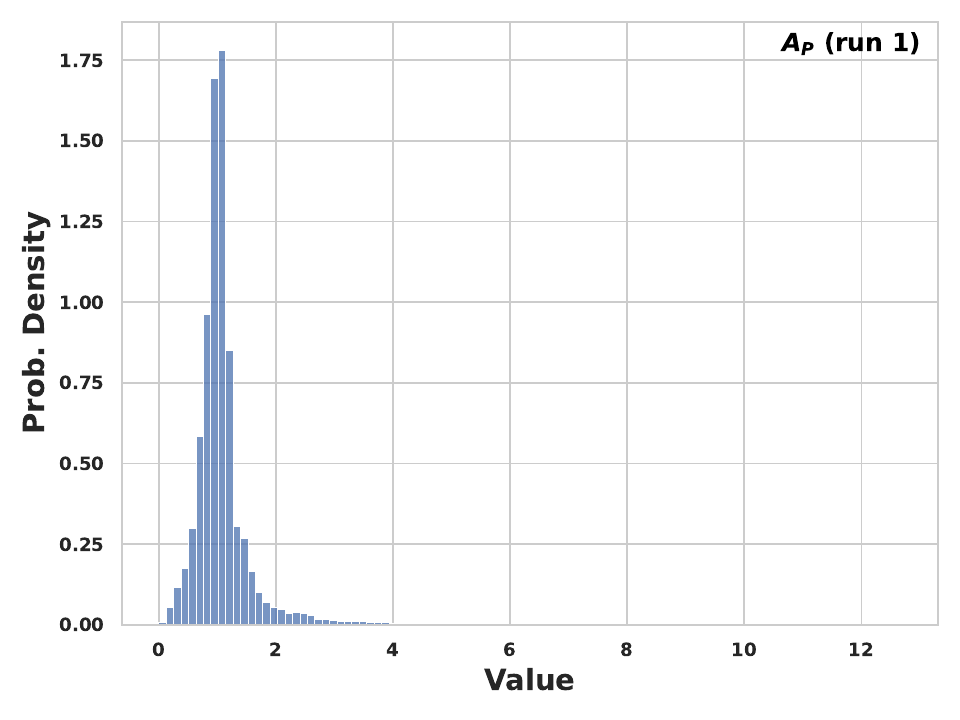}
		\caption{PLDRv51-SOC-110M-4}
		\label{figmainAPPLDRv51SOC110M4}
	\end{subfigure}
	\hfill
	\begin{subfigure}[b]{0.45\textwidth}
		\centering
		\includegraphics[width=1\textwidth]{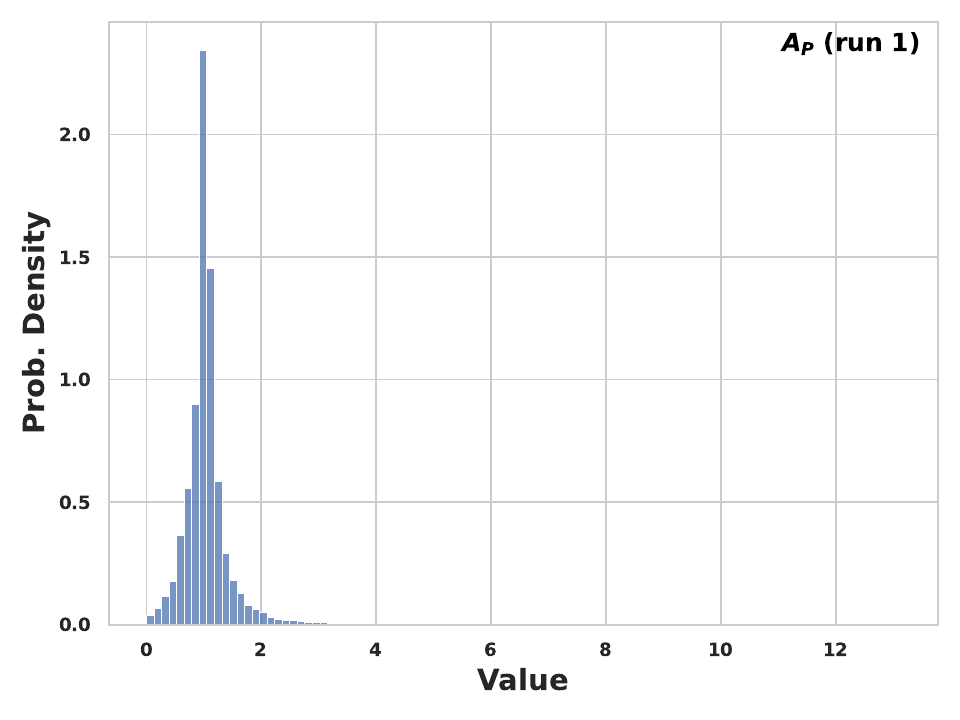}
		\caption{SUB-SOC-110M-2}
		\label{figmainAPSUBSOC110M2}
	\end{subfigure}
	\caption{Deductive output probability density distributions for all values in a model for PLDRv51-SOC-110M-4 and SUB-SOC-110M-2 binned in 100 buckets. The $\tA_{\textbf{P}}$ and $\tG_{LM}$ were plotted up to $\pm5\sigma$ for easier visibility of main distribution characteristics. $\tA$ and $\tA_{LM}$ distributions were plotted as log-linear.}
\end{figure}

\begin{figure}[!htb]
	\ContinuedFloat
	\centering
	\begin{subfigure}[b]{0.45\textwidth}
		\centering
		\includegraphics[width=1\textwidth]{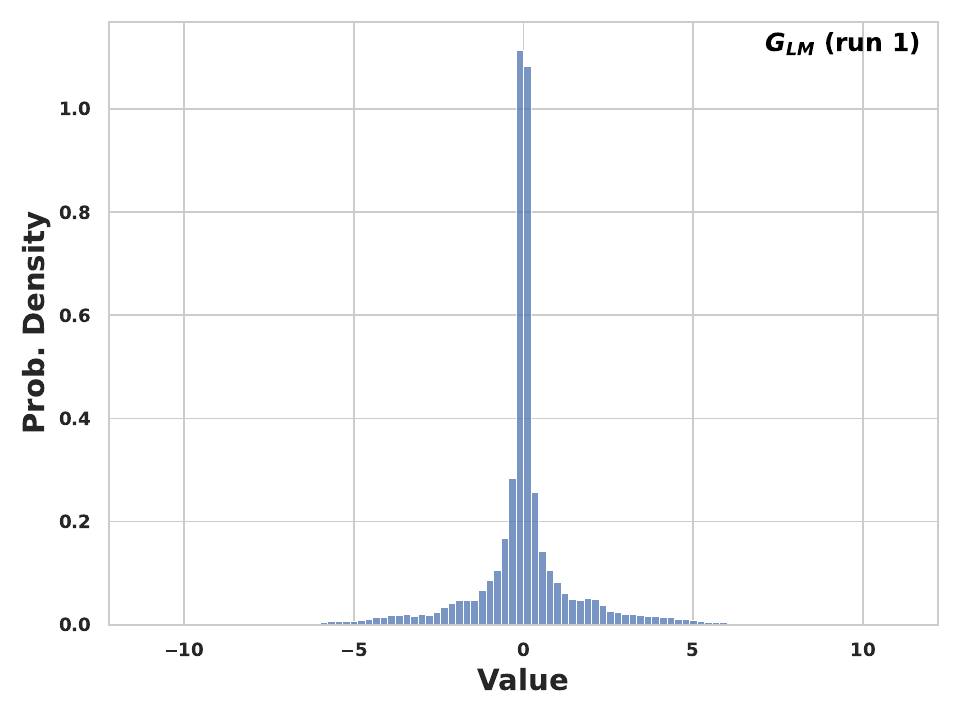}
		\caption{PLDRv51-SOC-110M-4}
		\label{figmainGLMPLDRv51SOC110M4}
	\end{subfigure}
	\hfill
	\begin{subfigure}[b]{0.45\textwidth}
		\centering
		\includegraphics[width=1\textwidth]{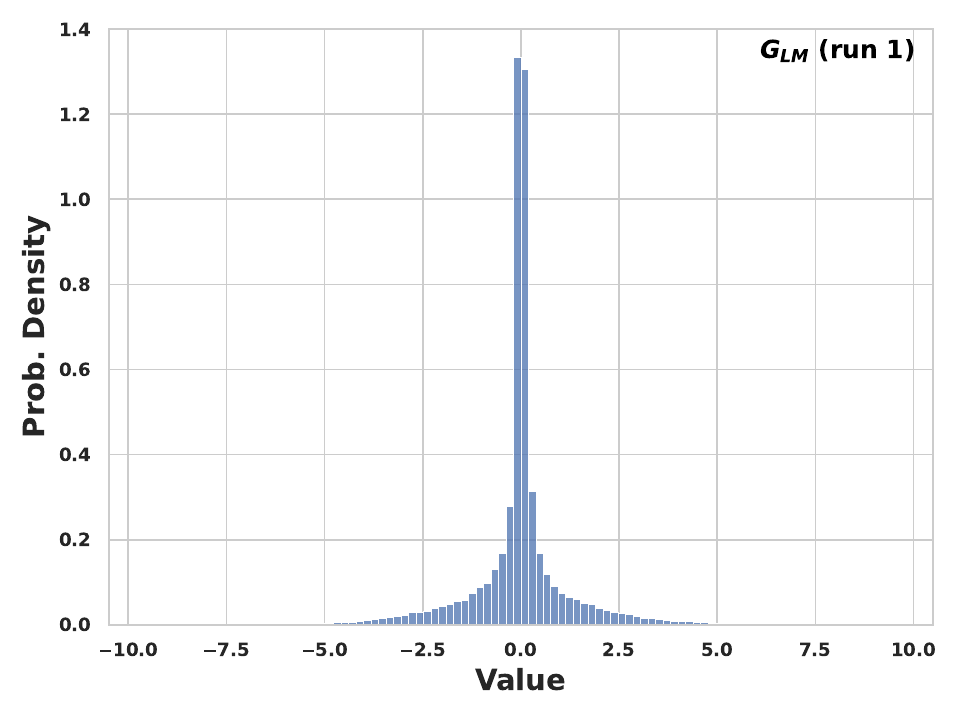}
		\caption{SUB-SOC-110M-2}
		\label{figmainGLMSUBSOC110M2}
	\end{subfigure}
		\hfill
	\begin{subfigure}[b]{0.45\textwidth}
		\centering
		\includegraphics[width=1\textwidth]{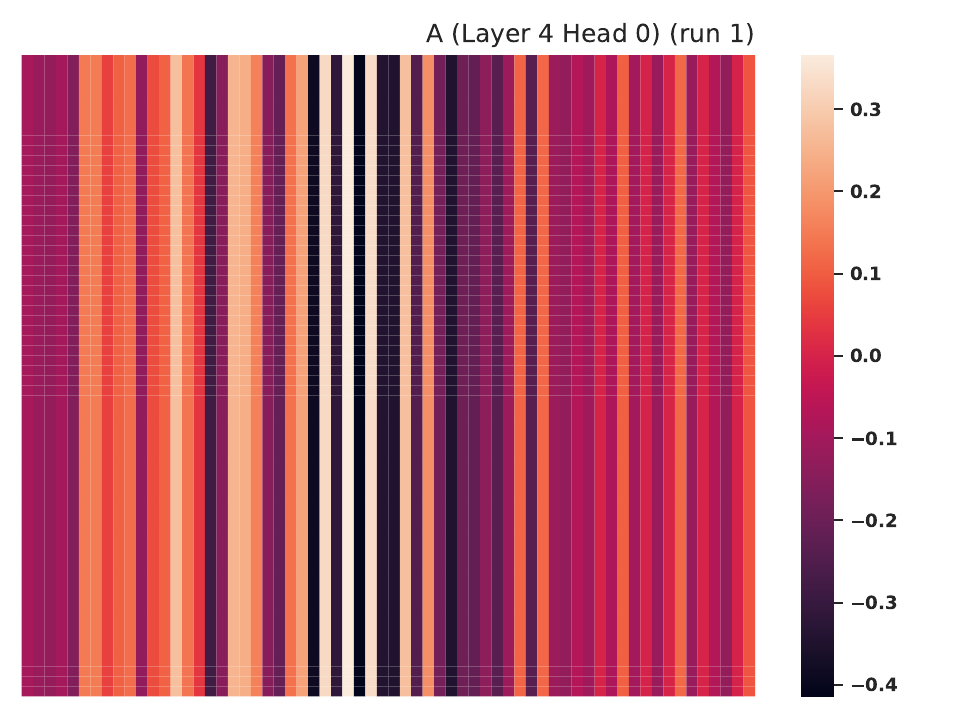}
		\caption{PLDRv51-SOC-110M-4}
		\label{figmainAhmPLDRv51SOC110M4}
	\end{subfigure}
	\hfill
	\begin{subfigure}[b]{0.45\textwidth}
		\centering
		\includegraphics[width=1\textwidth]{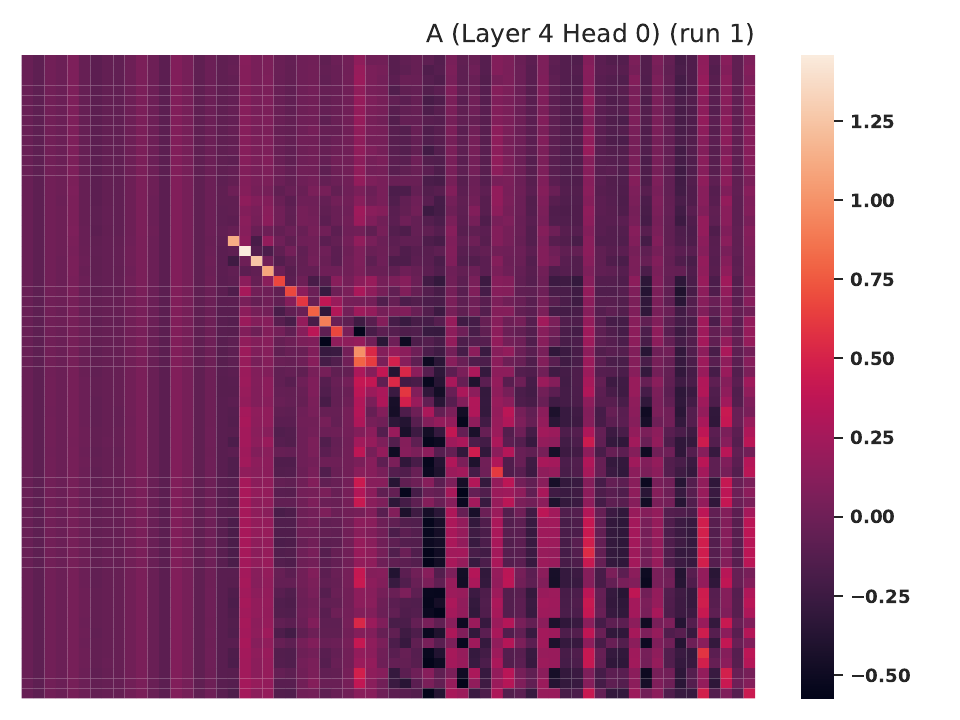}
		\caption{SUB-SOC-110M-2}
		\label{figmainAhmSUBSOC110M2}
	\end{subfigure}
	\caption{(Cont.) Deductive output probability density distributions for all values in a model for PLDRv51-SOC-110M-4 and SUB-SOC-110M-2 binned in 100 buckets. The $\tA_{\textbf{P}}$ and $\tG_{LM}$ were plotted up to $\pm5\sigma$ for easier visibility of main distribution characteristics. $\tA$ and $\tA_{LM}$ distributions were plotted as log-linear. The heatmaps of $\tA$ were averaged over all samples for same layer and head.}
	\label{figmainhistograms}
\end{figure}

\subsection{Comparison of Order Parameter and Benchmark Scores}

Benchmark scores are commonly used to evaluate LLMs for their reasoning and comprehension capability. However, these benchmarks typically perform better and more distinctive for LLMs with large parameter sizes and that are trained with a large amount of tokens scaled by increase in their size. We show in table \ref{tablebenchmarks} that order parameter is a precise indicator of reasoning and comprehension capabilities that can match the benchmark score evaluation. Since it is an intrinsic characteristic of the model, order parameter is independent of any benchmark dataset. The more order parameter of a model trained at near-criticality is close to zero, its average benchmark scores are higher. This trend still holds when the model is trained with more data, as it is the case for PLDRv51-SOC-110M-5. On the other hand, the order parameter is much larger for sub-critical models and their benchmark scores suffer. This comparison is further evidence that PLDR-LLM exhibits reasoning at self-organized criticality and is a self-contained, complete model architecture that can be evaluated independent of any input or benchmark dataset. As a result, small size  PLDR-LLMs can be evaluated as good as large size PLDR-LLMs that would require significant compute resources. 

PLDRv51-SOC-110M-5 shows the smallest order parameter values and highest average benchmark scores among all PLDR-LLMs pretrained. It has better average scores as well compared to a SDPA-LLM model with similar model size from literature, GPT-Neo-125M\footnote{https://huggingface.co/EleutherAI/gpt-neo-125m} \citep{gptneo, Gao2020pile}. It achieves this performance with modest compute requirements at a model parameter size of 110M and was trained over $\sim$41B tokens from RefinedWeb dataset.

\begin{table}
	\caption{Zero-shot benchmark evaluation results for all PLDR-LLMs trained at different criticality conditions. HS: Hellaswag, OBQA: OpenBookQA, WG: WinoGrande, TQA: TruthfulQA. For the benchmarks that showed unusually high scores for sub-critical and ablation models that can be easily confirmed as not capable of reasoning, a tiered average approach is used. Avg. 1 is cumulative average with ARC-c included, Avg. 2 is cumulative average with TQA also included. GPT-Neo-125M benchmark scores are from \citep{Gokden2025pldrllmkvgcache} that uses the same methodology for evaluation.}
	\label{tablebenchmarks}
	\centering
	\sisetup{table-format = 1.4e1, table-alignment-mode = format,  text-series-to-math = true , propagate-math-font = true}
	\resizebox{\textwidth}{!}{
		\begin{tabular}{c *{11}{S[table-alignment-mode = marker]} c}
			\toprule[1.2pt]
			& & & & & & & & & & & & {$RMSE_{1C}/|\mu_{C}|$} \\
			\cmidrule{13-13}
			& {ARC-e} & {HS} & {OBQA} & {PIQA} & {SIQA} & {WG} & {Avg. 0} & {ARC-c} & {Avg. 1} & {TQA} & {Avg. 2} & {\makecell{$\tG_{LM}$ \\[2pt] $\tA$}} \\
			\midrule[1.2pt]
			PLDRv51-SOC-110M-1 & 36.15 & 28.65 & 27.00 & 61.43 & 41.66 & 50.83 & 40.95 & 22.95 & 38.38 & 44.59 & 39.16 &  {\makecell{\num{2.1867E-02} \\ \num{8.3069E-03}}} \\
			\midrule
			PLDRv51-SOC-110M-2 & 36.87 & 29.27 & 26.20 & 62.79 & 41.91 & 52.41 & $\mathbf{\num{41.57}}$ & 20.99 & $\mathbf{\num{38.63}}$ & 44.33 & $\mathbf{\num{39.35}}$ & {\makecell{$\mathbf{\num{1.2528E-05}}$ \\ $\mathbf{\num{4.3525E-06}}$}} \\
			\midrule
			PLDRv51-SOC-110M-3 & 36.99 & 28.99 & 27.20 & 62.08 & 42.17 & 49.80 & 41.21 & 21.76 & 38.43 & 44.17 & 39.14 & {\makecell{\num{5.1342E-04} \\ \num{1.7730E-04}}} \\
			\midrule
			PLDRv51-SOC-110M-4 & 36.83 & 29.13 & 29.20 & 61.37 & 41.66 & 50.59 & $\mathbf{\num{41.46}}$ & 21.50 & $\mathbf{\num{38.61}}$ & 44.33 & $\mathbf{\num{39.33}}$ & {\makecell{$\mathbf{\num{9.2006E-06}}$ \\ $\mathbf{\num{1.3096E-06}}$}} \\
			\midrule
			PLDRv51-SOC-110M-5 & 38.55 & 30.55 & 29.80 & 63.98 & 43.09 & 49.72 & $\mathbf{\num{42.62}}$ & 22.95 & $\mathbf{\num{39.81}}$ & 43.00 & $\mathbf{\num{40.21}}$ & {\makecell{$\mathbf{\num{0}}$ \\ $\mathbf{\num{4.1660E-11}}$}} \\
			\midrule
			SUB-SOC-110M-1 & 24.07 & 26.03 & 26.00 & 51.63 & 36.80 & 48.38 & 35.49 & 26.79 & 34.24 & 48.91 & 36.08 & {\makecell{\num{1.5834E+01} \\ \num{3.0323E+00}}} \\
			\midrule
			SUB-SOC-110M-2 & 25.38 & 25.93 & 25.20 & 51.25 & 35.57 & 49.49 & 35.47 & 25.26 & 34.01 & 48.07 & 35.77 & {\makecell{\num{4.6945E+01} \\ \num{1.4708E+01}}} \\
			\midrule[1.2pt]
			ABL-SOC-110M-1 & 33.00 & 26.88 & 27.60 & 55.50 & 40.79 & 51.30 & 39.18 & 22.44 & 36.79 & 44.66 & 37.77 & {\makecell{\num{1.8421E+02} \\ \num{1.0042E+01}}} \\
			\midrule
			ABL-SOC-110M-2 & 24.87 & 26.03 & 26.60 & 51.85 & 36.80 & 50.04 & 36.03 & 24.91 & 34.44 & 50.41 & 36.44 & {\makecell{\num{4.4422E+01} \\ \num{5.0448E+00}}} \\
			\midrule
			ABL-SOC-110M-3 & 25.46 & 25.45 & 28.20 & 52.45 & 36.54 & 49.80 & 36.32 & 28.16 & 35.15 & 48.42 & 36.81 & {\makecell{\num{1.8211E+01} \\ \num{3.7942E+03}}} \\
			\midrule[1.2pt]
			GPT-Neo-125M & 39.39 & 30.40 & 26.20 & 62.46 & 42.07 & 50.91 & 41.91 & 23.12 & 39.22 & 45.58 & 40.02 & NA \\
			\bottomrule[1.2pt]
		\end{tabular}
	}
\end{table}

\section{Ablation Studies}

\begin{figure}[!htb]
	\centering
	\begin{subfigure}[b]{0.48\textwidth}
		\centering
		\includegraphics[width=1\textwidth]{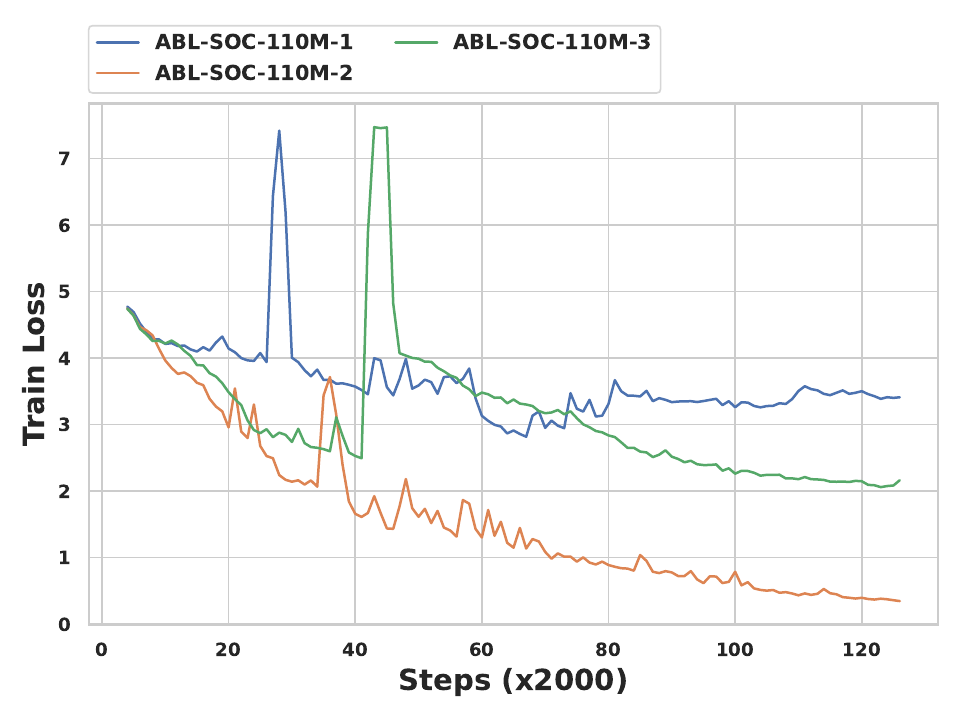}
		\caption{}
		\label{abllosscurves}
	\end{subfigure}
	\hfill
	\begin{subfigure}[b]{0.48\textwidth}
		\centering
		\includegraphics[width=1\textwidth]{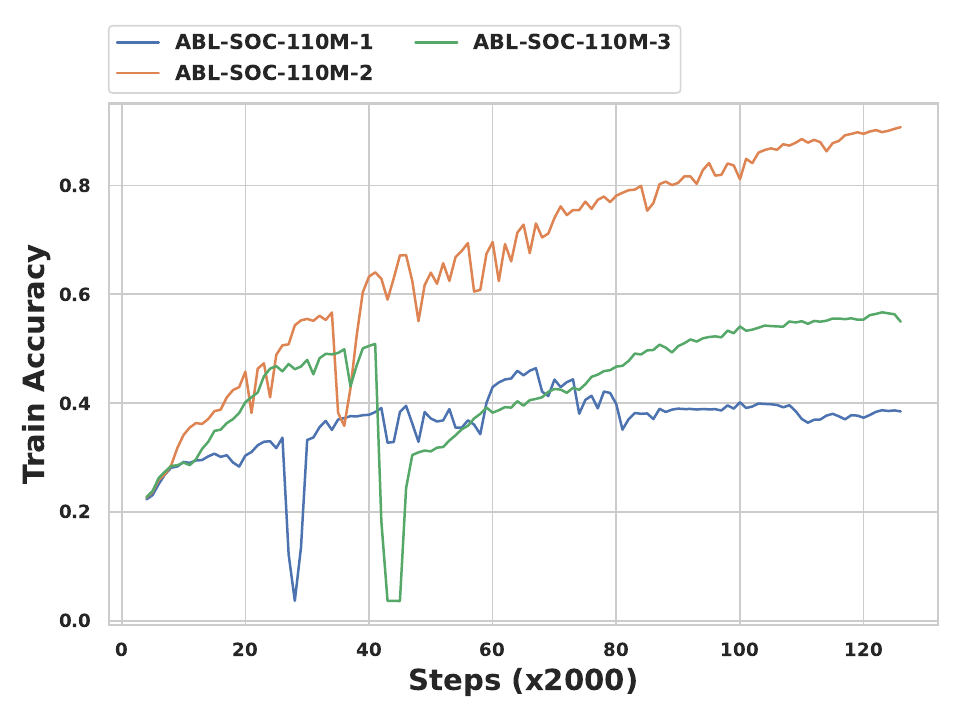}
		\caption{}
		\label{ablacccurves}
	\end{subfigure}
	\caption{Train loss (a) and accuracy (b) curves for the PLDR-LLMs pretrained as ablation study near-critical and sub-critical conditions and exhibiting dragon king events. Each data point was captured as a running average of 2000 steps.}
	\label{abllossacccurves}
\end{figure}

We performed ablation studies with a focus on warm-up step count and maximum learning rate pairs that eventually result in dragon king events (Table \ref{tablehyperparameters}). Dragon kings often arise in early stages of pretraining when learning rate is still high. The loss and accuracy curves for models that exhibit dragon kings are shown in fig. \ref{abllossacccurves}. They appear in the loss curve as a sharp peak characterized with a very high loss value. Dragon king also appears as a significant drop in accuracy curve at the same time. These effects were initially dismissed as failures in model optimization due to exploding gradients encountering steep walls on the loss manifold that throw the model off track \citep{Pascanu2013recurrent} and were not studied in detail. In the self-organized criticality picture, these extreme events can happen under specific and known conditions and are usually due to self-amplifying mechanisms that are caused by an imbalance in the driving impulse and dissipation force \citep{Mikaberidze2023DK,Sornette2012DK}. It is an indication of deviation from power law behaviour at criticality. In our experiments, we observed dragon king behavior both at the near-critical and sub-critical regions. A model can be driven into sub-critical region even the maximum learning rate is large enough but the warm-up step count does not balance the dissipation rate, which is the case for ABL-SOC-110M-2. Super-critical condition can briefly hold and dragon kings can appear in such otherwise sub-critical models. As it was mentioned before, we observed that it becomes more difficult to find a proper warm-up step count for low maximum learning rates. However, with specific warm-up or annealing schedules, we may expect to maintain criticality even at lower learning rates. This would require a more detailed study of driving and dissipation mechanisms of PLDR-LLMs during pretraining. After dragon king appears, the model still tries to follow near-critical behaviour as in ABL-SOC-110M-1 that  has a maximum learning rate of $1\times 10^{-3}$, but reasoning capability fails as indicated by large normalized RMSE by mean magnitude values in table \ref{tablermsecriticalall}. This observation is also supported by reduced benchmark scores in table \ref{tablebenchmarks}. 

Dragon kings are catastrophic events that appear in critical physical systems in nature and could be very informative of the course of progress in such systems. For PLDR-LLM and other critical systems, dragon kings are predictable events and can be avoided. The histogram distributions of deductive outputs and additional statistics of the ablation models are provided in the appendix.

\section{Discussion}

The observation that PLDR-LLM attains reasoning at self-organized criticality and the deductive outputs establish an input independent metastable steady state can help to explain several characteristics of LLMs in general. The improvement of LLM reasoning with scaling of model size and training data \citep{Kaplan2020scaling,Hoffmann2022chinchilla} also increases the number of parameters for the steady state which can capture more details of the higher dimensional symmetries of the pretraining data. Since steady state of deductive outputs is unchanged for any input in a model with good reasoning capabilities, scaling the model size becomes a very effective way to improve the reasoning capabilities. It is also easier to understand under the steady state condition, why models with deeper layers often outperform the wider models. The possible combinations of interactions between hidden states increases exponentially in deeper models, which improve the representation capacity of a system with fixed entries at steady state.

The self-organized criticality needs slowly driven updates with infinitesimal length to achieve and maintain the steady state at criticality. Therefore, slower changes of parameters are favored to maintain the delicate balance forming the steady state in each PLGA layer. For example, SwiGLU \citep{Dauphin2017GLU,Shazeer2020swiglu} allows a linear path for gradients while maintaining non-linearity and rotary positional embedding \citep{Su2021roformer} rotate the embedding vectors instead of changing their magnitude by addition of a positional embedding value. We expect that improvements to PLDR-LLM that help maintain the steady state at criticality easier can help develop better performing and more robust language models.

PLDR-LLM and its connection to self-organized criticality provide significant insights into how reasoning can arise from a language model. Self-organized criticality is observed in many physical systems, some of which have rich, reliable data to observe and others are very scarce in data. Moreover, these processes may belong to same universality classes. One exciting area of research on PLDR-LLM's capacity to generalize at criticality can be whether a model trained on a high resource field such as NLP can improve prediction capability on a field that is scarce of observations in a controlled environment such as understanding earthquake dynamics. In traditional machine learning, this is possible to a degree with methods such as fine-tuning or transfer learning that improves prediction of the inductive output. However, the model is still a black box with little understanding on how the low resource system is generalized. Access to the details of steady state dynamics through deductive outputs of PLDR-LLMs can further improve our understanding in complex physical systems that are otherwise hard to observe and thus generalize.

One interesting observation is that the steady state condition only perturbs negligibly under stochastic sampling conditions such as nucleus sampling used in this study. In SDPA-LLMs which do not have access to any of the deductive output dynamics, the steady state is a hidden variable (unknown to the observer) and the nucleus sampling of the inductive output is probabilistic. Query and key vectors define attention, fully-connected layers transform the attention at each layer and a choice is made for the next token based on the logits of the probabilities in the final layer. Unlike an SDPA-LLM, deductive outputs of PLDR-LLM at criticality do not get modified with any choice for the next token. This  situation suggests that the deductive outputs learn the representations of scaling functions, universality classes and renormalization groups such that for any input it is unchanged. This mechanism of generalization is fundamentally different than how the model learns for its fully-connected layers with activation functions.

It is hypothesized that brain also operates at criticality. Some evidence has been shown to support this hypothesis, though more experiments are  needed. As a computational model that is a fully controlled environment, PLDR-LLM can be an important tool to understand the dynamics of human brain further. Possible applications of PLDR-LLM as a laboratory test vehicle can be to understand and treat cognitive disorders, and uncover the differences in reasoning by LLMs compared to the brain.

\section{Conclusion}

We extended the investigations into unique characteristics of PLDR-LLM architecture observed since its inception from the perspective of self-organized criticality. We show that PLDR-LLM exhibits behaviour similar to second-order phase transitions during training. The reasoning capability is achieved when PLDR-LLMs are pretrained at criticality. The deductive outputs of PLDR-LLM attain a metastable steady-state at inference, and this suggests that PLGA learns the representations equivalent to scaling functions, universality classes and renormalization groups from the pretraining data. This type of generalization through a steady state condition allows the reasoning capability of the model to be quantified very precisely. We define an order parameter from the normalized RMSE by mean magnitude of global deductive output values from separate runs of generic input prompts with stochastic sampling. We show that this order parameter approaches closer to zero for models that also show higher benchmark scores for reasoning and comprehension. Our results indicate that PLDR-LLM is a self-contained model whose characteristics can be completely determined from the deductive outputs. This observation opens the path for PLDR-LLMs to be studied in every aspect without the need for training very large size models that require extensive computing resources for training and inference. The curated benchmark datasets are insightful for evaluating inductive output of a model, but not needed to quantify reasoning in the self-organized criticality perspective. We believe our understanding of PLDR-LLM and its reasoning dynamics presented here will lead to better analytical characterization of LLMs in general, the ability to better understand complex physical systems in other domains with low resources for observation, and how reasoning manifests in the brain.

\section*{Acknowledgments}

I am grateful to my parents for their support and patience. This research was conducted independently without support from a grant or corporation.

\bibliographystyle{unsrtnat}
\bibliography{pldrllm-soc-references}

\clearpage
\appendix

%tables for mean, std and RMSE
% Appendix file for PLDR SOC paper

\section*{Appendix}

\section{Mean and Std Dev for Pretrained PLDR-LLM Deductive Outputs}
\begin{table}[h]
	\caption{Mean ($\mu$) values of $\tA$ and $\tA_{LM}$ of all pretrained models at runs 1, 2 and Cached.}
	\label{tableallmodelsmean}
	\centering
	\sisetup{table-format = 1.4e1, table-alignment-mode = format}
	\resizebox{0.85\textwidth}{!}{
	\begin{tabular}{c c *{6}{S!{\quad}}}
		\toprule[1.2pt]
		&  & \multicolumn{3}{c}{$\tA$}  & \multicolumn{3}{c}{$\tA_{LM}$}  \\
		\cmidrule(lr){3-5} \cmidrule(lr){6-8}
		 & & {$\mu_{1}$} & {$\mu_{2}$} & {$\mu_{Cached}$} & {$\mu_{1}$} & {$\mu_{2}$} & {$\mu_{Cached}$}  \\
		\midrule[1.2pt]
		\multirow{5}{*}{\shortstack{NEAR\\CRITICAL}} & PLDRv51-SOC-110M-1 & 2.1041E-03 & 2.1041E-03 & 2.1041E-03 & 6.0489E-04 & 6.0489E-04 & 6.0489E-04  \\
		\cmidrule(lr){2-8}
		& PLDRv51-SOC-110M-2 & -9.8446E-04 & -9.8446E-04 & -9.8446E-04 & 1.6681E-04 & 1.6681E-04 & 1.6681E-04  \\
		\cmidrule(lr){2-8}
		& PLDRv51-SOC-110M-3 & 5.9329E-03 & 5.9329E-03 & 5.9329E-03 & 2.4056E-04 & 2.4056E-04 & 2.4056E-04  \\
		\cmidrule(lr){2-8}
		& PLDRv51-SOC-110M-4 & -1.6349E-03 & -1.6349E-03 & -1.6349E-03 & 1.6954E-04 & 1.6954E-04 & 1.6954E-04 \\
		\cmidrule(lr){2-8}
		& PLDRv51-SOC-110M-5 & -1.3555E-02 & -1.3555E-02 & -1.3555E-02 & 2.1645E-02 & 2.1645E-02 & 2.1645E-02 \\
		\midrule
		\multirow{2}{*}{\shortstack{SUB\\CRITICAL}} & SUB-SOC-110M-1 & 9.7475E-03 & 9.7477E-03 & 9.6593E-03 & 1.0627E-03 & 1.0635E-03 & 1.0392E-03 \\
		\cmidrule(lr){2-8}
		& SUB-SOC-110M-2 & -6.6022E-03 & -6.6019E-03 & -6.9380E-03 & 5.4364E-04 & 5.4439E-04 & 5.1420E-04 \\
		\midrule
		\multirow{3}{*}{ABLATION} & ABL-SOC-110M-1 & -1.0951E-02 & -1.0947E-02 & -1.1575E-02 & 7.9532E-03 & 7.9589E-03 & 8.0124E-03  \\
		\cmidrule(lr){2-8}
		& ABL-SOC-110M-2 & -1.3195E-02 & -1.3183E-02 & -1.3912E-02 & 3.9689E-03 & 3.9656E-03 & 3.9123E-03 \\
		\cmidrule(lr){2-8}
		& ABL-SOC-110M-3 & 1.5216E-04 & 1.5748E-04 & 1.8376E-05 & 3.3616E-04 & 3.3625E-04 & 3.3062E-04 \\
		\bottomrule[1.2pt]
	\end{tabular}
	}
	
	\caption{Mean ($\mu$) values of $\tA_{\textbf{P}}$ and $\tG_{LM}$ of all pretrained models at runs 1, 2 and Cached.}
	\label{tableallmodelsmean1}
	\centering
	\sisetup{table-format = 1.4e1, table-alignment-mode = format}
	\resizebox{0.85\textwidth}{!}{
		\begin{tabular}{c c *{6}{S!{\quad}}}
			\toprule[1.2pt]
			&  &  \multicolumn{3}{c}{$\tA_{\textbf{P}}$} & \multicolumn{3}{c}{$\tG_{LM}$}  \\
			\cmidrule(lr){3-5} \cmidrule(lr){6-8}
			 & & {$\mu_{1}$} & {$\mu_{2}$} & {$\mu_{Cached}$} & {$\mu_{1}$} & {$\mu_{2}$} & {$\mu_{Cached}$}  \\
			\midrule[1.2pt]
			\multirow{5}{*}{\shortstack{NEAR\\CRITICAL}} & PLDRv51-SOC-110M-1 & 1.2127E+00 & 1.2127E+00 & 1.2127E+00 & -9.2139E-03 & -9.2140E-03 & -9.2138E-03 \\
			\cmidrule(lr){2-8}
			 & PLDRv51-SOC-110M-2 & 1.2592E+00 & 1.2592E+00 & 1.2592E+00 & -4.9210E-03 & -4.9210E-03 & -4.9210E-03 \\
			\cmidrule(lr){2-8}
			& PLDRv51-SOC-110M-3 & 1.1864E+00 & 1.1864E+00 & 1.1864E+00 & 3.9397E-03 & 3.9397E-03 & 3.9397E-03 \\
			\cmidrule(lr){2-8}
			& PLDRv51-SOC-110M-4 &  1.2834E+00 & 1.2834E+00 & 1.2834E+00 & -2.3352E-03 & -2.3352E-03 & -2.3352E-03 \\
			\cmidrule(lr){2-8}
			& PLDRv51-SOC-110M-5 &  1.5005E+00 & 1.5005E+00 & 1.5005E+00 & -9.3573E-03 & -9.3573E-03 & -9.3573E-03 \\
			\midrule
			\multirow{2}{*}{\shortstack{SUB\\CRITICAL}} & SUB-SOC-110M-1 & 1.1717E+00 & 1.1717E+00 & 1.1719E+00 & 1.1005E-02 & 1.1015E-02 & 1.0833E-02 \\
			\cmidrule(lr){2-8}
				& SUB-SOC-110M-2 & 1.1240E+00 & 1.1237E+00 & 1.1224E+00 & -1.8509E-02 & -1.8685E-02 & -1.9625E-02 \\
			\midrule
			\multirow{3}{*}{ABLATION} & ABL-SOC-110M-1 & 3.9976E+00 & 3.9919E+00 & 4.0119E+00 & -4.8490E-02 & -4.8331E-02 & -4.6665E-02 \\
			\cmidrule(lr){2-8}
			& ABL-SOC-110M-2 & 1.1032E+00 & 1.1031E+00 & 1.1042E+00 & -1.3235E-02 & -1.3150E-02 & -1.2866E-02 \\
			\cmidrule(lr){2-8}
			& ABL-SOC-110M-3 & 1.2135E+00 & 1.2135E+00 & 1.2153E+00 & -2.2128E-02 & -2.2210E-02 & -2.2251E-02 \\
			\bottomrule[1.2pt]
		\end{tabular}
	}
	
	\caption{Standard deviation ($\sigma$) values of $\tA$ and $\tA_{LM}$ of all pretrained models at runs 1, 2 and Cached.}
	\label{tableallmodelssigma}
	\centering
	\sisetup{table-format = 1.4e1, table-alignment-mode = format}
	\resizebox{0.85\textwidth}{!}{
		\begin{tabular}{c c *{6}{S!{\quad}}}
			\toprule[1.2pt]
			&  & \multicolumn{3}{c}{$\tA$}  & \multicolumn{3}{c}{$\tA_{LM}$} \\
			\cmidrule(lr){3-5} \cmidrule(lr){6-8}
			& & {$\sigma_{1}$} & {$\sigma_{2}$} & {$\sigma_{Cached}$} & {$\sigma_{1}$} & {$\sigma_{2}$} & {$\sigma_{Cached}$}  \\
			\midrule[1.2pt]
			\multirow{5}{*}{\shortstack{NEAR\\CRITICAL}} & PLDRv51-SOC-110M-1 & 5.7461E-02 & 5.7461E-02 & 5.7461E-02 & 4.7671E-03 & 4.7671E-03 & 4.7671E-03  \\
			\cmidrule(lr){2-8}
			& PLDRv51-SOC-110M-2 & 8.1209E-02 & 8.1209E-02 & 8.1209E-02 & 1.4358E-03 & 1.4358E-03 & 1.4358E-03  \\
			\cmidrule(lr){2-8}
			& PLDRv51-SOC-110M-3 & 1.0658E-01 & 1.0658E-01 & 1.0658E-01 & 1.7124E-03 & 1.7124E-03 & 1.7124E-03 \\
			\cmidrule(lr){2-8}
			& PLDRv51-SOC-110M-4 & 9.2655E-02 & 9.2655E-02 & 9.2655E-02 & 6.8940E-04 & 6.8940E-04 & 6.8940E-04 \\
			\cmidrule(lr){2-8}
			& PLDRv51-SOC-110M-5 & 2.4249E-01 & 2.4249E-01 & 2.4249E-01 & 1.1569E+01 & 1.1569E+01 & 1.1569E+01 \\
			\midrule
			\multirow{2}{*}{\shortstack{SUB\\CRITICAL}} & SUB-SOC-110M-1 & 7.0436E-02 & 7.0462E-02 & 6.9683E-02 & 1.0331E-02 & 1.0328E-02 & 1.0204E-02 \\
			\cmidrule(lr){2-8}
			& SUB-SOC-110M-2 & 1.5189E-01 & 1.5190E-01 & 1.5245E-01 & 6.7867E-03 & 6.7761E-03 & 6.9008E-03 \\
			\midrule
			\multirow{3}{*}{ABLATION} & ABL-SOC-110M-1 & 2.2558E-01 & 2.2574E-01 & 2.1577E-01 & 8.1948E-01 & 8.2113E-01 & 8.1877E-01 \\
			\cmidrule(lr){2-8}
			& ABL-SOC-110M-2 & 1.4103E-01 & 1.4106E-01 & 1.4308E-01 & 3.6208E-02 & 3.6157E-02 & 3.6568E-02 \\
			\cmidrule(lr){2-8}
			& ABL-SOC-110M-3 & 8.8414E-02 & 8.8436E-02 & 8.9070E-02 & 3.3533E-03 & 3.3539E-03 & 3.3990E-03 \\
			\bottomrule[1.2pt]
		\end{tabular}
}

	\caption{Standard deviation ($\sigma$) values of $\tA_{\textbf{P}}$ and $\tG_{LM}$ of all pretrained models at runs 1, 2 and Cached.}
	\label{tableallmodelssigma1}
	\centering
	\sisetup{table-format = 1.4e1, table-alignment-mode = format}
	\resizebox{0.85\textwidth}{!}{
	\begin{tabular}{c c *{6}{S!{\quad}}}
		\toprule[1.2pt]
		&  &  \multicolumn{3}{c}{$\tA_{\textbf{P}}$} & \multicolumn{3}{c}{$\tG_{LM}$}  \\
		\cmidrule(lr){3-5} \cmidrule(lr){6-8}
		& & {$\sigma_{1}$} & {$\sigma_{2}$} & {$\sigma_{Cached}$} & {$\sigma_{1}$} & {$\sigma_{2}$} & {$\sigma_{Cached}$}  \\
		\midrule[1.2pt]
		\multirow{5}{*}{\shortstack{NEAR\\CRITICAL}} & PLDRv51-SOC-110M-1 &  2.3447E+00 & 2.3447E+00 & 2.3447E+00 & 1.9497E+00 & 1.9497E+00 & 1.9497E+00 \\
		\cmidrule(lr){2-8}
		& PLDRv51-SOC-110M-2 & 2.6758E+00 & 2.6758E+00 & 2.6758E+00 & 2.2230E+00 & 2.2230E+00 & 2.2230E+00 \\
		\cmidrule(lr){2-8}
		& PLDRv51-SOC-110M-3 & 2.0695E+00 & 2.0695E+00 & 2.0695E+00 & 1.8114E+00 & 1.8114E+00 & 1.8114E+00 \\
		\cmidrule(lr){2-8}
		& PLDRv51-SOC-110M-4 & 2.5361E+00 & 2.5361E+00 & 2.5361E+00 & 2.2214E+00 & 2.2214E+00 & 2.2214E+00 \\
		\cmidrule(lr){2-8}
		& PLDRv51-SOC-110M-5 & 5.8997E+00 & 5.8997E+00 & 5.8997E+00 & 3.3020E+00 & 3.3020E+00 & 3.3020E+00 \\
		\midrule
		\multirow{2}{*}{\shortstack{SUB\\CRITICAL}} & SUB-SOC-110M-1 & 2.0562E+00 & 2.0516E+00 & 2.0576E+00 & 1.9203E+00 & 1.9190E+00 & 1.9200E+00 \\
		\cmidrule(lr){2-8}
		& SUB-SOC-110M-2 & 2.6259E+00 & 2.5036E+00 & 2.6059E+00 & 1.9082E+00 & 1.8766E+00 & 1.8974E+00 \\
		\midrule
		\multirow{3}{*}{ABLATION} & ABL-SOC-110M-1 & 1.4098E+02 & 1.4015E+02 & 1.4443E+02 & 3.6662E+01 & 3.6502E+01 & 3.7342E+01 \\
		\cmidrule(lr){2-8}
		& ABL-SOC-110M-2 & 1.6127E+00 & 1.5825E+00 & 1.9342E+00 & 1.8571E+00 & 1.8524E+00 & 1.9366E+00 \\
		\cmidrule(lr){2-8}
		& ABL-SOC-110M-3 & 2.7258E+00 & 2.7307E+00 & 2.8533E+00 & 2.2626E+00 & 2.2646E+00 & 2.3034E+00 \\
		\bottomrule[1.2pt]
	\end{tabular}
}
	
\end{table}
\newpage
\section{RMSE and Normalized RMSE for Pretrained PLDR-LLM Deductive Outputs}
\begin{table}[h]
	\caption{RMSE values of deductive outputs of all pretrained models at runs 1, 2 and Cached.}
	\label{tableallmodelsRMSE}
	\centering
	\sisetup{table-format = 1.4e1, table-alignment-mode = format}
	\resizebox{\textwidth}{!}{
		\begin{tabular}{cc*{8}{S!{\quad}}}
			\toprule[1.2pt]
			&  & \multicolumn{2}{c}{$\tA$} & \multicolumn{2}{c}{$\tA_{LM}$} & \multicolumn{2}{c}{$\tA_{\textbf{P}}$} & \multicolumn{2}{c}{$\tG_{LM}$} \\
			\cmidrule(lr){3-4} \cmidrule(lr){5-6} \cmidrule(lr){7-8} \cmidrule(lr){9-10}
			& &  {$RMSE_{12}$} & {$RMSE_{1C}$} & {$RMSE_{12}$} & {$RMSE_{1C}$} & {$RMSE_{12}$} & {$RMSE_{1C}$} & {$RMSE_{12}$} & {$RMSE_{1C}$} \\
			\midrule[1.2pt]
			\multirow{5}{*}{\shortstack{NEAR\\CRITICAL}} & PLDRv51-SOC-110M-1 & 1.7823E-05 & 1.7479E-05 & 4.5307E-07 & 4.1603E-07 & 7.9668E-04 & 1.1559E-03 & 1.6772E-04 & 2.0148E-04 \\
			\cmidrule(lr){2-10}
			& PLDRv51-SOC-110M-2 & 4.1713E-09 & 4.2848E-09 & 1.3083E-10 & 1.4341E-10 & 1.2817E-07 & 1.2633E-07 & 6.2531E-08 & 6.1651E-08 \\
			\cmidrule(lr){2-10}
			& PLDRv51-SOC-110M-3 & 8.1415E-07 & 1.0519E-06 & 1.5399E-10 & 1.9643E-10 & 3.4293E-06 & 3.8472E-06 & 1.8074E-06 & 2.0227E-06 \\
			\cmidrule(lr){2-10}
			& PLDRv51-SOC-110M-4 & 2.1092E-09 & 2.1410E-09 & 3.0347E-11 & 3.1737E-11 & 4.7302E-08 & 4.6144E-08 & 2.1934E-08 & 2.1485E-08 \\
			\cmidrule(lr){2-10}
			& PLDRv51-SOC-110M-5 & 4.7764E-13 & 5.6471E-13 & 2.3071E-14 & 2.7074E-14 & 0 & 0 & 0 & 0 \\
			\midrule
			\multirow{2}{*}{\shortstack{SUB\\CRITICAL}} & SUB-SOC-110M-1 &  2.8131E-02 & 2.9290E-02 & 3.8573E-03 & 3.8069E-03 & 2.1001E-01 & 3.9297E-01 & 9.6239E-02 & 1.7153E-01 \\
			\cmidrule(lr){2-10}
			& SUB-SOC-110M-2 & 4.7210E-02 & 1.0205E-01 & 1.7820E-03 & 3.9981E-03 & 1.5788E+00 & 2.2192E+00 & 6.4775E-01 & 9.2130E-01 \\
			\midrule
			\multirow{3}{*}{ABLATION} & ABL-SOC-110M-1 & 8.5889E-02 & 1.1624E-01 & 9.8798E-02 & 1.1974E-01 & 2.4309E+01 & 3.4371E+01 & 6.4290E+00 & 8.5961E+00 \\
			\cmidrule(lr){2-10}
			& ABL-SOC-110M-2 & 5.9923E-02 & 7.0185E-02 & 1.1403E-02 & 1.2688E-02 & 3.9351E-01 & 1.2008E+00 & 1.6470E-01 & 5.7152E-01 \\
			\cmidrule(lr){2-10}
			& ABL-SOC-110M-3 & 3.8996E-02 & 6.9723E-02 & 1.3239E-03 & 2.5769E-03 & 3.1748E-01 & 8.2056E-01 & 1.5110E-01 & 4.0520E-01 \\
			\bottomrule[1.2pt]
		\end{tabular}		
	}
	
	\caption{Normalized RMSE by mean magnitude of deductive outputs of all pretrained models at runs 1, 2 and Cached.}
	\label{tableallmodelsNRMSE}
	\centering
	\sisetup{table-format = 1.4e1, table-alignment-mode = format}
	\resizebox{\textwidth}{!}{
	\begin{tabular}{cc*{8}{S!{\quad}}}
		\toprule[1.2pt]
		&  & \multicolumn{2}{c}{$\tA$}& \multicolumn{2}{c}{$\tA_{LM}$} & \multicolumn{2}{c}{$\tA_{\textbf{P}}$} & \multicolumn{2}{c}{$\tG_{LM}$}  \\
		\cmidrule(lr){3-4} \cmidrule(lr){5-6} \cmidrule(lr){7-8} \cmidrule(lr){9-10}
		&  & {$RMSE_{12}/|\mu_{1}|$} & {$RMSE_{1C}/|\mu_{C}|$} & {$RMSE_{12}/|\mu_{1}|$} & {$RMSE_{1C}/|\mu_{C}|$} & {$RMSE_{12}/|\mu_{1}|$} & {$RMSE_{1C}/|\mu_{C}|$} & {$RMSE_{12}/|\mu_{1}|$} & {$RMSE_{1C}/|\mu_{C}|$} \\
		\midrule[1.2pt]
		\multirow{5}{*}{\shortstack{NEAR\\CRITICAL}} & PLDRv51-SOC-110M-1 & 8.4709E-03 & 8.3069E-03 & 7.4902E-04 & 6.8777E-04 & 6.5694E-04 & 9.5313E-04 & 1.8203E-02 & 2.1867E-02 \\
		\cmidrule(lr){2-10}
		& PLDRv51-SOC-110M-2 & 4.2371E-06 & 4.3525E-06 & 7.8430E-07 & 8.5975E-07 & 1.0178E-07 & 1.0032E-07 & 1.2707E-05 & 1.2528E-05 \\
		\cmidrule(lr){2-10}
		& PLDRv51-SOC-110M-3 & 1.3723E-04 & 1.7730E-04 & 6.4012E-07 & 8.1658E-07 & 2.8905E-06 & 3.2427E-06 & 4.5876E-04 & 5.1342E-04 \\
		\cmidrule(lr){2-10}
		& PLDRv51-SOC-110M-4 & 1.2901E-06 & 1.3096E-06 & 1.7899E-07 & 1.8719E-07 & 3.6856E-08 & 3.5953E-08 & 9.3926E-06 & 9.2006E-06 \\
		\cmidrule(lr){2-10}
		& PLDRv51-SOC-110M-5 & 3.5237E-11 & 4.1660E-11 & 1.0659E-12 & 1.2508E-12 & 0 & 0 & 0 & 0 \\
		\midrule
		\multirow{2}{*}{\shortstack{SUB\\CRITICAL}} & SUB-SOC-110M-1 & 2.8859E+00 & 3.0323E+00 & 3.6298E+00 & 3.6632E+00 & 1.7923E-01 & 3.3531E-01 & 8.7449E+00 & 1.5834E+01 \\
		\cmidrule(lr){2-10}
		& SUB-SOC-110M-2 & 7.1506E+00 & 1.4708E+01 & 3.2778E+00 & 7.7753E+00 & 1.4047E+00 & 1.9771E+00 & 3.4997E+01 & 4.6945E+01 \\
		\midrule
		\multirow{3}{*}{ABLATION} & ABL-SOC-110M-1 & 7.8433E+00 & 1.0042E+01 & 1.2422E+01 & 1.4944E+01 & 6.0810E+00 & 8.5672E+00 & 1.3258E+02 & 1.8421E+02 \\
		\cmidrule(lr){2-10}
		& ABL-SOC-110M-2 & 4.5414E+00 & 5.0448E+00 & 2.8730E+00 & 3.2431E+00 & 3.5674E-01 & 1.0874E+00 & 1.2445E+01 & 4.4422E+01 \\
		\cmidrule(lr){2-10}
		& ABL-SOC-110M-3 & 2.5628E+02 & 3.7942E+03 & 3.9382E+00 & 7.7942E+00 & 2.6163E-01 & 6.7517E-01 & 6.8287E+00 & 1.8211E+01 \\
		\bottomrule[1.2pt]
	\end{tabular}
	}
\end{table}

%figures for deductive outputs
\section{Global Density Distributions and Heatmaps for Deductive Outputs of PLDR-LLMs}

%%%%%%%%%%%%%%%%%%%%%%%%%%%%%%%%%%%%%%%%%%%%%%%%%%%%%%%%%%%%%%%%%%%%%%%%%%%%%%%%%%%%%%%%%%%%%%%%%%%%%%%%%%
%A DISTRIBUTIONS
%%%%%%%%%%%%%%%%%%%%%%%%%%%%%%%%%%%%%%%%%%%%%%%%%%%%%%%%%%%%%%%%%%%%%%%%%%%%%%%%%%%%%%%%%%%%%%%%%%%%%%%%%%

\begin{figure}[!htb]
	\centering
	\begin{subfigure}[b]{0.45\textwidth}
		\centering
		\includegraphics[width=1\textwidth]{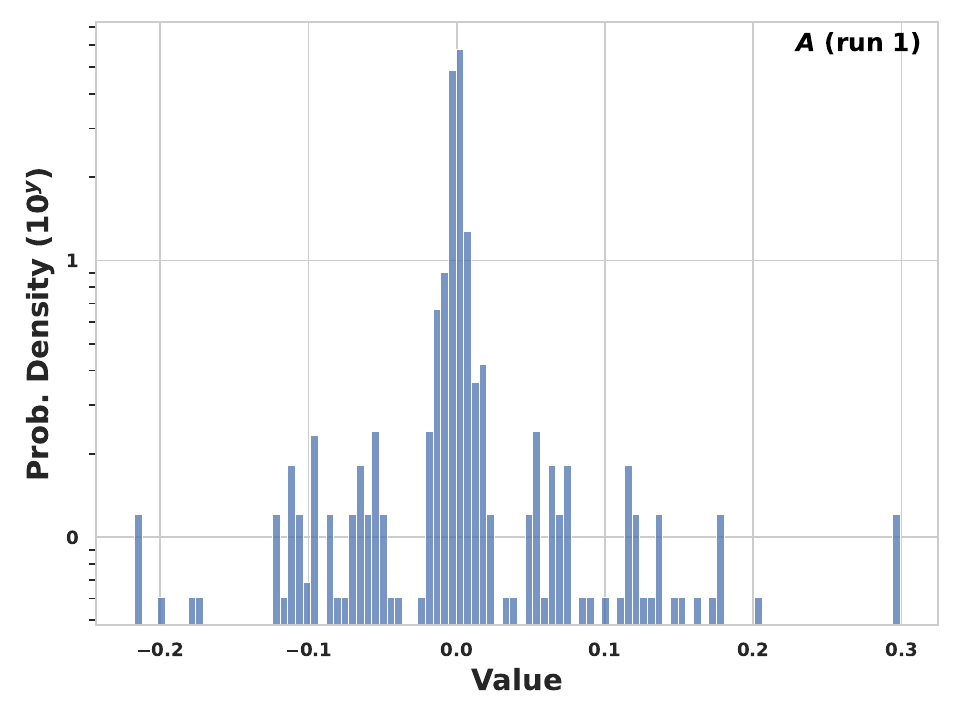}
		\caption{PLDRv51-SOC-110M-1}
		\label{PLDRv51SOC110M1}
	\end{subfigure}
	\hfill
	\begin{subfigure}[b]{0.45\textwidth}
		\centering
		\includegraphics[width=1\textwidth]{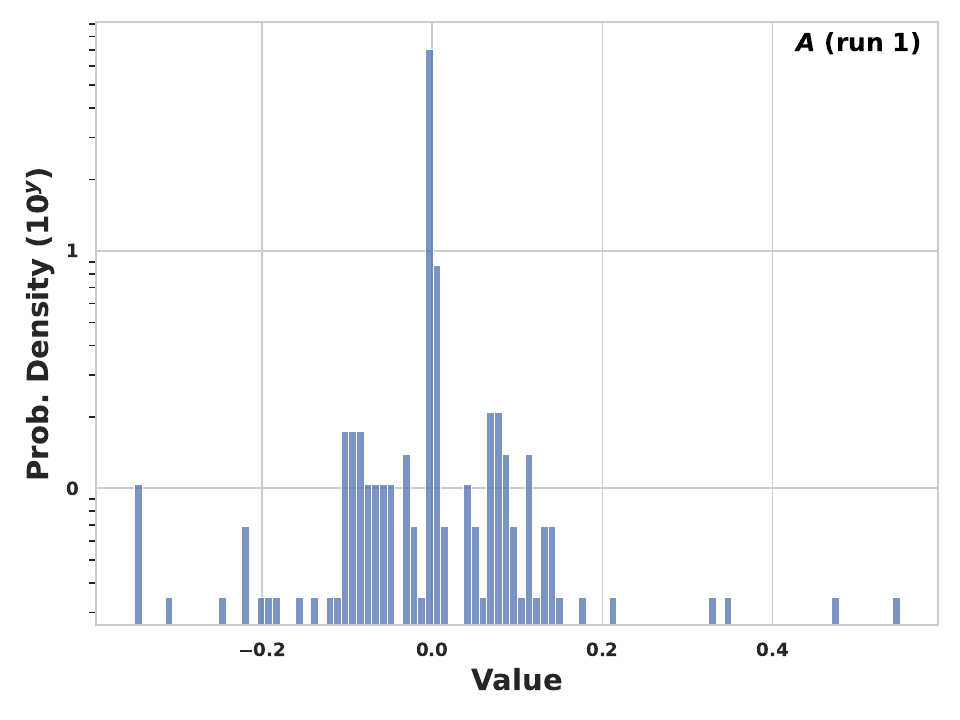}
		\caption{PLDRv51-SOC-110M-2}
		\label{PLDRv51SOC110M2}
	\end{subfigure}
	\hfill
	\begin{subfigure}[b]{0.45\textwidth}
		\centering
		\includegraphics[width=1\textwidth]{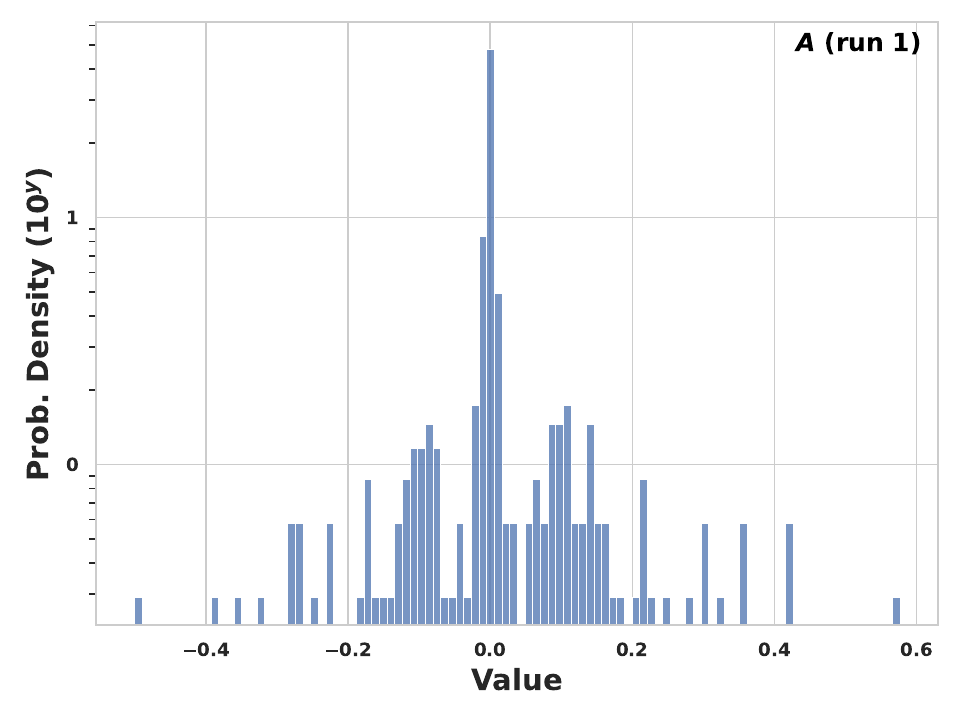}
		\caption{PLDRv51-SOC-110M-3}
		\label{PLDRv51SOC110M3}
	\end{subfigure}
	\hfill
	\begin{subfigure}[b]{0.45\textwidth}
		\centering
		\includegraphics[width=1\textwidth]{A_histplot_ylog_xlin_PLDRv51-SOC-110M-4_run1}
		\caption{PLDRv51-SOC-110M-4}
		\label{PLDRv51SOC110M4}
	\end{subfigure}
	\hfill
	\begin{subfigure}[b]{0.45\textwidth}
	\centering
	\includegraphics[width=1\textwidth]{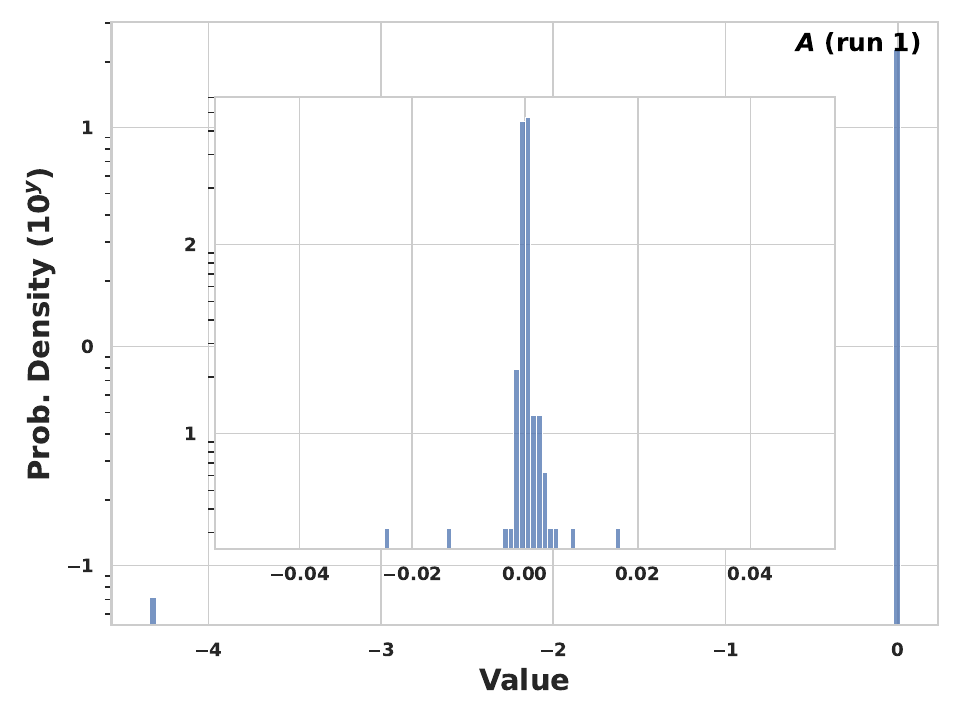}
	\caption{PLDRv51-SOC-110M-5}
	\label{PLDRv51SOC110M5}
	\end{subfigure}
	\hfill
	\begin{subfigure}[b]{0.45\textwidth}
		\centering
		\includegraphics[width=1\textwidth]{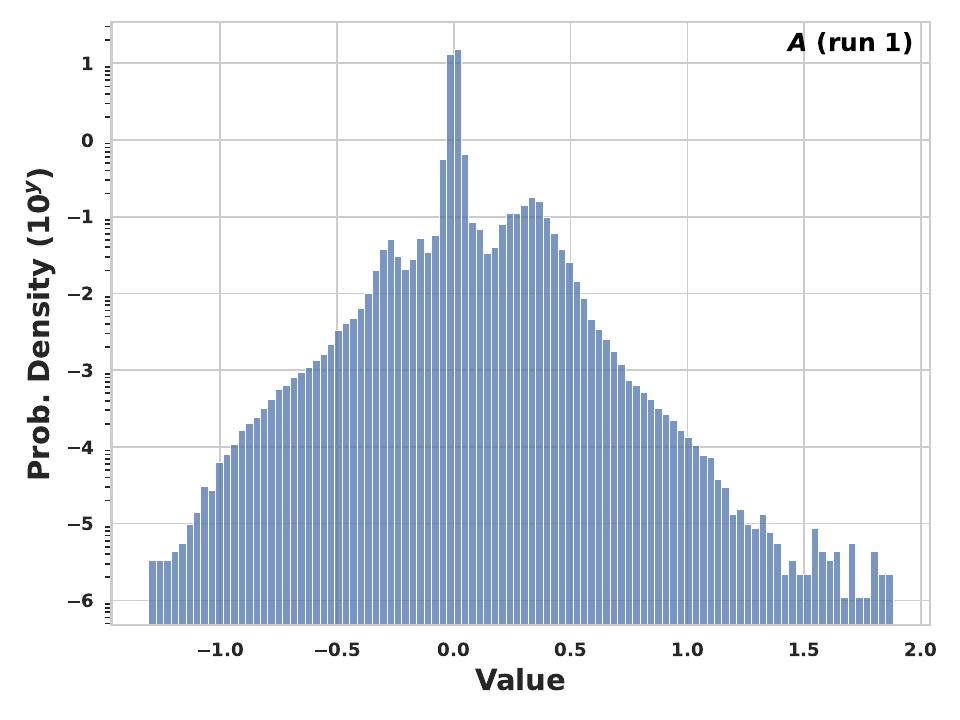}
		\caption{SUB-SOC-110M-1}
		\label{SUBSOC110M1}
	\end{subfigure}
	\caption{$\tA$ probability density distributions for all models binned in 100 buckets for main plots and insets.}
\end{figure}

\begin{figure}[!htb]
	\ContinuedFloat
	\centering
	\hfill
	\begin{subfigure}[b]{0.45\textwidth}
	\centering
	\includegraphics[width=1\textwidth]{A_histplot_ylog_xlin_SUB-SOC-110M-2_run1}
	\caption{SUB-SOC-110M-2}
	\label{SUBSOC110M2}
	\end{subfigure}
	\hfill
	\begin{subfigure}[b]{0.45\textwidth}
		\centering
		\includegraphics[width=1\textwidth]{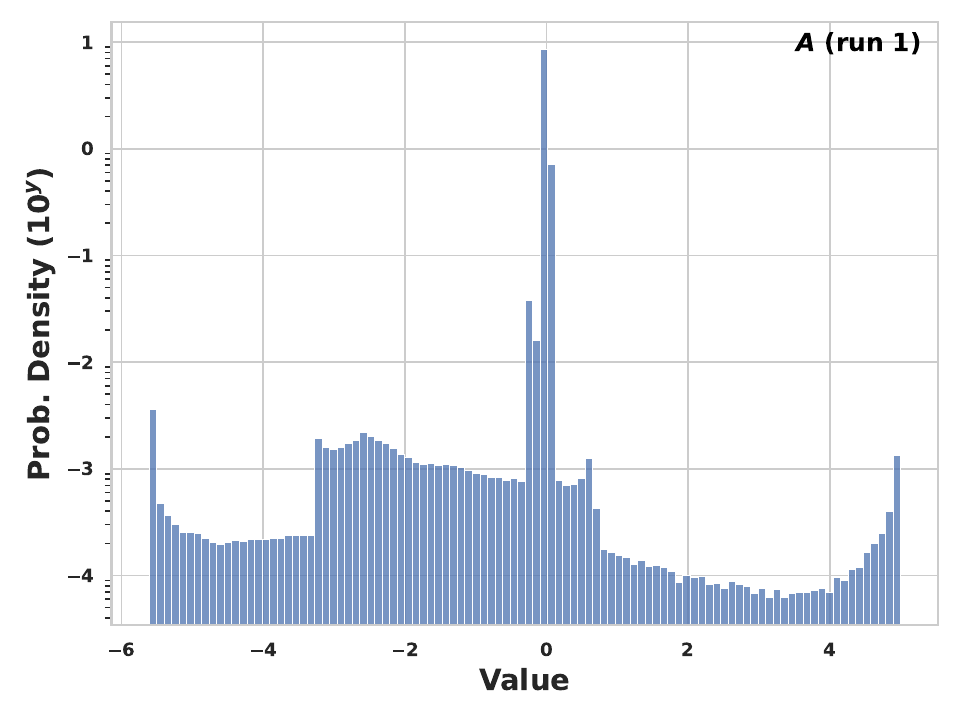}
		\caption{ABL-SOC-110M-1}
		\label{ABLSOC110M1}
	\end{subfigure}
	\hfill
	\begin{subfigure}[b]{0.45\textwidth}
		\centering
		\includegraphics[width=1\textwidth]{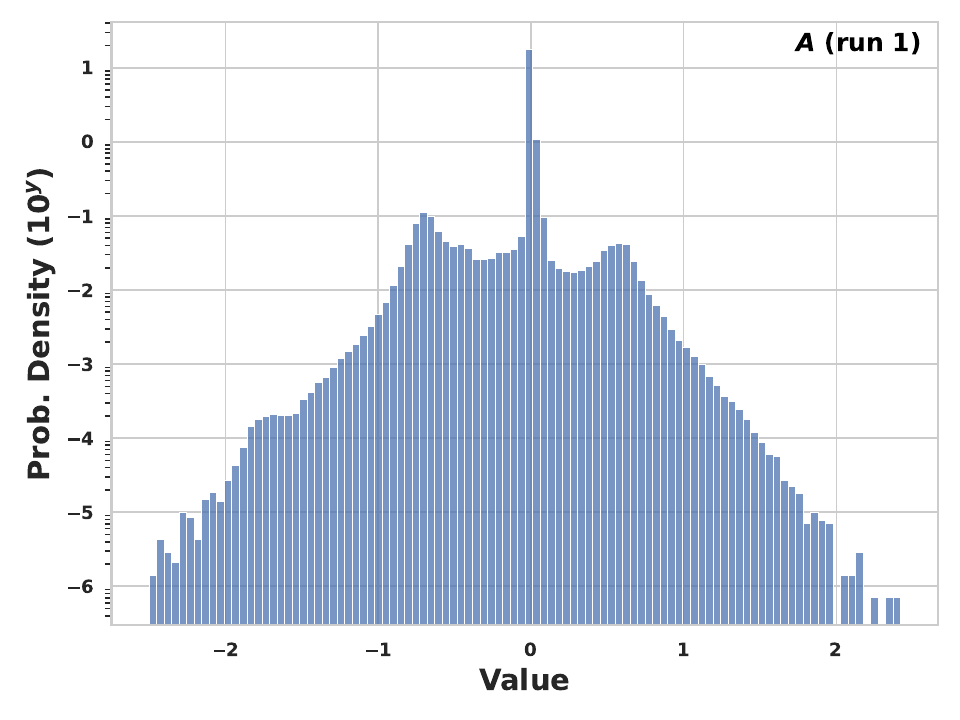}
		\caption{ABL-SOC-110M-2}
		\label{ABLSOC110M2}
	\end{subfigure}
	\hfill
	\begin{subfigure}[b]{0.45\textwidth}
		\centering
		\includegraphics[width=1\textwidth]{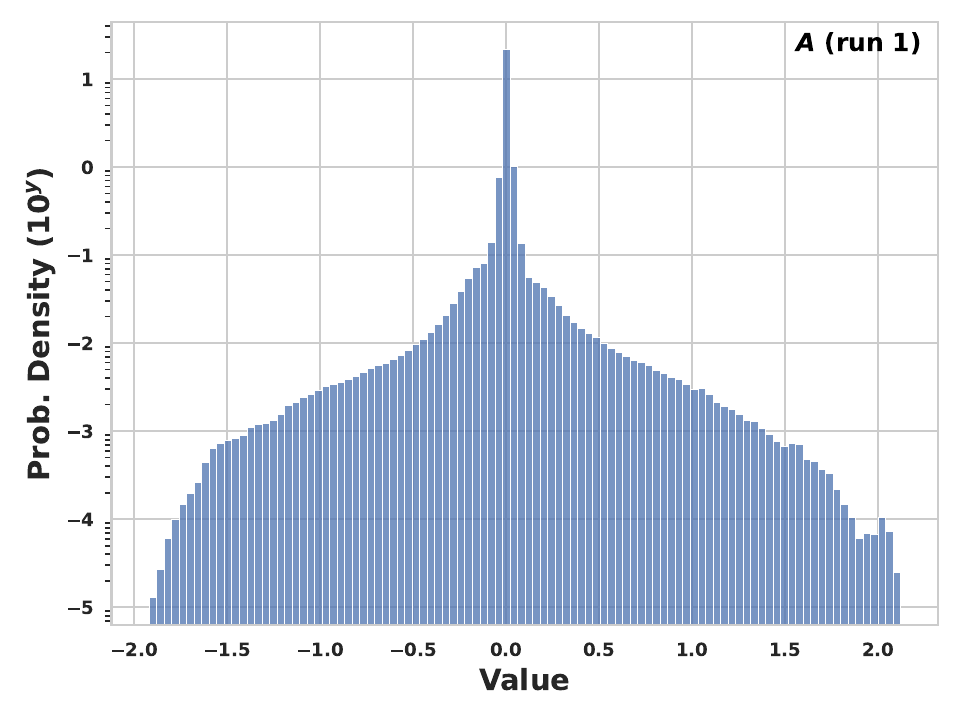}
		\caption{ABL-SOC-110M-3}
		\label{ABLSOC110M3}
	\end{subfigure}
	\caption{(Cont.) $\tA$ probability density distributions for all models binned in 100 buckets for main plots and insets.}
	\label{Ahistallmodels}
\end{figure}

%%%%%%%%%%%%%%%%%%%%%%%%%%%%%%%%%%%%%%%%%%%%%%%%%%%%%%%%%%%%%%%%%%%%%%%%%%%%%%%%%%%%%%%%%%%%%%%%%%%%%%%%%%%%%%%%%%%%%%%%%%%%%%%%%%%%%%%%%%
%ALM DISTRIBUTIONS
%%%%%%%%%%%%%%%%%%%%%%%%%%%%%%%%%%%%%%%%%%%%%%%%%%%%%%%%%%%%%%%%%%%%%%%%%%%%%%%%%%%%%%%%%%%%%%%%%%%%%%%%%%%%%%%%%%%%%%%%%%%%%%%%%%%%%%%%%%

\begin{figure}[!htb]
	\centering
	\begin{subfigure}[b]{0.45\textwidth}
		\centering
		\includegraphics[width=1\textwidth]{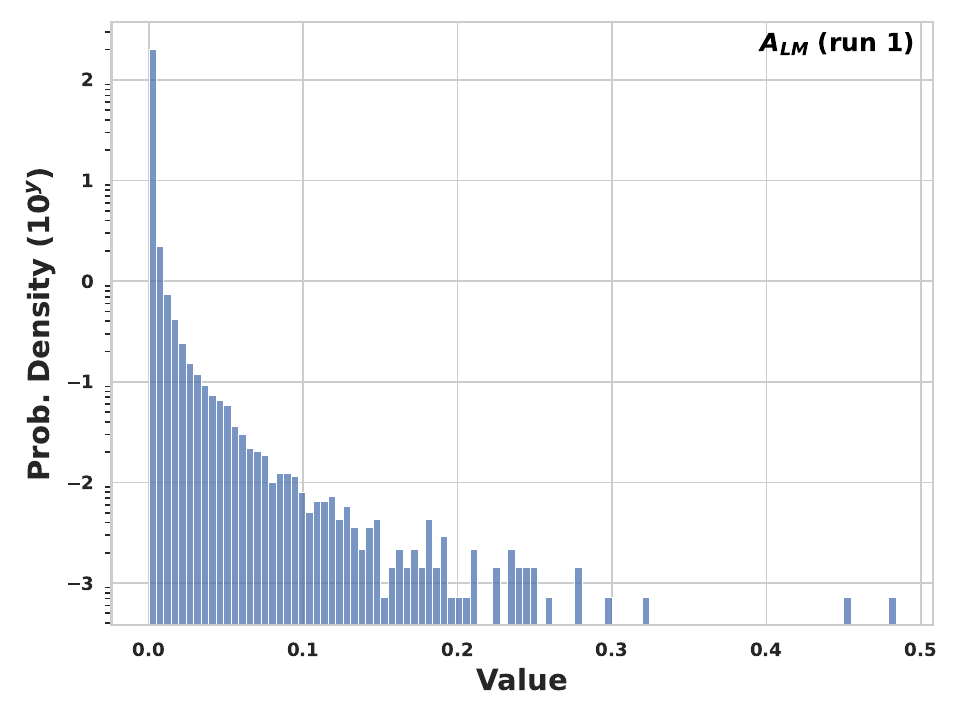}
		\caption{PLDRv51-SOC-110M-1}
		\label{ALMPLDRv51SOC110M1}
	\end{subfigure}
	\hfill
	\begin{subfigure}[b]{0.45\textwidth}
		\centering
		\includegraphics[width=1\textwidth]{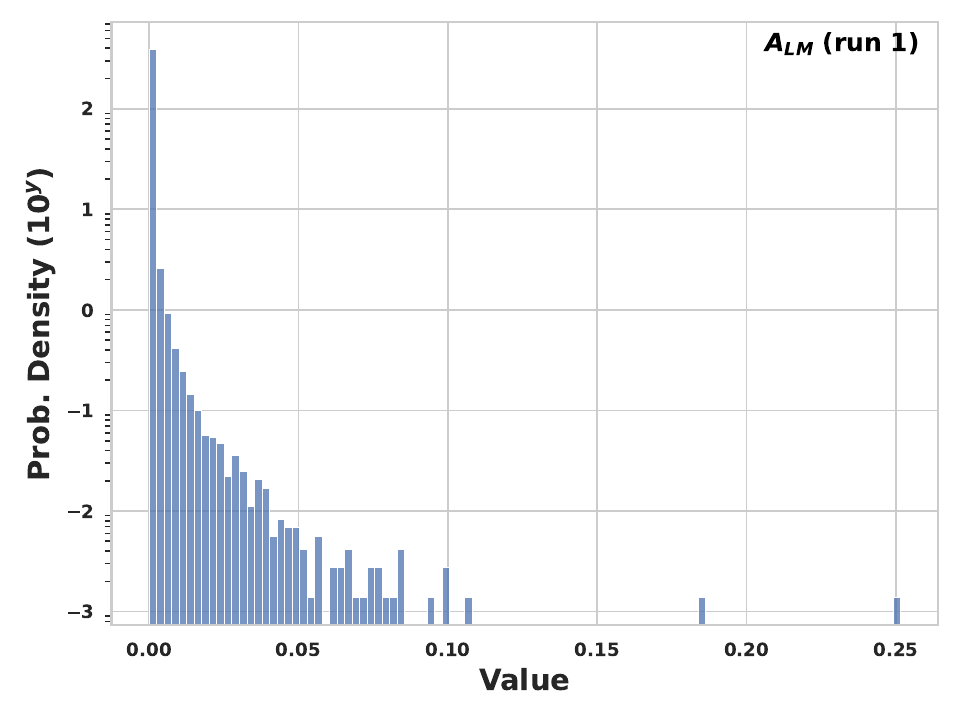}
		\caption{PLDRv51-SOC-110M-2}
		\label{ALMPLDRv51SOC110M2}
	\end{subfigure}
	\hfill
	\begin{subfigure}[b]{0.45\textwidth}
		\centering
		\includegraphics[width=1\textwidth]{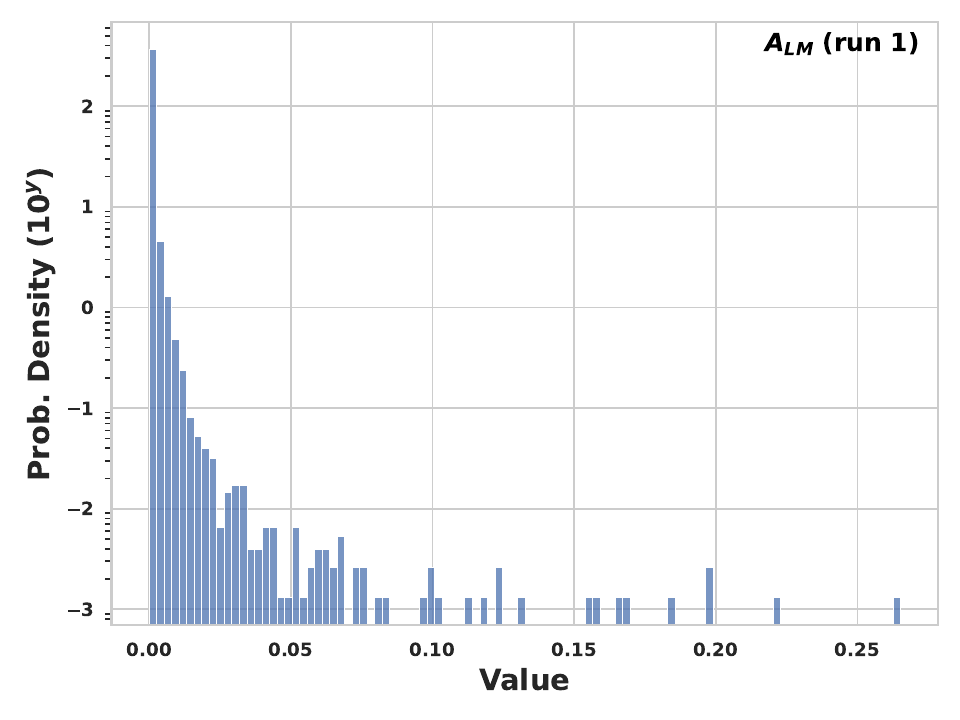}
		\caption{PLDRv51-SOC-110M-3}
		\label{ALMPLDRv51SOC110M3}
	\end{subfigure}
	\hfill
	\begin{subfigure}[b]{0.45\textwidth}
			\centering
			\includegraphics[width=1\textwidth]{ALM_histplot_ylog_xlin_PLDRv51-SOC-110M-4_run1}
			\caption{PLDRv51-SOC-110M-4}
			\label{ALMPLDRv51SOC110M4}
	\end{subfigure}
	\hfill
	\begin{subfigure}[b]{0.45\textwidth}
		\centering
		\includegraphics[width=1\textwidth]{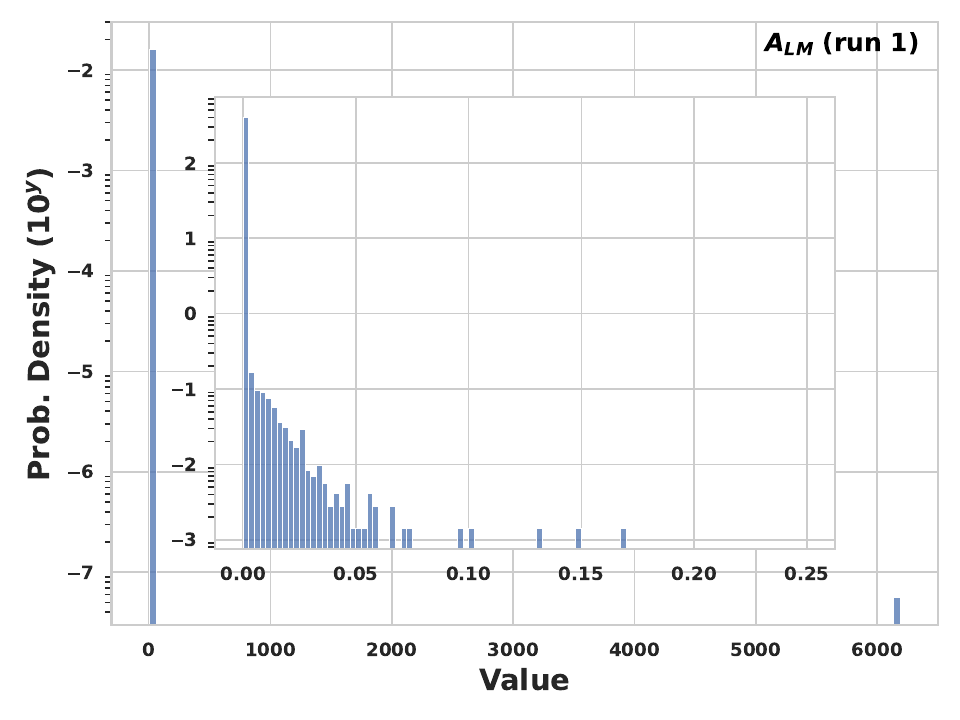}
		\caption{PLDRv51-SOC-110M-5}
		\label{ALMPLDRv51SOC110M5}
	\end{subfigure}
	\hfill
	\begin{subfigure}[b]{0.45\textwidth}
		\centering
		\includegraphics[width=1\textwidth]{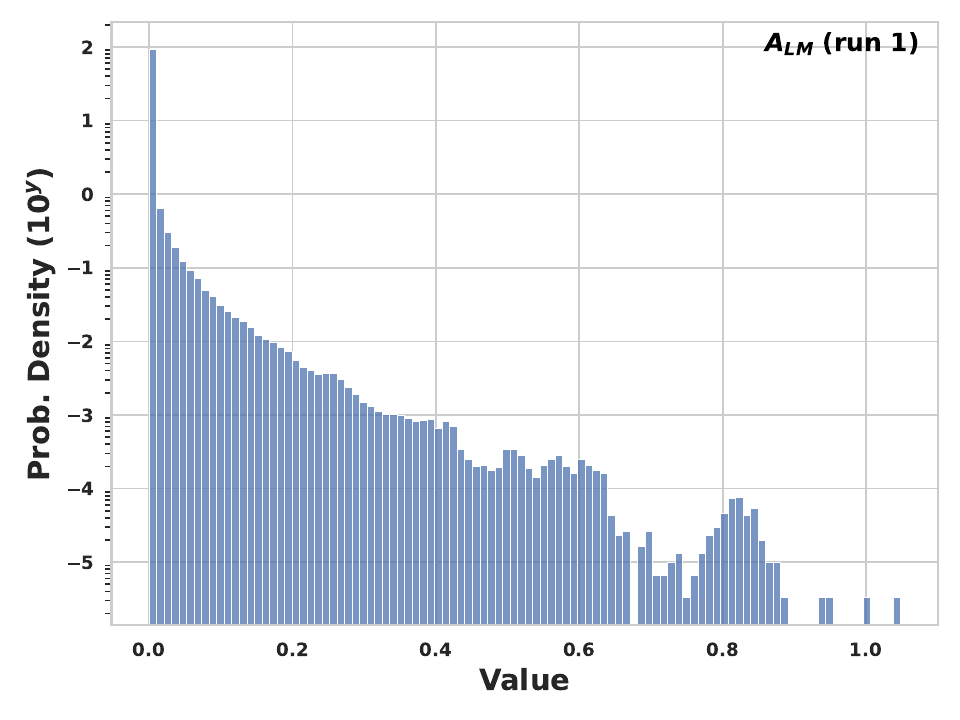}
		\caption{SUB-SOC-110M-1}
		\label{ALMSUBSOC110M1}
	\end{subfigure}
	\caption{$\tA_{LM}$ probability density distributions for all models binned in 100 buckets for main plots and insets.}
\end{figure}

\begin{figure}[!htb]
	\ContinuedFloat
	\centering
	\begin{subfigure}[b]{0.45\textwidth}
		\centering
		\includegraphics[width=1\textwidth]{ALM_histplot_ylog_xlin_SUB-SOC-110M-2_run1}
		\caption{SUB-SOC-110M-2}
		\label{ALMSUBSOC110M2}
	\end{subfigure}
	\hfill
	\begin{subfigure}[b]{0.45\textwidth}
		\centering
		\includegraphics[width=1\textwidth]{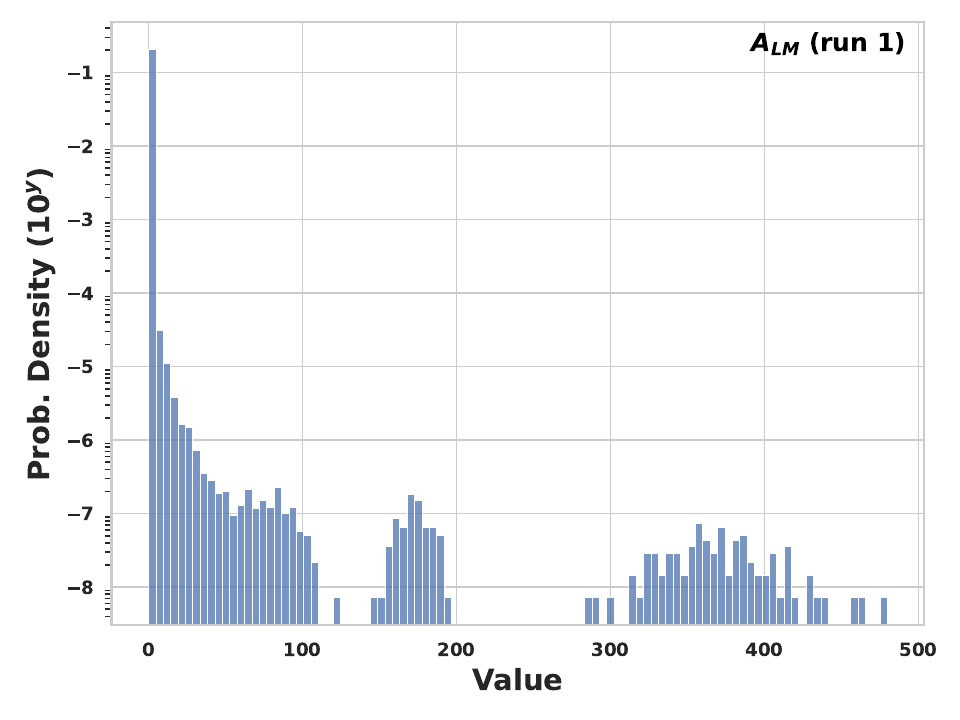}
		\caption{ABL-SOC-110M-1}
		\label{ALMABLSOC110M1}
	\end{subfigure}
	\hfill
	\begin{subfigure}[b]{0.45\textwidth}
		\centering
		\includegraphics[width=1\textwidth]{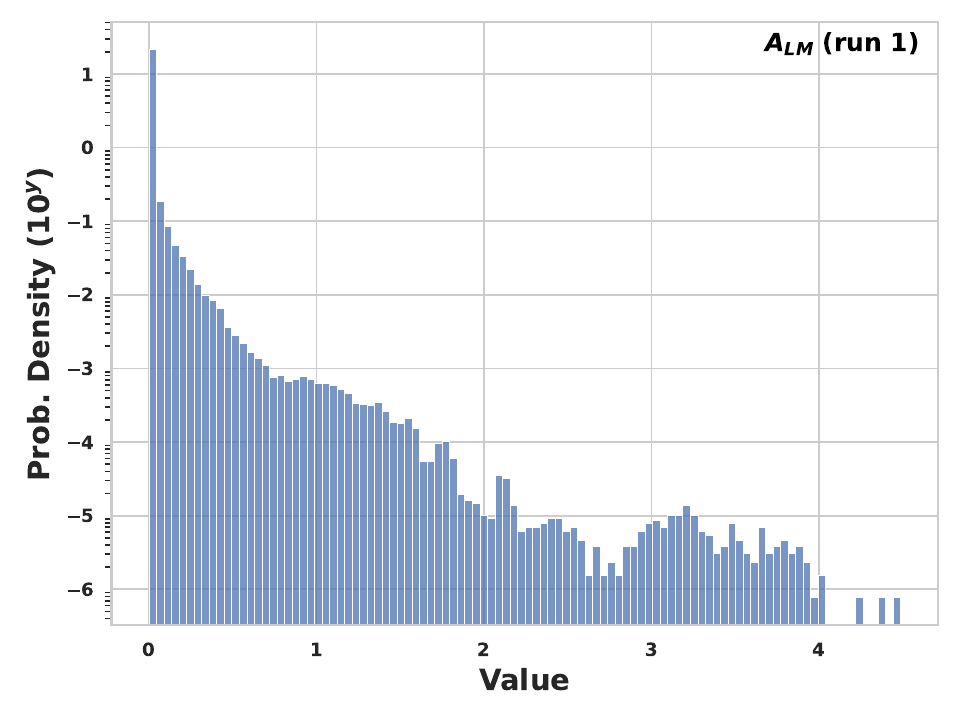}
		\caption{ABL-SOC-110M-2}
		\label{ALMABLSOC110M2}
	\end{subfigure}
	\hfill
	\begin{subfigure}[b]{0.45\textwidth}
		\centering
		\includegraphics[width=1\textwidth]{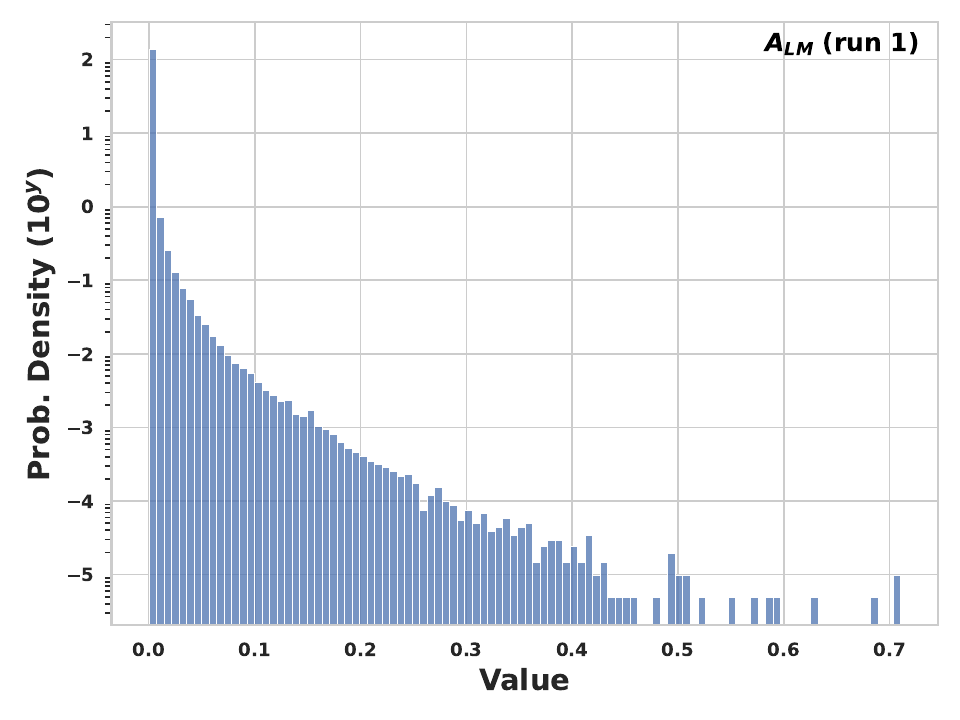}
		\caption{ABL-SOC-110M-3}
		\label{ALMABLSOC110M3}
	\end{subfigure}
	
	\caption{(Cont.) $\tA_{LM}$ probability density distributions for all models binned in 100 buckets for main plots and insets.}
	\label{ALMhistallmodels}
\end{figure}

%%%%%%%%%%%%%%%%%%%%%%%%%%%%%%%%%%%%%%%%%%%%%%%%%%%%%%%%%%%%%%%%%%%%%%%%%%%%%%%%%%%%%%%%%%%%%%%%%%%%%%%%%%%%%%%%%%%%%%%%%%%%%%%%%%%%%%
%Ap DISTRIBUTIONS
%%%%%%%%%%%%%%%%%%%%%%%%%%%%%%%%%%%%%%%%%%%%%%%%%%%%%%%%%%%%%%%%%%%%%%%%%%%%%%%%%%%%%%%%%%%%%%%%%%%%%%%%%%%%%%%%%%%%%%%%%%%%%%%%%%%%%%

\begin{figure}[!htb]
	\centering
	\begin{subfigure}[b]{0.45\textwidth}
		\centering
		\includegraphics[width=1\textwidth]{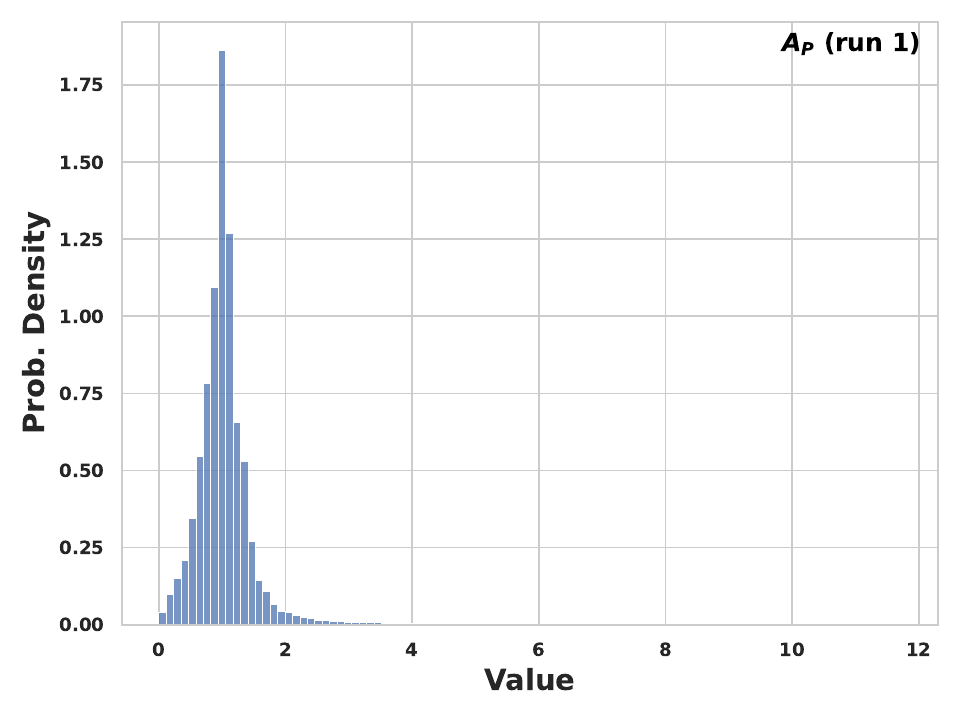}
		\caption{PLDRv51-SOC-110M-1}
		\label{ApPLDRv51SOC110M1}
	\end{subfigure}
	\hfill
	\begin{subfigure}[b]{0.45\textwidth}
		\centering
		\includegraphics[width=1\textwidth]{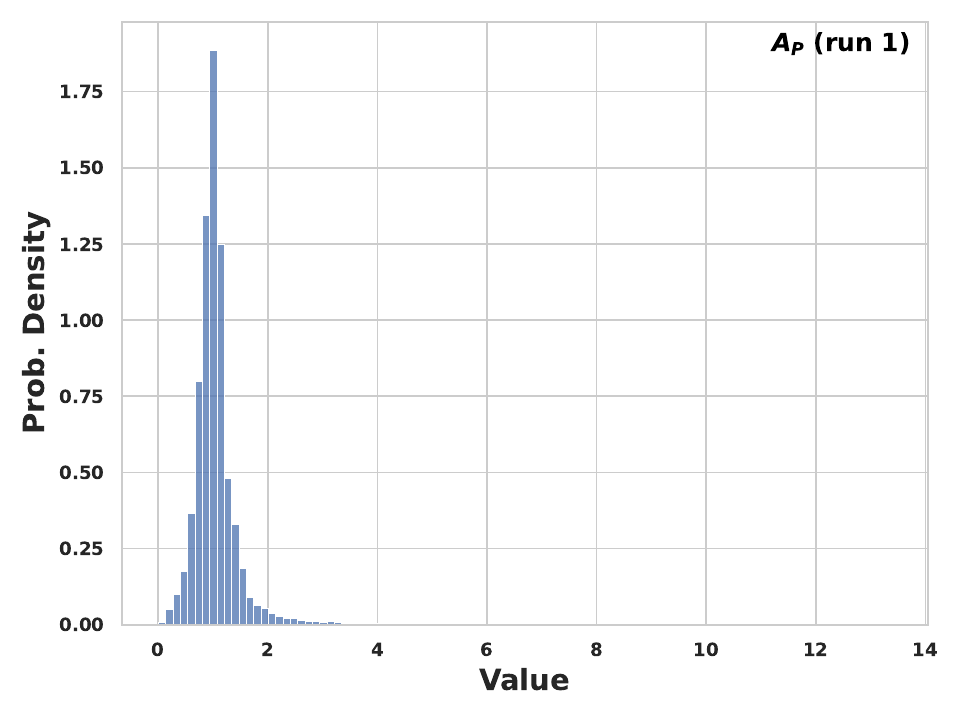}
		\caption{PLDRv51-SOC-110M-2}
		\label{ApPLDRv51SOC110M2}
	\end{subfigure}
	\hfill
	\begin{subfigure}[b]{0.45\textwidth}
		\centering
		\includegraphics[width=1\textwidth]{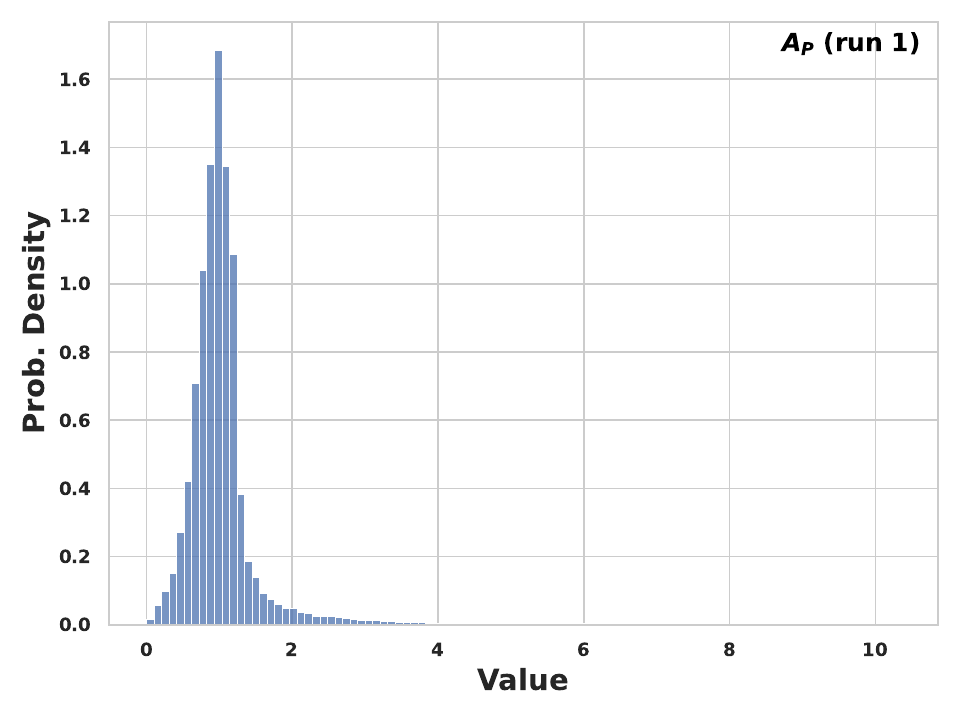}
		\caption{PLDRv51-SOC-110M-3}
		\label{ApPLDRv51SOC110M3}
	\end{subfigure}
	\hfill
	\begin{subfigure}[b]{0.45\textwidth}
			\centering
			\includegraphics[width=1\textwidth]{Ap_histplot_PLDRv51-SOC-110M-4_run1}
			\caption{PLDRv51-SOC-110M-4}
			\label{ApPLDRv51SOC110M4}
	\end{subfigure}
	\hfill
	\begin{subfigure}[b]{0.45\textwidth}
		\centering
		\includegraphics[width=1\textwidth]{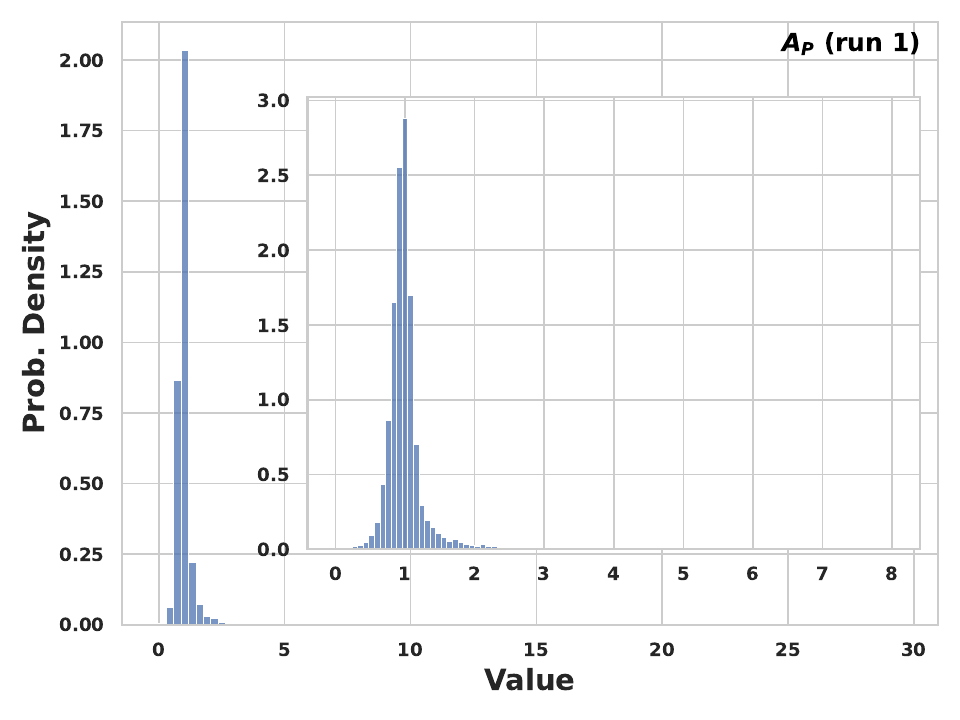}
		\caption{PLDRv51-SOC-110M-5}
		\label{ApPLDRv51SOC110M5}
	\end{subfigure}
	\hfill
	\begin{subfigure}[b]{0.45\textwidth}
		\centering
		\includegraphics[width=1\textwidth]{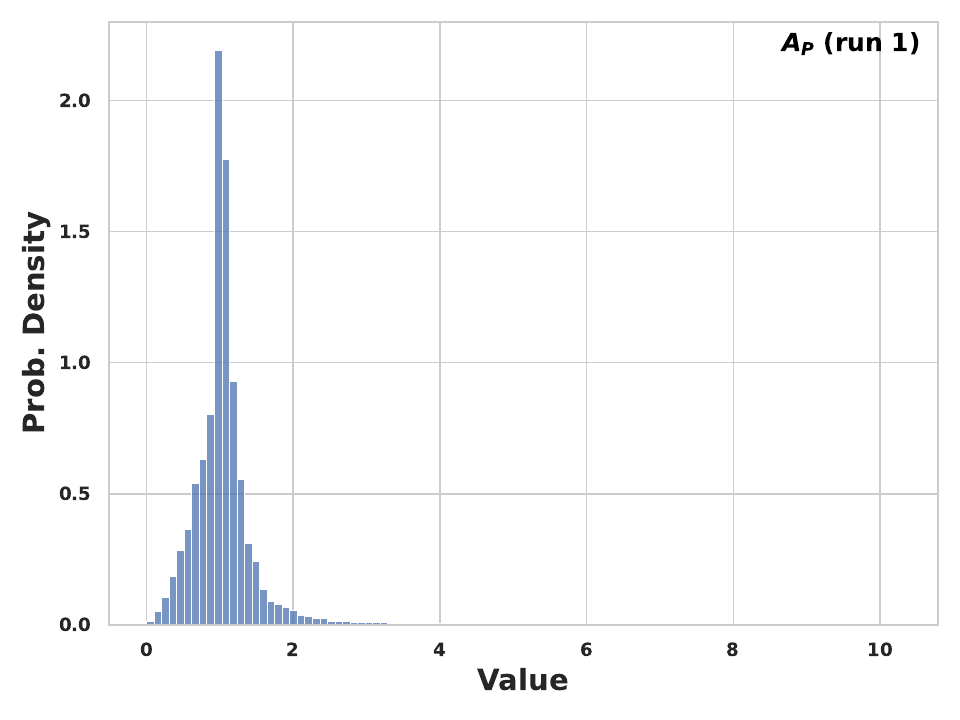}
		\caption{SUB-SOC-110M-1}
		\label{ApSUBSOC110M1}
	\end{subfigure}
	\caption{$\tA_{\textbf{P}}$ probability density distributions for all models binned in 100 buckets for main plots and insets. The $\tA_{\textbf{P}}$ were plotted up to $\pm5\sigma$ for easier visibility of main distribution characteristics.}
	\end{figure}

	\begin{figure}[!htb]
	\ContinuedFloat
	\centering
	\begin{subfigure}[b]{0.45\textwidth}
		\centering
		\includegraphics[width=1\textwidth]{Ap_histplot_SUB-SOC-110M-2_run1}
		\caption{SUB-SOC-110M-2}
		\label{ApSUBSOC110M2}
	\end{subfigure}
	\hfill
	\begin{subfigure}[b]{0.45\textwidth}
		\centering
		\includegraphics[width=1\textwidth]{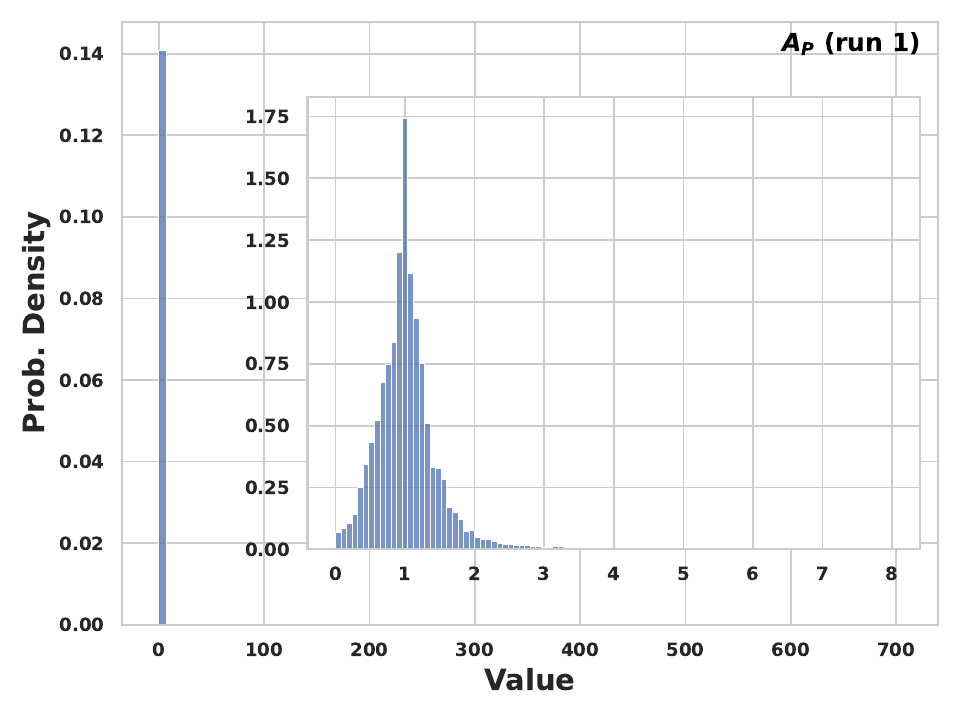}
		\caption{ABL-SOC-110M-1}
		\label{ApABLSOC110M1}
	\end{subfigure}
	\hfill
	\begin{subfigure}[b]{0.45\textwidth}
		\centering
		\includegraphics[width=1\textwidth]{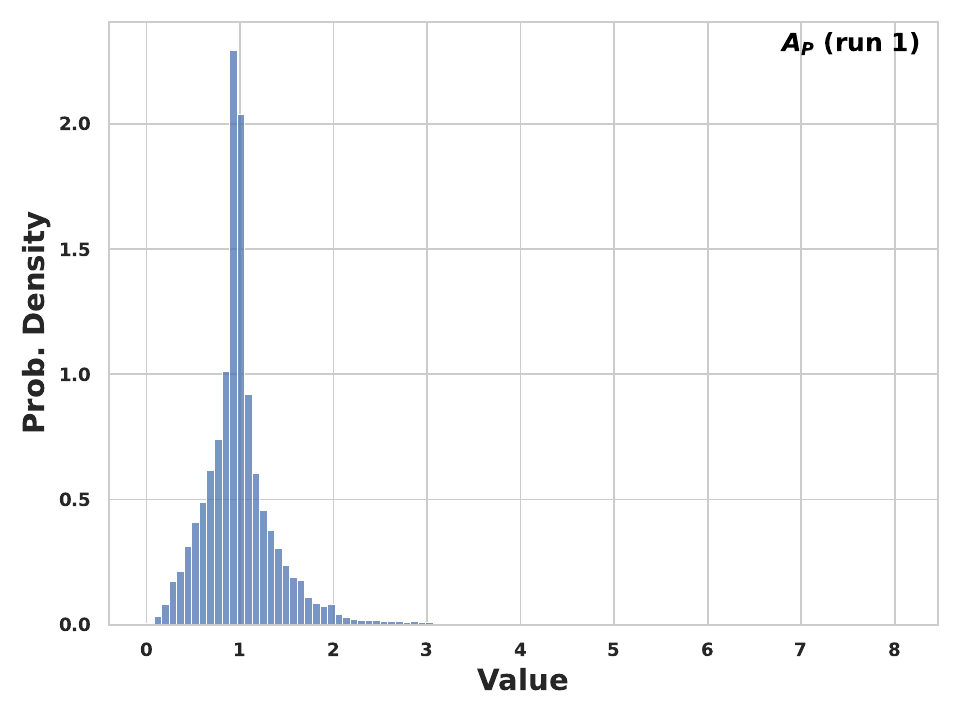}
		\caption{ABL-SOC-110M-2}
		\label{ApABLSOC110M2}
	\end{subfigure}
	\hfill
	\begin{subfigure}[b]{0.45\textwidth}
		\centering
		\includegraphics[width=1\textwidth]{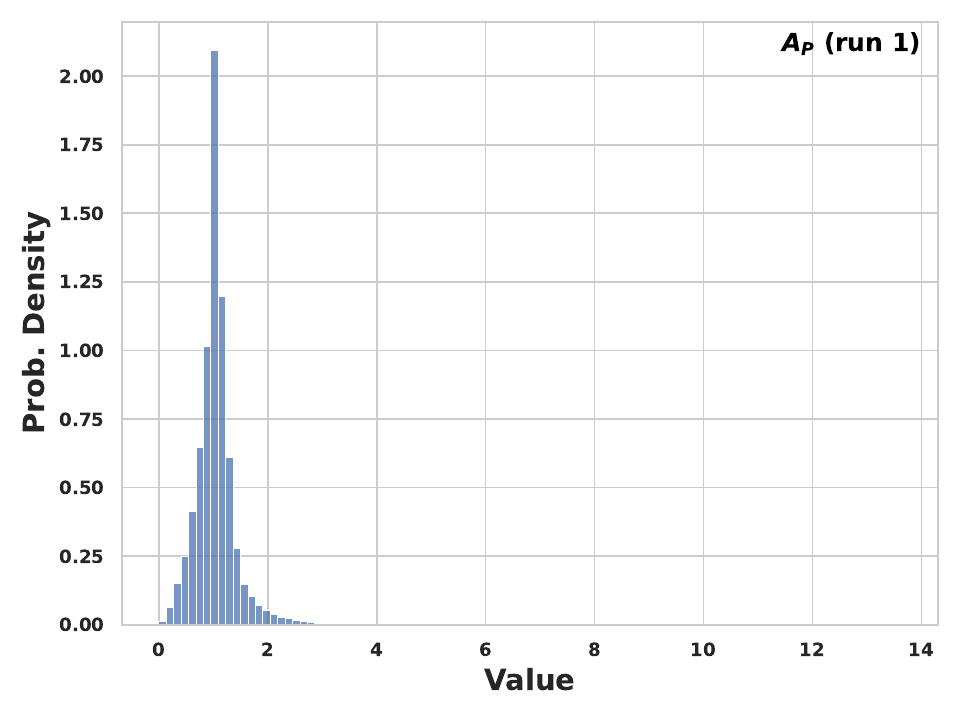}
		\caption{ABL-SOC-110M-3}
		\label{ApABLSOC110M3}
	\end{subfigure}
	\caption{(Cont.) $\tA_{\textbf{P}}$ probability density distributions for all models binned in 100 buckets for main plots and insets. The $\tA_{\textbf{P}}$ were plotted up to $\pm5\sigma$ for easier visibility of main distribution characteristics.}
	\label{Aphistallmodels}
\end{figure}

%%%%%%%%%%%%%%%%%%%%%%%%%%%%%%%%%%%%%%%%%%%%%%%%%%%%%%%%%%%%%%%%%%%%%%%%%%%%%%%%%%%%%%%%%%%%%%%%%%%%%%%%%%%%%%%%%%%%%%%%%%%%%%%%%%%%%%
%GLM DISTRIBUTIONS
%%%%%%%%%%%%%%%%%%%%%%%%%%%%%%%%%%%%%%%%%%%%%%%%%%%%%%%%%%%%%%%%%%%%%%%%%%%%%%%%%%%%%%%%%%%%%%%%%%%%%%%%%%%%%%%%%%%%%%%%%%%%%%%%%%%%%%

\begin{figure}[!htb]
	\centering
	\begin{subfigure}[b]{0.45\textwidth}
		\centering
		\includegraphics[width=1\textwidth]{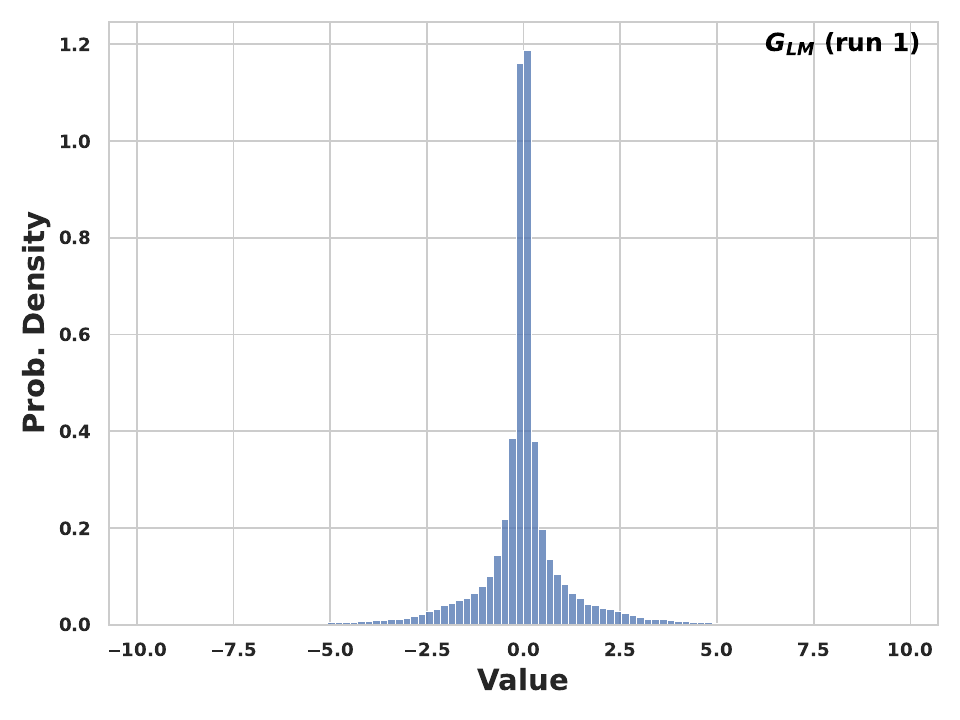}
		\caption{PLDRv51-SOC-110M-1}
		\label{GLMPLDRv51SOC110M1}
	\end{subfigure}
	\hfill
	\begin{subfigure}[b]{0.45\textwidth}
		\centering
		\includegraphics[width=1\textwidth]{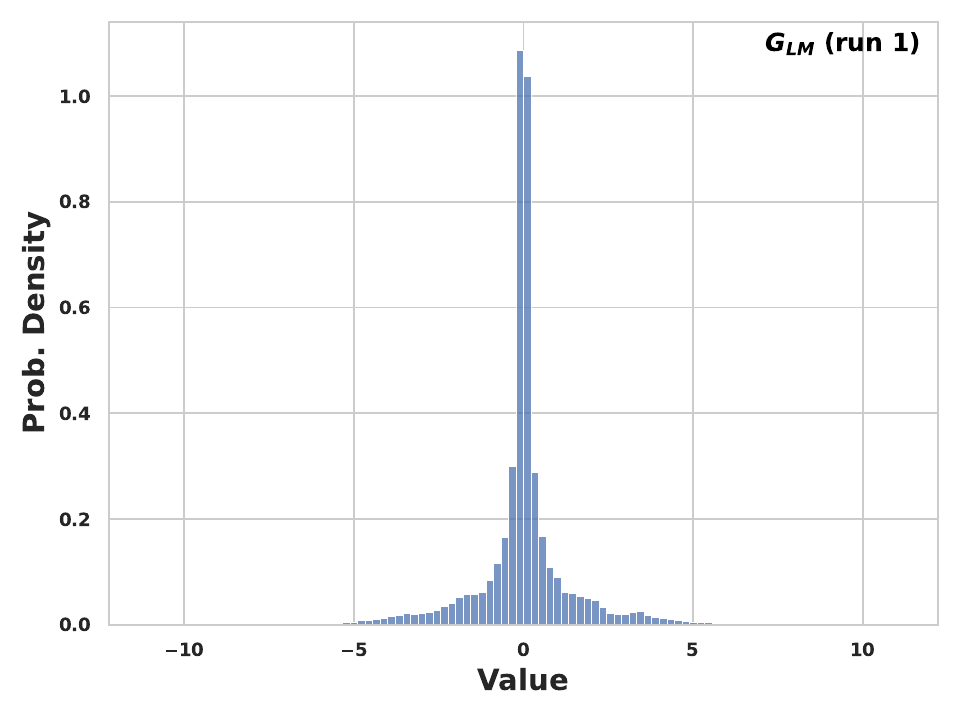}
		\caption{PLDRv51-SOC-110M-2}
		\label{GLMPLDRv51SOC110M2}
	\end{subfigure}
	\hfill
	\begin{subfigure}[b]{0.45\textwidth}
		\centering
		\includegraphics[width=1\textwidth]{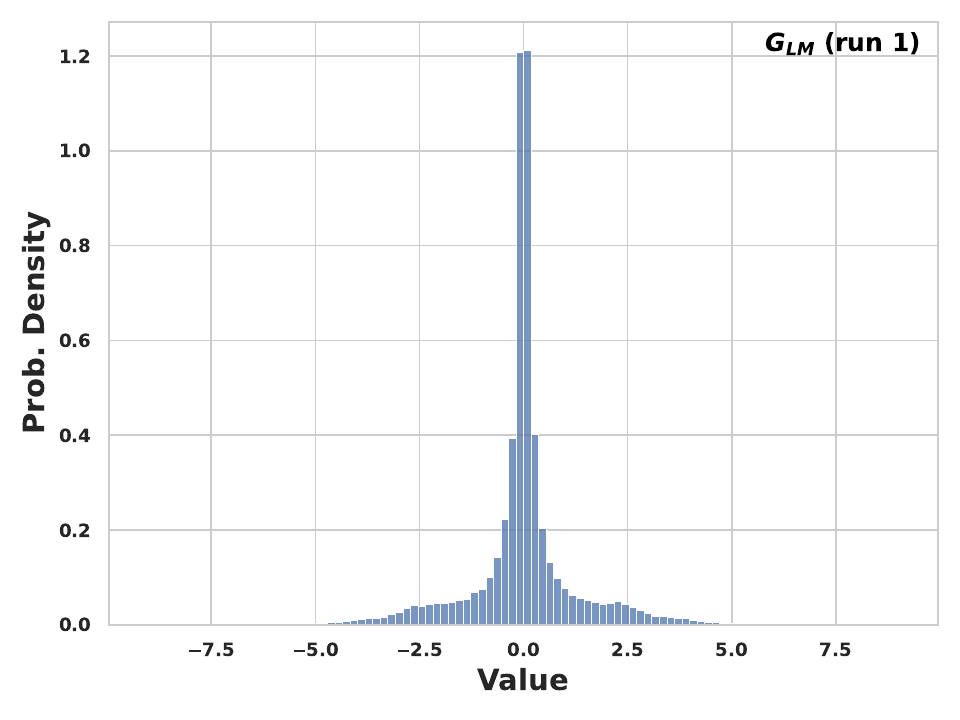}
		\caption{PLDRv51-SOC-110M-3}
		\label{GLMPLDRv51SOC110M3}
	\end{subfigure}
	\hfill
	\begin{subfigure}[b]{0.45\textwidth}
			\centering
			\includegraphics[width=1\textwidth]{GLM_histplot_PLDRv51-SOC-110M-4_run1}
			\caption{PLDRv51-SOC-110M-4}
			\label{GLMPLDRv51SOC110M4}
	\end{subfigure}
	\hfill
	\begin{subfigure}[b]{0.45\textwidth}
		\centering
		\includegraphics[width=1\textwidth]{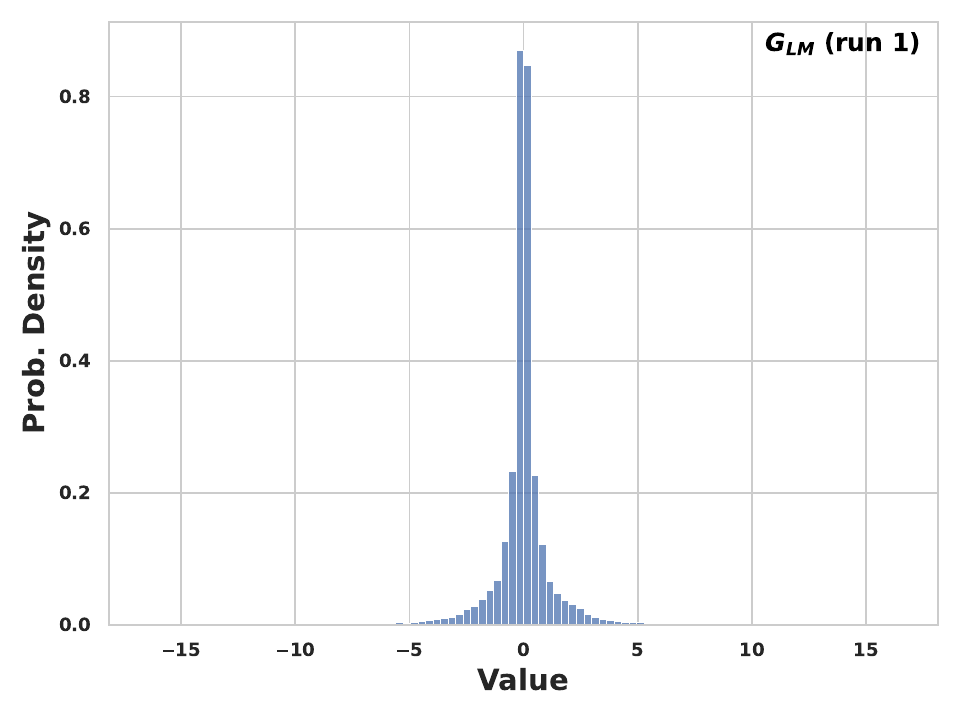}
		\caption{PLDRv51-SOC-110M-5}
		\label{GLMPLDRv51SOC110M5}
	\end{subfigure}
	\hfill
	\begin{subfigure}[b]{0.45\textwidth}
		\centering
		\includegraphics[width=1\textwidth]{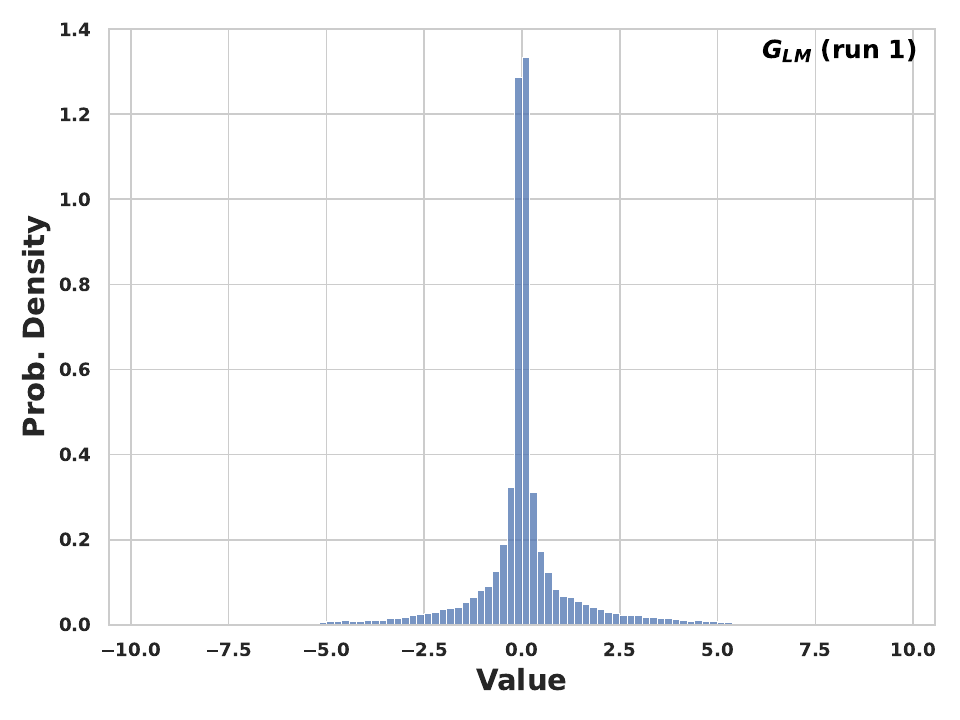}
		\caption{SUB-SOC-110M-1}
		\label{GLMSUBSOC110M1}
	\end{subfigure}
	\caption{$\tG_{LM}$ probability density distributions for all models binned in 100 buckets. The $\tG_{LM}$ were plotted up to $\pm5\sigma$ for easier visibility of main distribution characteristics.}
\end{figure}

\begin{figure}[!htb]
	\ContinuedFloat
	\centering
	\begin{subfigure}[b]{0.45\textwidth}
		\centering
		\includegraphics[width=1\textwidth]{GLM_histplot_SUB-SOC-110M-2_run1}
		\caption{SUB-SOC-110M-2}
		\label{GLMSUBSOC110M2}
	\end{subfigure}
	\hfill
	\begin{subfigure}[b]{0.45\textwidth}
		\centering
		\includegraphics[width=1\textwidth]{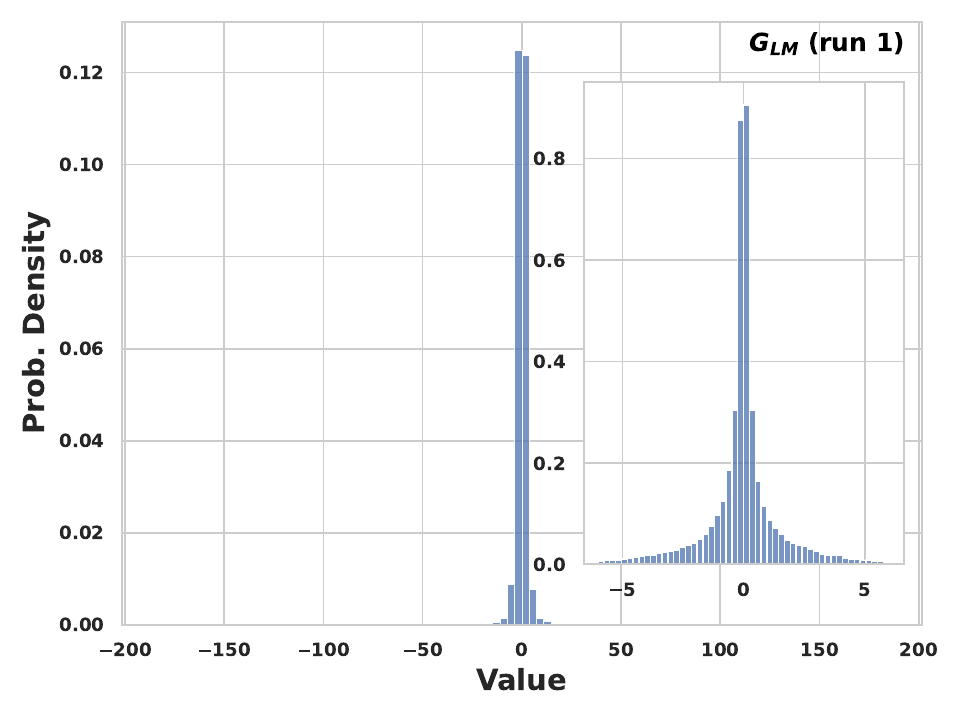}
		\caption{ABL-SOC-110M-1}
		\label{GLMABLSOC110M1}
	\end{subfigure}
	\hfill
	\begin{subfigure}[b]{0.45\textwidth}
		\centering
		\includegraphics[width=1\textwidth]{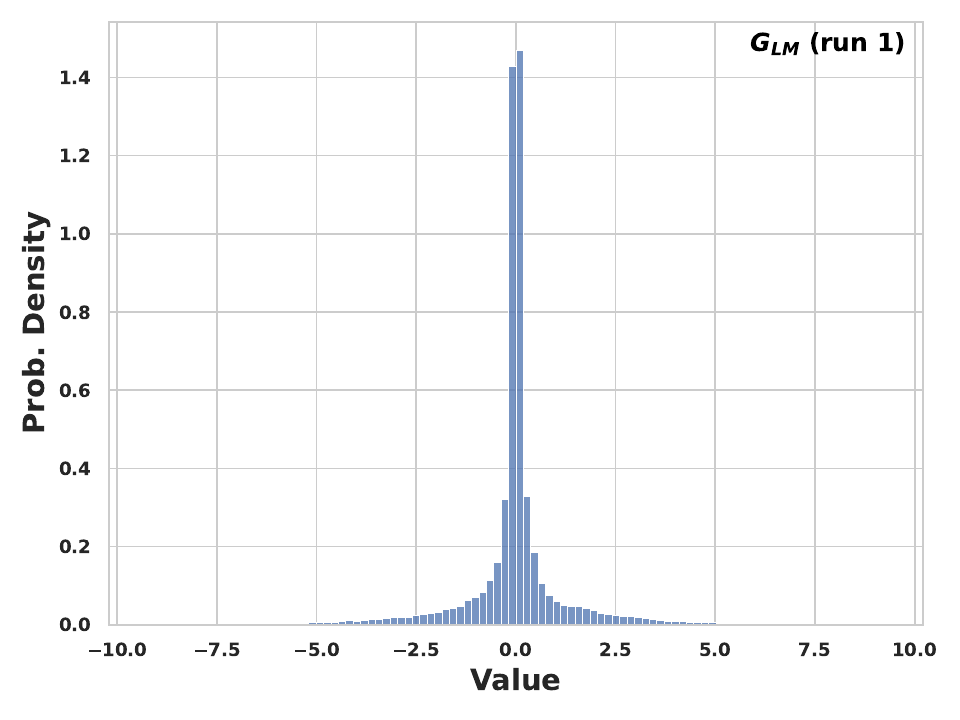}
		\caption{ABL-SOC-110M-2}
		\label{GLMABLSOC110M2}
	\end{subfigure}
	\hfill
	\begin{subfigure}[b]{0.45\textwidth}
		\centering
		\includegraphics[width=1\textwidth]{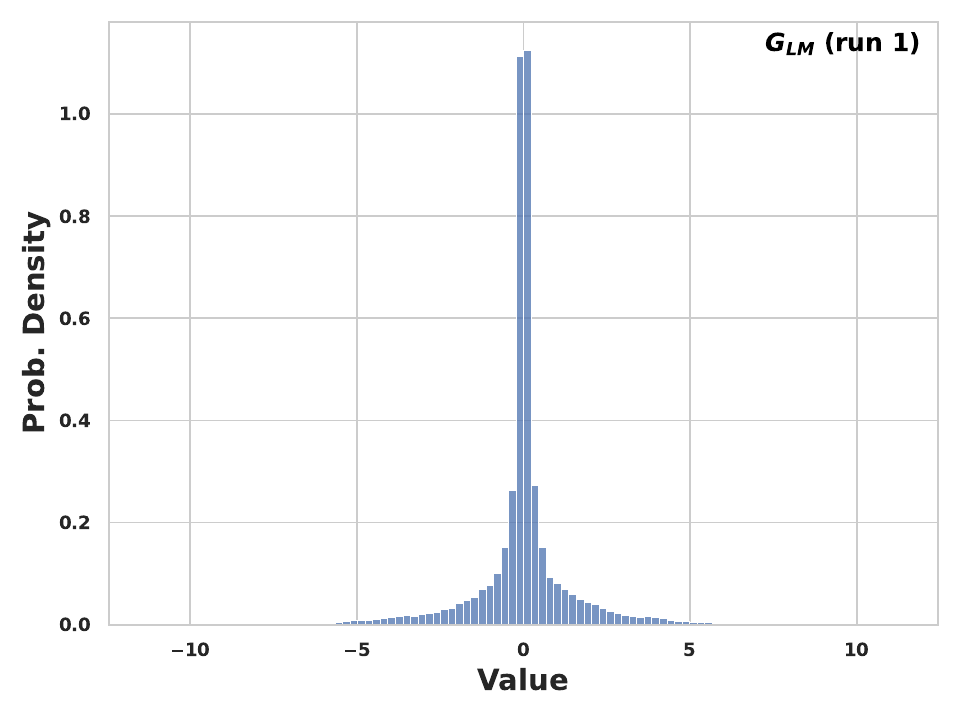}
		\caption{ABL-SOC-110M-3}
		\label{GLMABLSOC110M3}
	\end{subfigure}	
	\caption{(Cont.) $\tG_{LM}$ probability density distributions for all models binned in 100 buckets. Inset for ABL-SOC-110M-1 was binned in 50 buckets. The $\tG_{LM}$ were plotted up to $\pm5\sigma$ for easier visibility of main distribution characteristics.}
	\label{GLMhistallmodels}
\end{figure}

%%%%%%%%%%%%%%%%%%%%%%%%%%%%%%%%%%%%%%%%%%%%%%%%%%%%%%%%%%%%%%%%%%%%%%%%%%%%%%%%%%%%%%%%%%%%%%%%%%%%%%%%%%%%%%%%%%%%%%%%%%%%%%%%%%%%%%
%A HEATMAP DISTRIBUTIONS
%%%%%%%%%%%%%%%%%%%%%%%%%%%%%%%%%%%%%%%%%%%%%%%%%%%%%%%%%%%%%%%%%%%%%%%%%%%%%%%%%%%%%%%%%%%%%%%%%%%%%%%%%%%%%%%%%%%%%%%%%%%%%%%%%%%%%%

\begin{figure}[!htb]
	\centering
	\begin{subfigure}[b]{0.45\textwidth}
		\centering
		\includegraphics[width=1\textwidth]{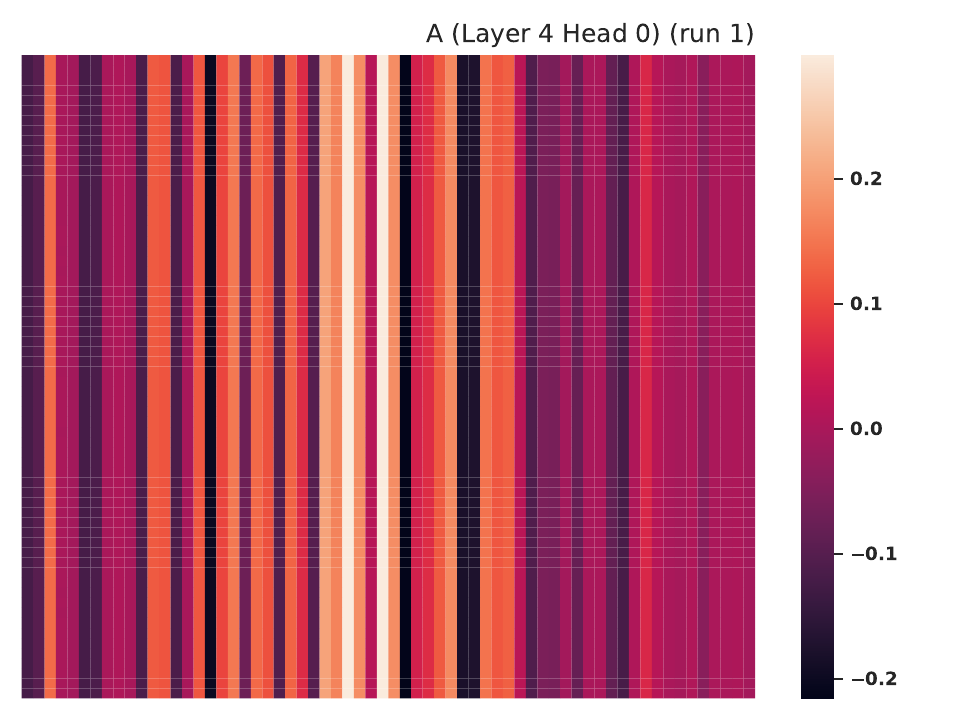}
		\caption{PLDRv51-SOC-110M-1}
		\label{AhmPLDRv51SOC110M1}
	\end{subfigure}
	\hfill
	\begin{subfigure}[b]{0.45\textwidth}
		\centering
		\includegraphics[width=1\textwidth]{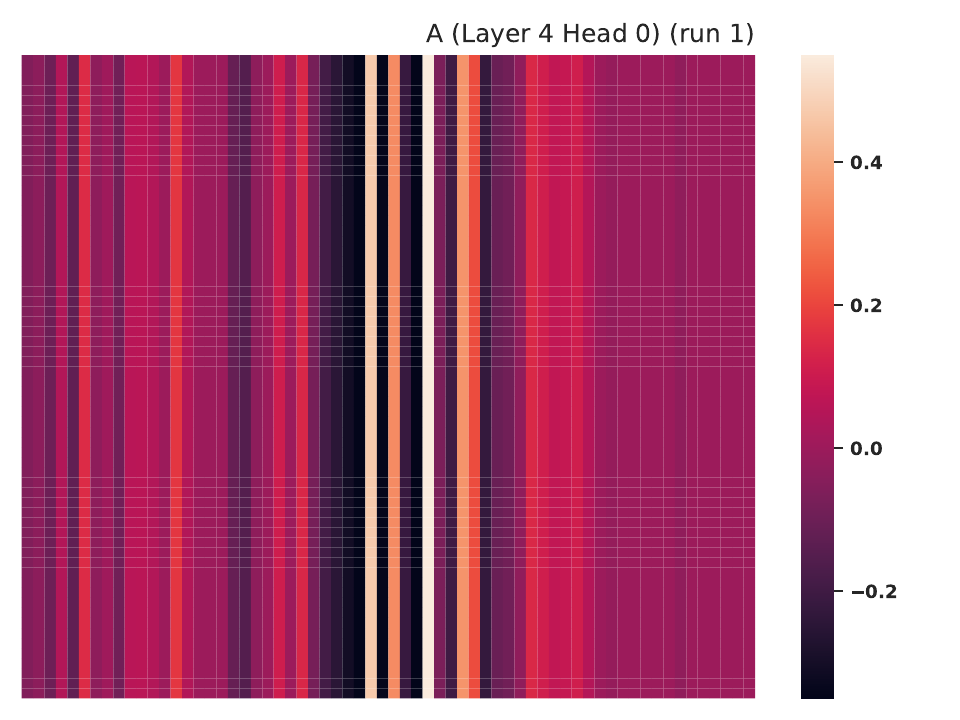}
		\caption{PLDRv51-SOC-110M-2}
		\label{AhmPLDRv51SOC110M2}
	\end{subfigure}
	\hfill
	\begin{subfigure}[b]{0.45\textwidth}
		\centering
		\includegraphics[width=1\textwidth]{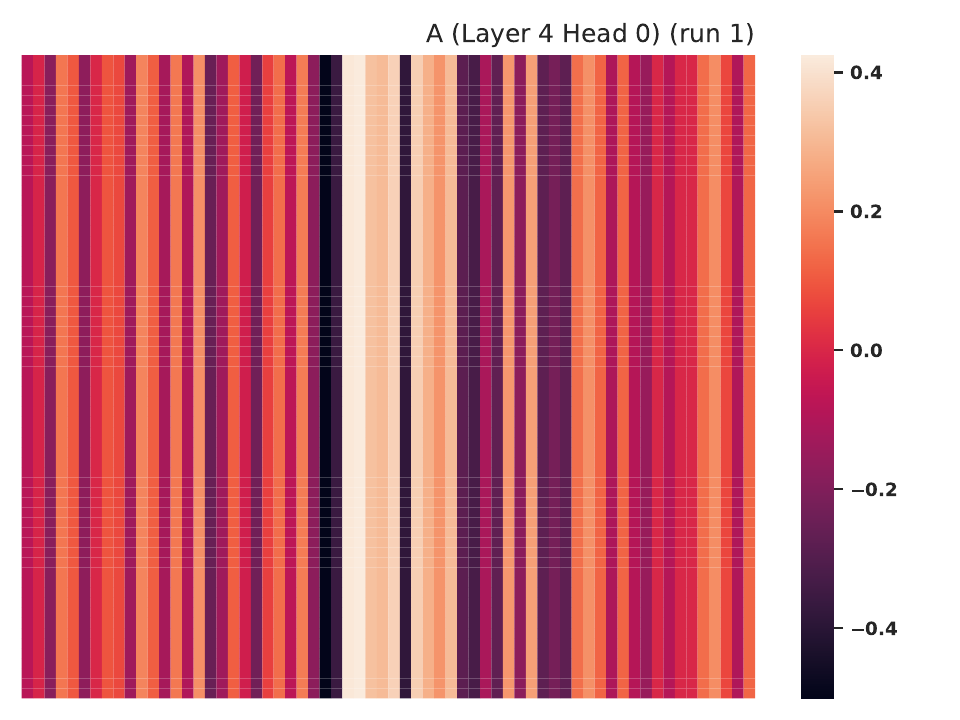}
		\caption{PLDRv51-SOC-110M-3}
		\label{AhmPLDRv51SOC110M3}
	\end{subfigure}
	\hfill
	\begin{subfigure}[b]{0.45\textwidth}
		\centering
		\includegraphics[width=1\textwidth]{A_heatmapplot_PLDRv51-SOC-110M-4_run1}
		\caption{PLDRv51-SOC-110M-4}
		\label{AhmPLDRv51SOC110M4}
	\end{subfigure}
	\hfill
	\begin{subfigure}[b]{0.45\textwidth}
		\centering
		\includegraphics[width=1\textwidth]{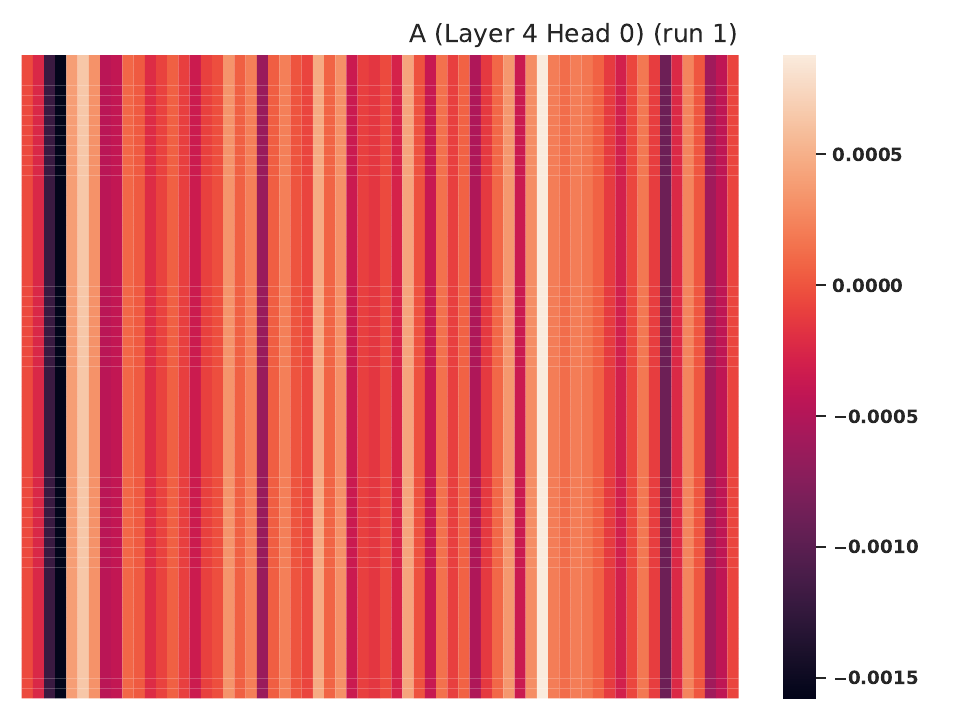}
		\caption{PLDRv51-SOC-110M-5}
		\label{AhmPLDRv51SOC110M5}
	\end{subfigure}
	\hfill
	\begin{subfigure}[b]{0.45\textwidth}
		\centering
		\includegraphics[width=1\textwidth]{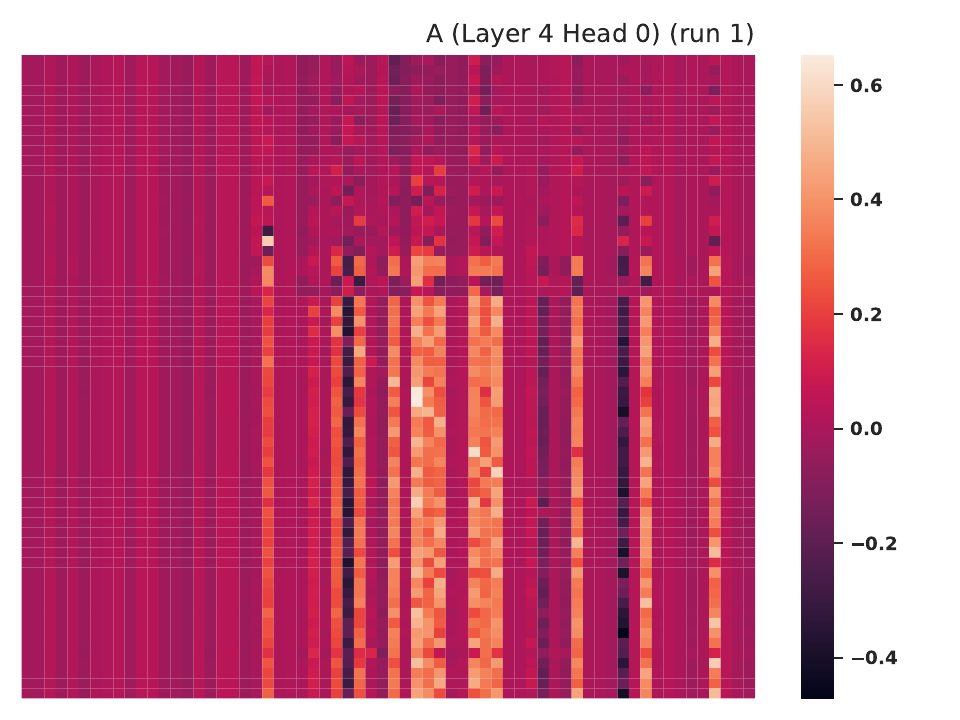}
		\caption{SUB-SOC-110M-1}
		\label{AhmSUBSOC110M1}
	\end{subfigure}
	\caption{$\tA$ heat maps at last decoder layer and for a single head of all models averaged over all samples.}
\end{figure}

\begin{figure}[!htb]
	\ContinuedFloat
	\centering
	\begin{subfigure}[b]{0.45\textwidth}
		\centering
		\includegraphics[width=1\textwidth]{A_heatmapplot_SUB-SOC-110M-2_run1}
		\caption{SUB-SOC-110M-2}
		\label{AhmSUBSOC110M2}
	\end{subfigure}
	\hfill
	\begin{subfigure}[b]{0.45\textwidth}
		\centering
		\includegraphics[width=1\textwidth]{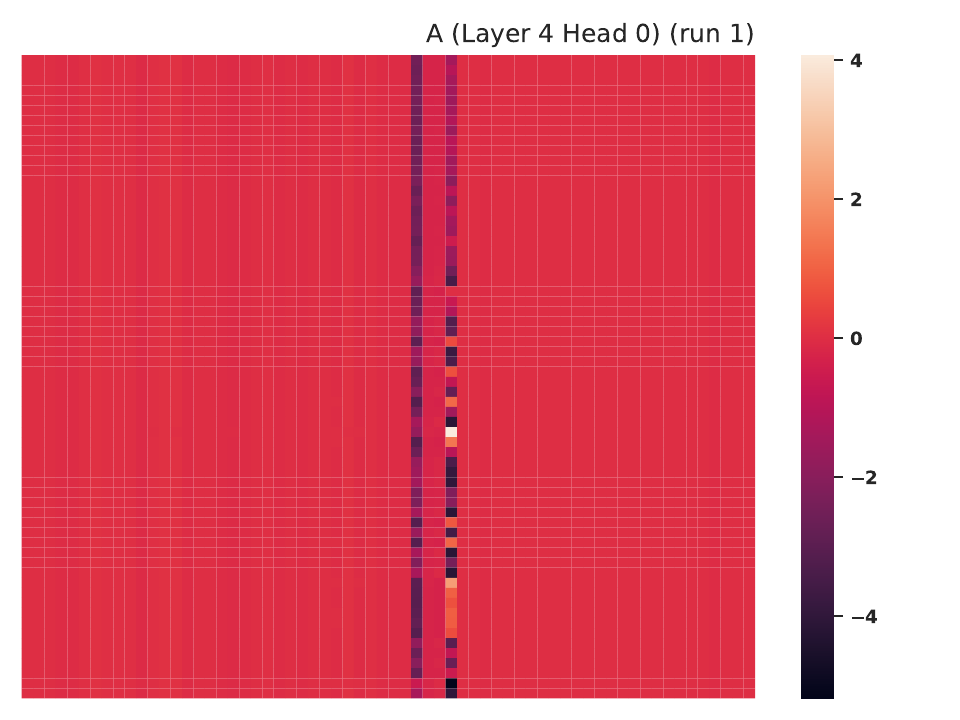}
		\caption{ABL-SOC-110M-1}
		\label{AhmABLSOC110M1}
	\end{subfigure}
	\hfill
	\begin{subfigure}[b]{0.45\textwidth}
		\centering
		\includegraphics[width=1\textwidth]{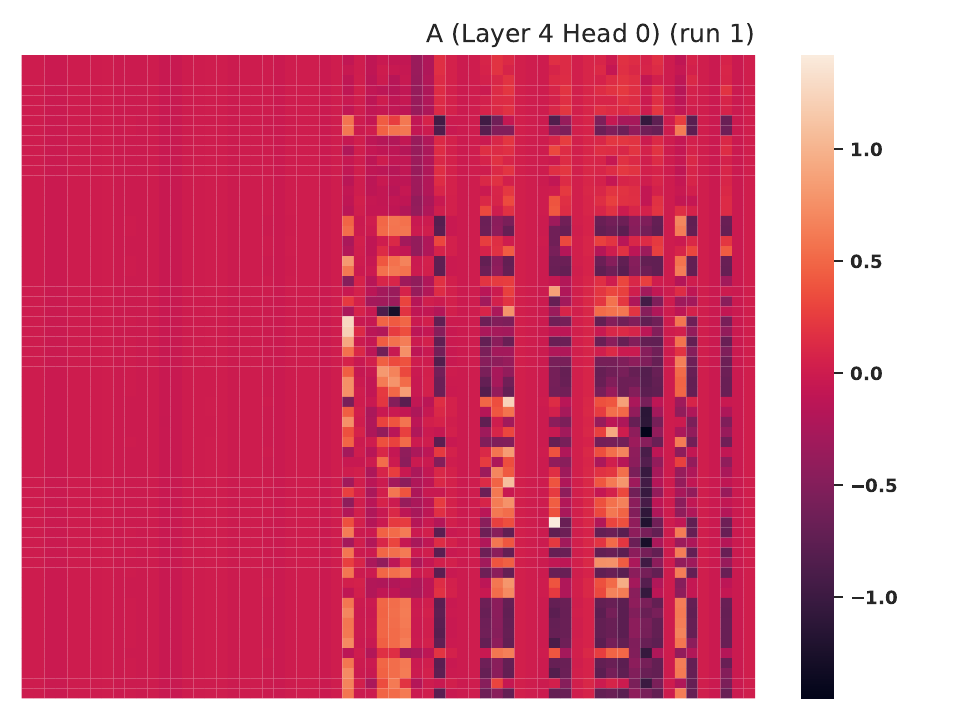}
		\caption{ABL-SOC-110M-2}
		\label{AhmABLSOC110M2}
	\end{subfigure}
	\hfill
	\begin{subfigure}[b]{0.45\textwidth}
		\centering
		\includegraphics[width=1\textwidth]{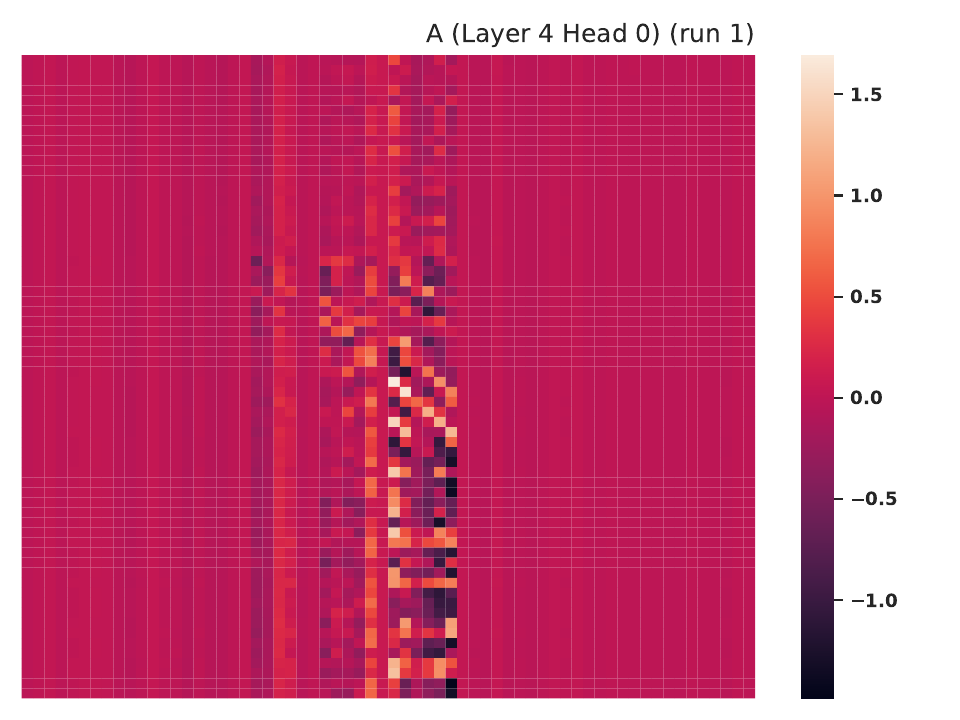}
		\caption{ABL-SOC-110M-3}
		\label{AhmABLSOC110M3}
	\end{subfigure}
	\caption{(Cont.) $\tA$ heat maps at last decoder layer and for a single head of all models averaged over all samples.}
	\label{Ahmhistallmodels}
\end{figure}

\clearpage
\section{Benchmark Datasets}

\textbf{ARC}. The AI2 Reasoning Challenge (ARC) dataset consists of multiple-choice grade school questions from $3^{rd}$ to $9^{th}$ grade. It consists of an easy set and a challenge set. The challenge set contains the questions answered incorrectly by both a retrieval based algorithm and a word co-occurrence algorithm \citep{Clark2018arc}.

\textbf{Hellaswag}. Harder Endings, Longer contexts, and Low-shot Activities for Situations With Adversarial Generations dataset is a commonsense natural language inference dataset that was prepared using adversarial filtering to create problems that are challenging to models, yet easy for humans \citep{Zellers2019hellaswag}.

\textbf{WinoGrande}. WinoGrande is a more challenging version of Winograd Schema Challenge that is a commonsense reasoning benchmark based on a set of pronoun resolution problems designed to be unsolvable for statistical models that rely on selectional preferences or word associations \citep{Keisuke2019winogrande}.

\textbf{TruthfulQA}. TruthfulQA is a benchmark that aims to measure truthfullness of a model. It consists of questions covering 38 categories such as health, law, finance and politics. The model should avoid imitating human contexts in pretraining dataset to perform well, since the questions are selected from the ones humans would answer incorrectly due to a false belief or misconception \citep{Lin2021truthfulqa}.

\textbf{OpenBookQA}. OpenBookQA is a question answering dataset that consists of about 6000 questions accompanied with scientific facts. To answer the questions correctly the model needs to combine with extra common knowledge beyond the facts included in the dataset \citep{Mihaylov2018obqa}.

\textbf{PIQA}. Physical Interaction:Question Answering dataset is a physical commonsense benchmark that aims to evaluate model performance for concepts that are traditionally only seen or experienced in the real world \citep{Bisk2020piqa}.

\textbf{SIQA}. Social Intelligence QA dataset is a social commonsense reasoning benchmark that aims to evaluate model performance for social situations. It consists of 38000 multiple-choice questions for probing emotional and social intelligence in a variety of everyday situations \citep{Sap2019siqa}.

\textbf{IMDB Review}. IMDB Review dataset is a collection of 50000 reviews with each movie having no more than 30 reviews. It was compiled for sentiment analysis and consists of an even number of highly polarized negative ($\leq4$ out of $10$) and positive ($\geq7$ out of $10$) reviews \citep{Maas2011imdb}.

\end{document}